\documentclass[10pt,twocolumn,letterpaper]{article}

\usepackage{iccv}
\usepackage{xcolor}
\usepackage[utf8]{inputenc} 
\usepackage[T1]{fontenc}    
\usepackage{url}            
\usepackage{booktabs}       
\usepackage{amsfonts}       
\usepackage{nicefrac}     
\usepackage{microty pe}     
\usepackage{amsmath}
\usepackage{algorithm}
\usepackage{algpseudocode}
\usepackage{graphicx}
\usepackage{xspace}
\usepackage{multirow}

\usepackage[pagebackref=true,breaklinks=true,letterpaper=true,colorlinks,bookmarks=false,citecolor=blue]{hyperref}

\hypersetup{
pdftitle={Self-Supervised Real-to-Sim Scene Generation},
pdfsubject={Self-Supervised Real-to-Sim Scene Generation}
}

\usepackage{amsmath,amsfonts,bm}

\def\1{\bm{1}}

\DeclareMathAlphabet{\mathsfit}{\encodingdefault}{\sfdefault}{m}{sl}
\SetMathAlphabet{\mathsfit}{bold}{\encodingdefault}{\sfdefault}{bx}{n}

\definecolor{darkblue}{rgb}{0.0, 0.0, 0.55}

\usepackage{url}

\newcommand\blfootnote[1]{%
  \begingroup
  \renewcommand\thefootnote{}\footnote{#1}%
  \addtocounter{footnote}{-1}%
  \endgroup
}

\iccvfinalcopy 

\def\iccvPaperID{6596} 

\ificcvfinal\fi

\begin{document}


\title{Self-Supervised Real-to-Sim Scene Generation}


\author{
\vspace{1mm}
Aayush Prakash 
\hspace{1.0cm}
Shoubhik Debnath
\hspace{1.0cm}
Jean-Francois Lafleche
\\
\vspace{1em}
Eric Cameracci
\hspace{1.0cm}
Gavriel State
\hspace{1.0cm}
Stan Birchfield
\hspace{1.0cm}
Marc T. Law
\\
NVIDIA \\
}

\maketitle
\ificcvfinal\thispagestyle{empty}\fi

\begin{abstract}

Synthetic data is emerging as a promising solution to the scalability issue of supervised deep learning, especially when real data are difficult to acquire or hard to annotate. Synthetic data generation, however, can itself be prohibitively expensive when domain experts have to manually and painstakingly oversee the process. Moreover, neural networks trained on synthetic data often do not perform well on real data because of the domain gap.
To solve these challenges, we propose Sim2SG, a self-supervised automatic scene generation technique for matching the distribution of real data. 
Importantly, Sim2SG does not require supervision from the real-world dataset, thus making it applicable in situations for which such annotations are difficult to obtain. 
Sim2SG is designed to bridge both the content and appearance gaps, by matching the content of real data, and by matching the features in the source and target domains. 
We select scene graph (SG) generation as the downstream task, due to the limited availability of labeled datasets. 
Experiments demonstrate significant improvements over leading baselines in reducing the domain gap both qualitatively and quantitatively, on several synthetic datasets as well as the real-world KITTI dataset.\blfootnote{Correspondence to aayush382.iitkgp@gmail.com, sdebnath@nvidia.com, sbirchfield@nvidia.com, marcl@nvidia.com}

\end{abstract}

\section{Introduction}
\label{sc:intro}

Synthetic data, for which annotations can be automatically generated, is a promising solution to overcome the well-known supervised learning bottleneck of labeling real data.
For this approach to succeed, such synthetic data must look like real data, in both \emph{appearance} and \emph{content}.
Differences in appearance and content together comprise the so-called ``domain gap'' between synthetic and real data~\cite{datasetbias11,meta-sim}.
\emph{Appearance} refers to aspects like color, texture, shape, and lighting, whereas \emph{content} refers to the number, position, and orientation of objects in the scene, as well as their relationships to one another.


At one extreme, synthetic scenes can be created manually by domain experts to reduce these gaps, but such solutions are expensive and therefore do not scale well~\cite{synscapes,RosCVPR16,richter2016playing,VirtualKITTI}.
At the opposite extreme, domain randomization intentionally leverages these gaps to facilitate sim-to-real transfer~\cite{drjosh17,sdr18,tremblay2018training}; these approaches, however, fail to capture the complexity and distribution of real scenes, thus fundamentally limiting their performance.

\begin{figure}
\centering
\includegraphics[width=1.0\linewidth]{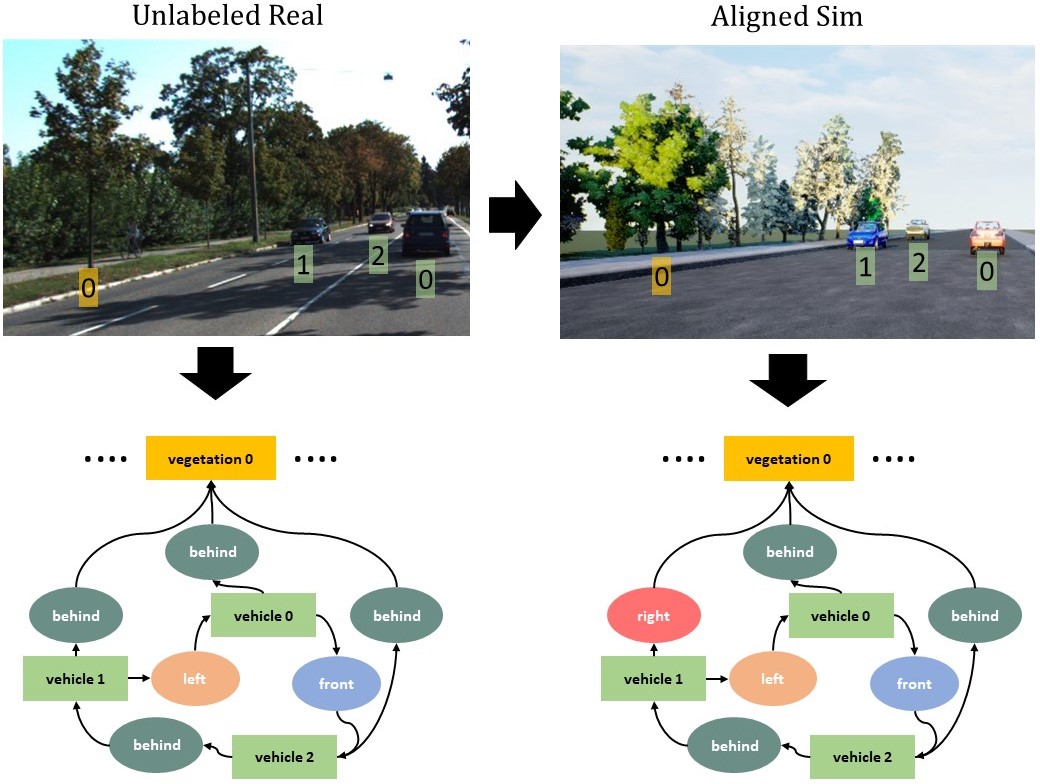}
\caption{We present Sim2SG, a self-supervised real-to-sim scene generation technique that matches the distribution of real data, for the purpose of training a network to infer scene graphs. Sim2SG does not require costly supervision from the real-world dataset.  
(Only one tree is shown in the graph to avoid clutter.)}
\label{fig:teasure}
\end{figure}

Recent approaches such as Meta-Sim~\cite{meta-sim}, Meta-Sim2~\cite{devaranjan2020metasim2}, and SceneGen~\cite{tan2021scenegen} attempt to reduce the content gap by automatically learning to generate synthetic data that matches the distribution of real data.
However, Meta-Sim can only learn the position and orientation of objects, and therefore cannot align the structure of the synthetic scenes (e.g., number and types of objects in a scene) to real data.
Similarly, Meta-Sim2 only learns the distribution of one type of object (cars), and thus cannot match the surrounding context.
Both Meta-Sim and Meta-Sim2 rely on a complex, realistic simulator designed on a set of hand-crafted heuristic rules to ensure that data are generated properly.
SceneGen addresses these limitations but requires access to a large amount of labeled real data, which undermines the underlying purpose of synthetic data.
None of these approaches addresses the appearance gap.

In this paper, we address both the content and appearance gaps, and we do so in a self-supervised manner that requires no real-world labels.
Given an unlabeled real dataset, our method aims to automatically generate synthetic data that matches the distribution of the real data.
See Figure~\ref{fig:teasure}.
We propose Sim2SG (Simulation to Scene Graph), a \emph{synthesis-by-analysis} framework, that generates scenes via self-learning~\cite{pseudolabelyang} loop comprising of two alternating stages: \emph{synthesis}  and  \emph{analysis}. 
During the synthesis stage we leverage a synthetic data generator to create scenes.
By ensuring that the number of objects, as well as their type and placement, are similar, the synthesized data resembles the real data, and therefore the content gap is reduced.
During the analysis stage, we use the generated synthetic data for training.
To further reduce both the content and appearance gaps, the corresponding latent and output distributions are aligned via Gradient Reversal Layers~\cite{ganin2016domain}. 


We show the effectiveness of our method in the scene graph generation task~\cite{Dai_2017,Yang_2018,Zellers_2018}.
A scene graph (SG) summarizes entities in the scene and plausible relationships among them.
The difficulty of hand-labeling scene graphs has limited the community to a small number of datasets~\cite{VRD,visual_genome}.
We experimentally demonstrate our method in three distinct environments: synthetic CLEVR~\cite{johnson2017clevr}, synthetic Dining-Sim, and real KITTI~\cite{kitti}. 
We nearly close the domain gap in the CLEVR environment and show significant improvements over respective baselines in Dining-Sim and KITTI. 
Through ablations, we validate our contributions regarding appearance and content gaps.

\textbf{Contributions:} Our contributions are three-fold: (1)~To the best of our knowledge, we are first to do self-supervised aligned scene generation.
(2)~We propose a novel synthesis-by-analysis framework that addresses both the content and appearance gaps without using any real labels.
(3)~Experimentally, we show that Sim2SG obtains significant improvements on downstream tasks over baselines in all three scenarios:  CLEVR, Dining-Sim, and KITTI. We also present ablations to illustrate the effectiveness of our technique. 
\section{Related Work}
\label{related_work}

\textbf{Synthetic Data}
has been used for many tasks including, but not limited to, object detection \cite{meta-sim,sdr18,tremblay2018training}, semantic segmentation \cite{richter2016playing,RosCVPR16,apostolia2017arxiv}, optical flow modeling \cite{Sintel,DFIB15iccv}, scene flow \cite{mayer2015arx}, classification \cite{pmlr-v139-acuna21a,borrego2018arxiv}, stereo \cite{qiu2016arx:uncv,zhang2016arx:unst}, 3D keypoint extraction \cite{Suwajanakorn2018nips:latent}, object pose estimation~\cite{Mueller2017ue4, tremblay2018deep} and
3D reconstruction \cite{mccormac2016scenenet}.
There are also several simulators~\cite{Dosovitskiy17, robothor, kolve2017ai2,to2018ndds,unity2020perception,denninger2019blenderproc,xiang2020sapien} available for generating synthetic data. 
However, to the best of our knowledge, synthetic data has not been applied to scene graph generation. 

\textbf{Domain Gap} 
is the performance gap when the network is trained on a synthetic source domain and evaluated on real target data. 
Most prior work addresses the appearance gap by image translations~\cite{Chen_2019,studentteachervisda,cycada,munit,gradgan,cyclegan}, clever feature alignment~\cite{Chen_2018,Li_2019,Luo_2019,Saito_2019,Xu_2020,li2020spatial} and domain randomization~\cite{sdr18,drjosh17,tremblay2018training}. 
There are few works which handle the content gap~\cite{azizzadenesheli2019regularized,lipton2018detecting, tan2019generalized,zhao2019learning} by addressing the \emph{label shift} between the two domains. However, they do not exploit the unlabeled images from the target domain. We, on the other hand, leverage the images from the target domain to reduce the domain gap further. To this end, we exploit a scene graph generation task which is more complex than classification and relies on self-training \cite{pseudolabelyang}. 
The idea of self-training with \emph{pseudo labels} is used in \cite{Li_2019,tan2019generalized} to learn models from the target distribution. However, the labels predicted by the model on the target are often inaccurate because of the domain gap~\cite{zheng2020rectifying}.
We instead propose to rely on a synthetic data generator to produce accurate labels.
Some approaches \cite{Chang_2019, sun2019unsupervised} similar to ours also train their task model on top of domain invariant features for image classification and image segmentation. However, they do not address the content gap.

\textbf{Scene Generation} has been used in several machine learning driven approaches.
For instance, some methods propose ways to learn how to synthesize indoor scenes~\cite{wang2019planit, zhou2019scenegraphnet, qi2018human, li2019grains}.
LayoutVAE~\cite{jyothi2019layoutvae} generates scenes conditioned on label sets.
Deep priors and models~\cite{wang2018deep,ritchie2019fast} synthesize scenes by sequentially placing objects. 
A few techniques~\cite{zhou2019scenegraphnet, wang2019planit} also represent scenes as scene graphs with relationships among objects in the same way as our approach.
SceneGen~\cite{tan2021scenegen} learns to generate traffic scenes by modeling the attributes of all objects.
However, all these methods either do not address the content gap with real data or use labeled real data for supervision. 
Meta-Sim~\cite{meta-sim} learns the distribution of position and rotation of objects in the scene and also requires small labeled real data.
Meta-Sim2~\cite{devaranjan2020metasim2} additionally learns the distribution of number and type of objects in the scene without needing labeled real data. However, it does not match the distribution of context. Both Meta-Sim and Meta-Sim2 rely on a simulator designed on handcrafted rules. Unlike Meta-Sim and Meta-Sim2, our method captures relationships among objects and therefore generates more accurate scenes.




\section{Reducing the domain gap}
\label{sec:domgap}

\begin{figure*}
\centering
\includegraphics[width=1.0\linewidth]{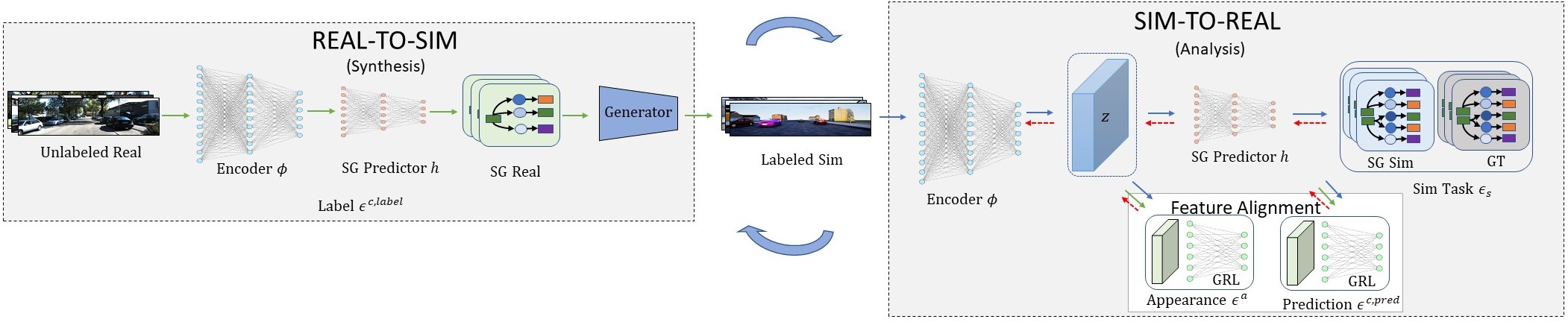}
\caption{Overview of Sim2SG, a self-supervised synthesis-by-analysis framework that generates synthetic data in a loop comprising of two alternating stages: synthesis and analysis.  During synthesis (real-to-sim), we infer scene graphs from real data to generate synthetic data that matches the distribution of real data, thus bridging the content label gap, $\epsilon^{c,label}$.
During analysis (sim-to-real), we train a scene graph prediction network ($\phi$, $h$) using the synthetic data. Additionally, Gradient Reversal Layers (GRLs) are used to align the latent features $z$ and output features $h(z)$ to bridge the appearance $\epsilon^{a}$ and content prediction $\epsilon^{c,pred}$ gap, respectively.}

\label{fig:overview}
\end{figure*}

Let $\langle x_r, y_r \rangle \sim q(x,y)$ be the real data and labels (where the labels $y_r$ are not known). 
Our goal is to generate synthetic data  $\langle x_s, y_s \rangle  \sim p(x,y)$ such that the distributions $p$ and $q$ match (\ie, $p \approx q$).
We assume that both real and synthetic domains share the same categories of objects as well as scenario (\eg, both are driving scenes). 
The difference between $p$ and $q$ is the domain gap.





We now study this domain gap between synthetic and real domains. 
Let $\phi$ be a network that encodes an input image $x$ into a latent representation $z \in \mathcal{Z}$, and let $h$ be a network that estimates some desired quantity from $z$.  (In our case, $h$ infers a scene graph.)
We consider as in~\cite{wu2019domain}
that the task error in the real domain, $\epsilon_r(\phi, h)$, is a function of three terms:
\begin{equation}
\begin{split}
    & ~~~~~~ \epsilon_r(\phi, h) = \int q(z) e_r \, dz \\ 
                        & =  \underbrace{\int p(z) e_s dz}_{\epsilon_s(\phi, h)} + \underbrace{\int q(z) (e_r - e_s) dz}_{\epsilon^{c} (\phi, h)} +
                        \underbrace{\int (q(z)- p(z))e_s dz}_{\epsilon^{a} (\phi, h)} 
\end{split}
\label{eq:target}
\end{equation}
where $e_r \equiv |h(\phi(x_r)) - y_r|$ is the real risk, and similarly $e_s \equiv |h(\phi(x_s)) - y_s|$ is the synthetic risk; and where the distributions of shared features for synthetic and real domains are denoted by $p(z)$ and $q(z)$, respectively.
In this equation, $\epsilon_s(\phi, h)$ is the training error on the synthetic domain, $ \epsilon^{c} (\phi, h)$ is the risk gap between the domains, and $ \epsilon^{a} (\phi, h)$ is the feature gap between the two domains. 
In the following, we drop the terms  $\phi, h$ for notational simplicity. 


If the error $\epsilon_r$ reduces to zero on the target domain, we have closed the domain gap.
To reduce this error, we must reduce both 
$\epsilon^{c}$ and $ \epsilon^{a}$.
We call the former the \emph{content gap} because it refers to the difference in the distributions of the two domains regarding the number of objects and their class, placement, pose and scale.
The difficulty of minimizing the content gap is that computing it is intractable since we do not have access to labels $y_r$ from the real domain, and hence $e_r$ cannot be computed. 
We call the gap $ \epsilon^{a}$ the \emph{appearance gap} because it refers to differences in texture, color, lighting, and reflectance of the scene.
Recent work~\cite{geirhos2018imagenettrained} also shows that latent representations $z$ frequently account for appearance.
In the following, we describe our approach to reduce both the content and appearance gaps.

\section{Simulation to Scene Graph (Sim2SG)}

\begin{figure*}
\vspace{-2mm}
    \centering
    \addtolength{\tabcolsep}{-4.6pt}
    \begin{tabular}{cccc}
    
      \includegraphics[width=0.24\textwidth]{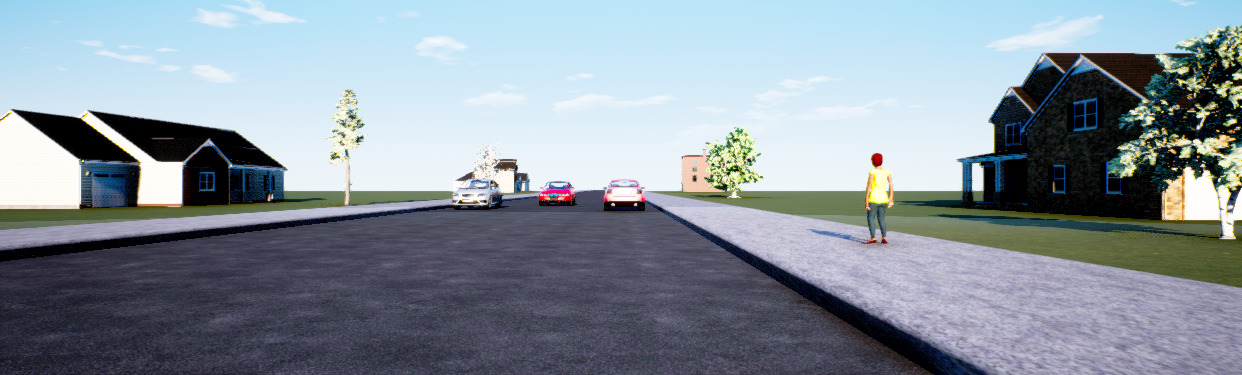} &
      \includegraphics[width=0.24\textwidth]{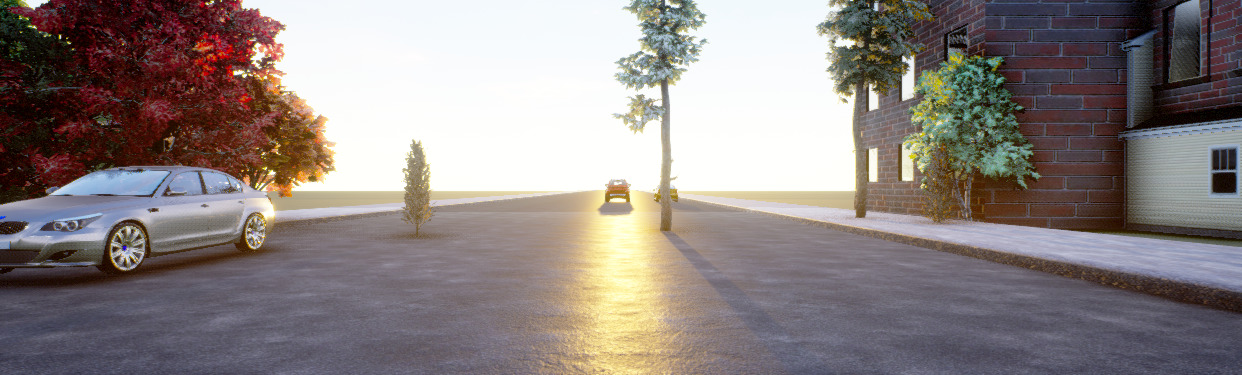} &
      \includegraphics[width=0.24\textwidth]{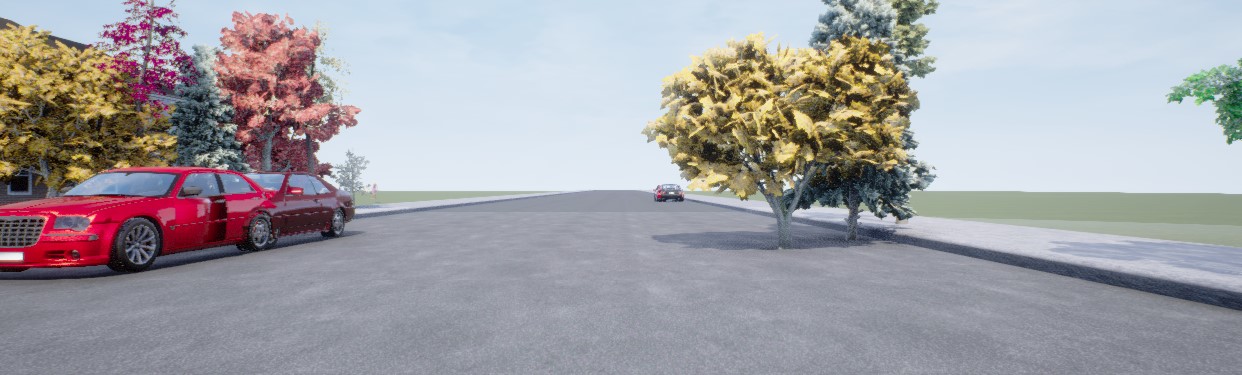} &
      \includegraphics[width=0.24\textwidth]{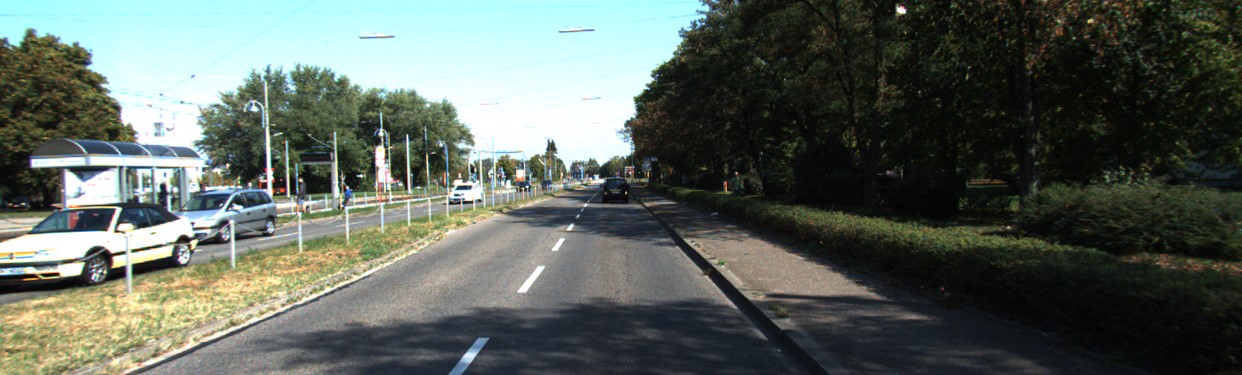} \\

       \includegraphics[width=0.24\textwidth]{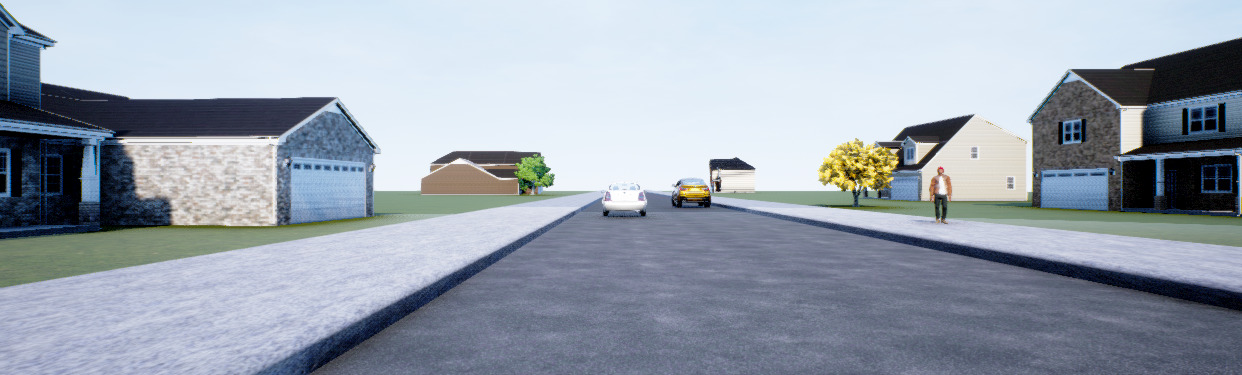} &
      \includegraphics[width=0.24\textwidth]{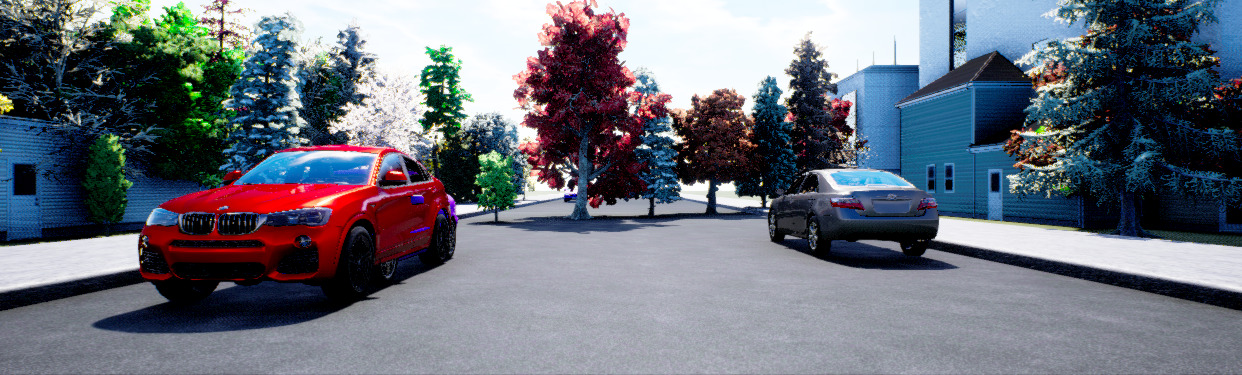} &
      \includegraphics[width=0.24\textwidth]{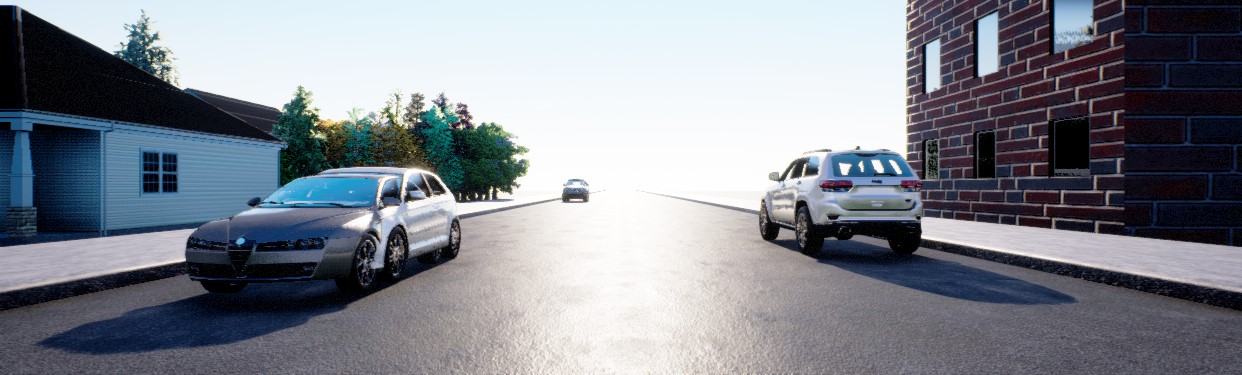} &
      \includegraphics[width=0.24\textwidth]{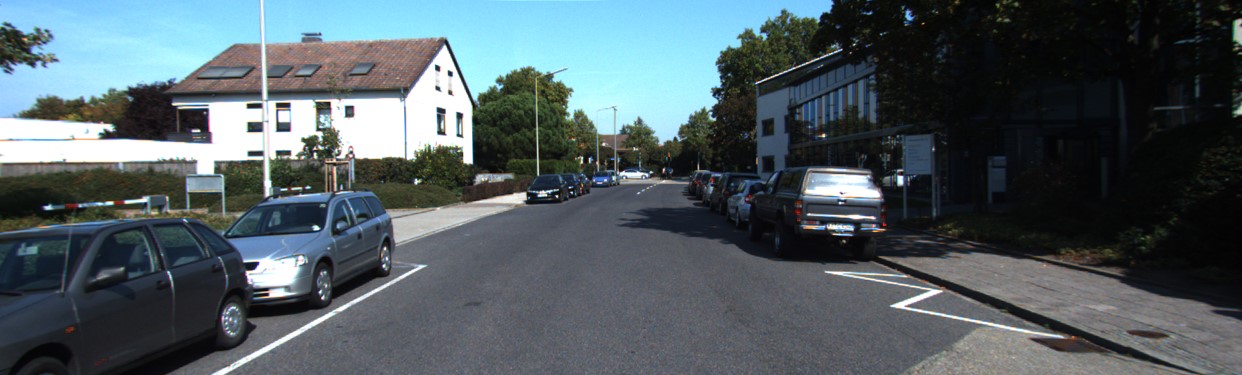} \\

       \includegraphics[width=0.24\textwidth]{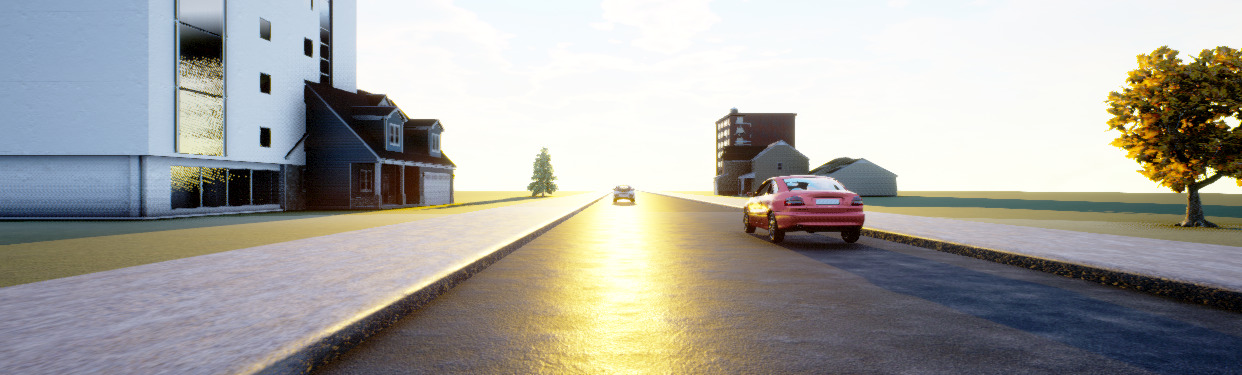} &
      \includegraphics[width=0.24\textwidth]{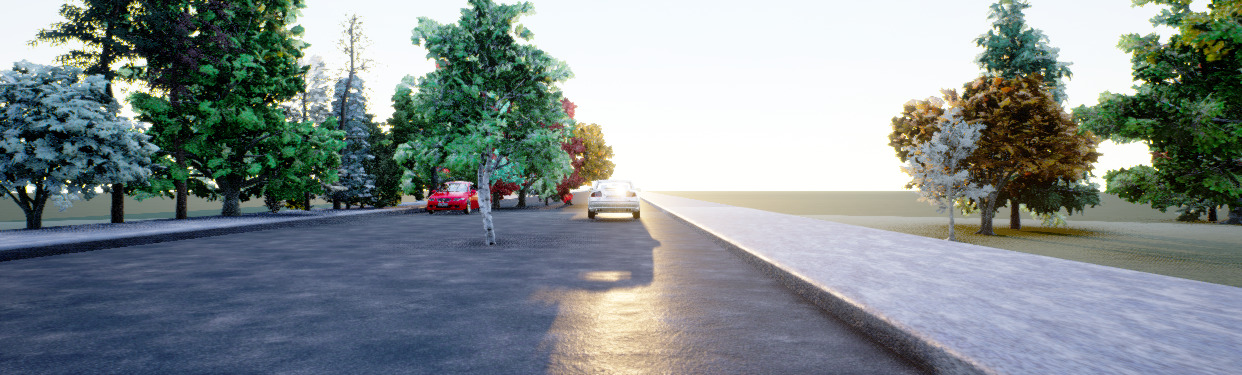} &
      \includegraphics[width=0.24\textwidth]{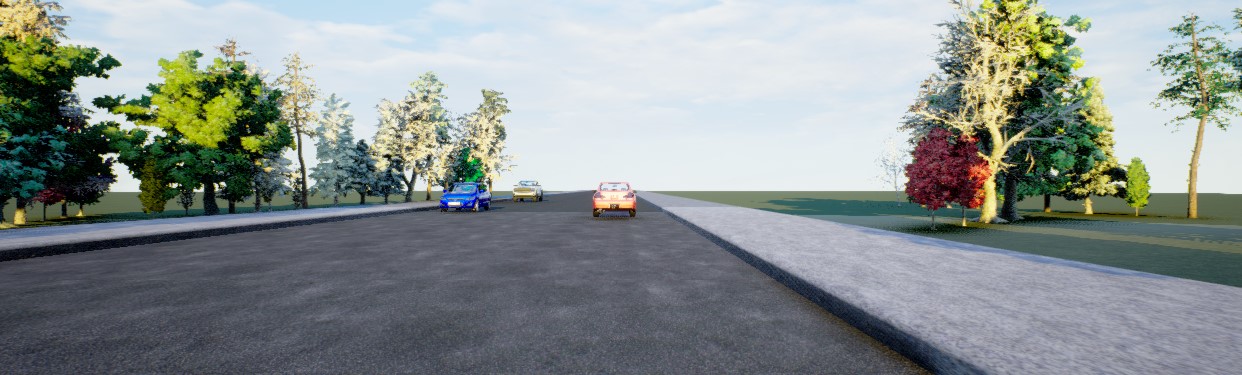} &
      \includegraphics[width=0.24\textwidth]{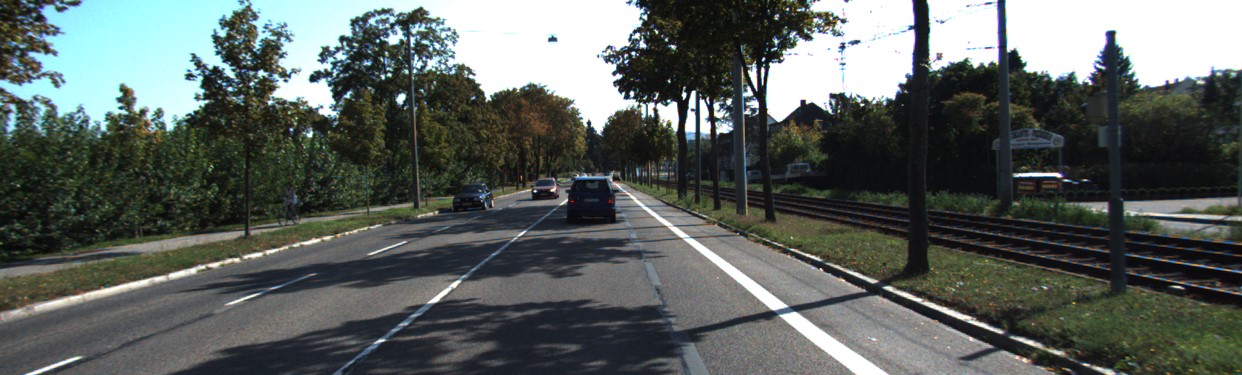} \\
      

       \includegraphics[width=0.24\textwidth]{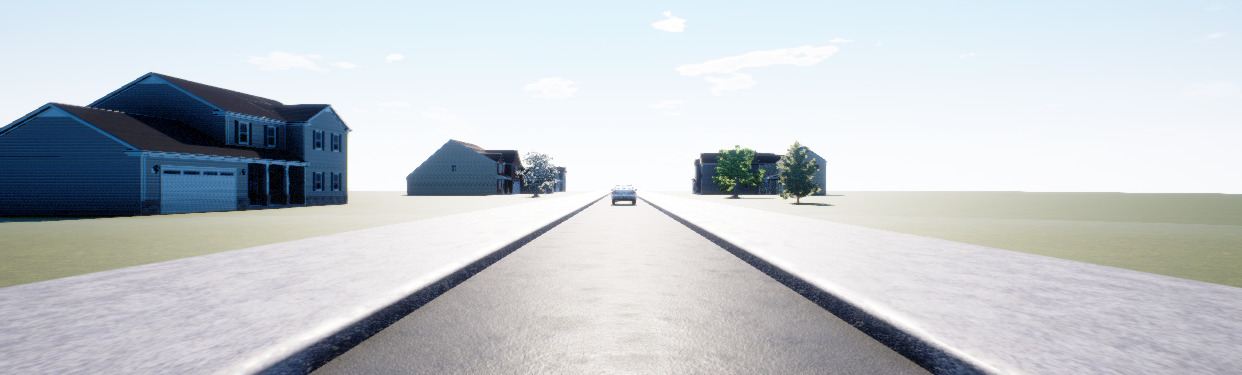} &
      \includegraphics[width=0.24\textwidth]{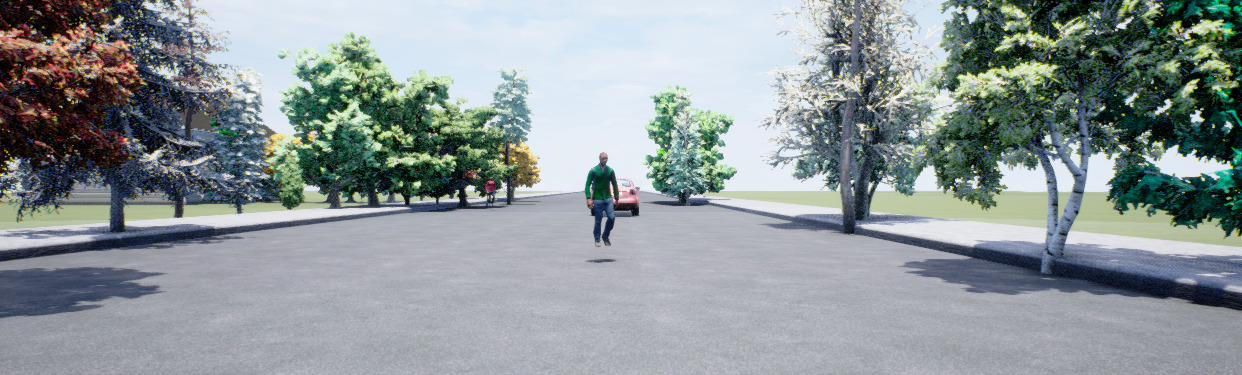} &
      \includegraphics[width=0.24\textwidth]{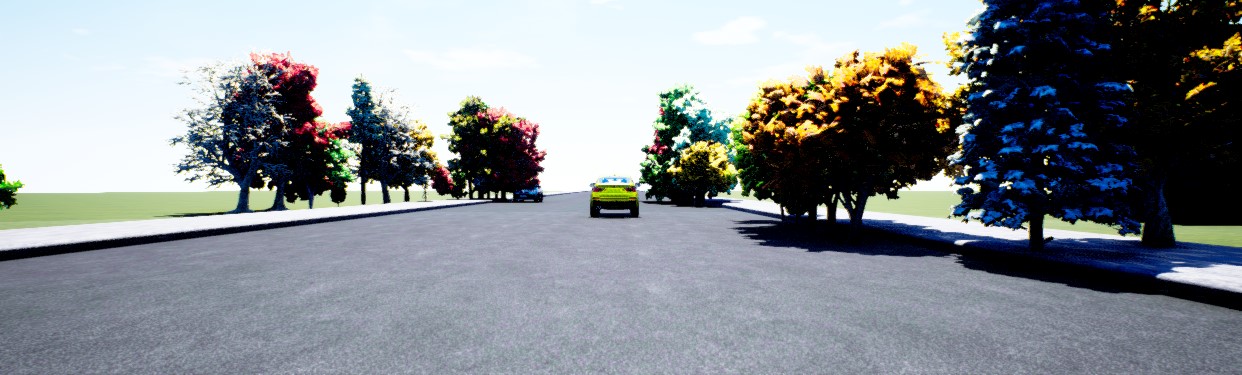} &
      \includegraphics[width=0.24\textwidth]{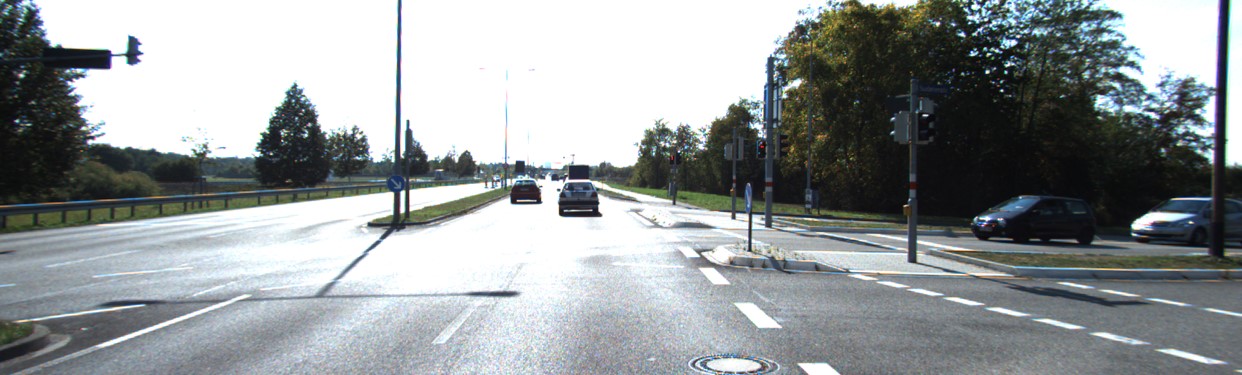} \\

       \includegraphics[width=0.24\textwidth]{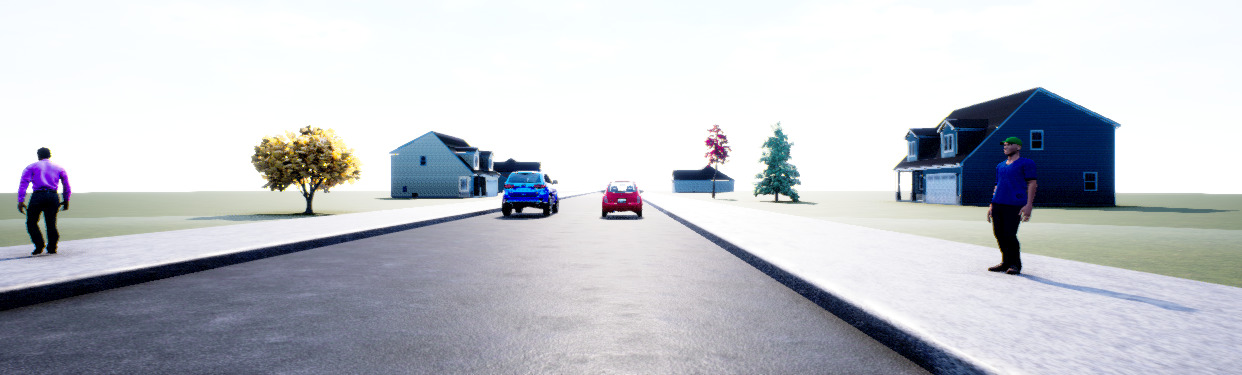} &
      \includegraphics[width=0.24\textwidth]{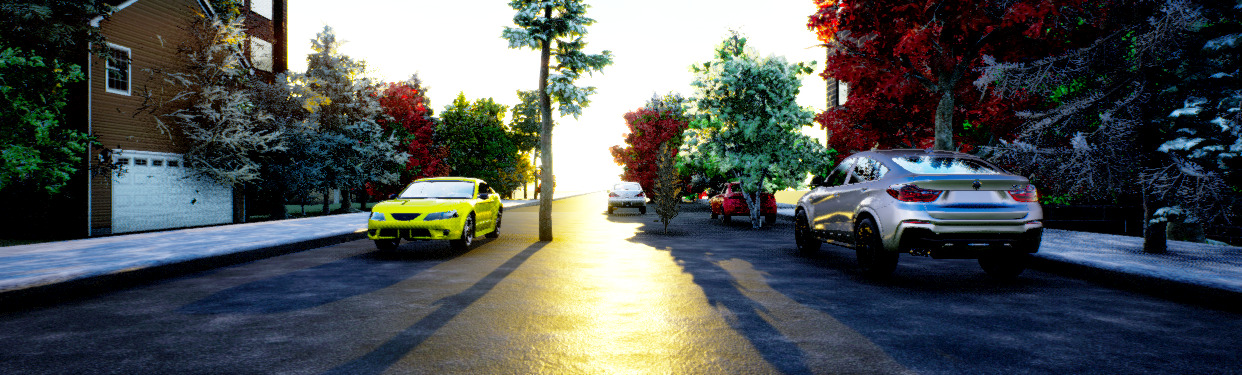} &
      \includegraphics[width=0.24\textwidth]{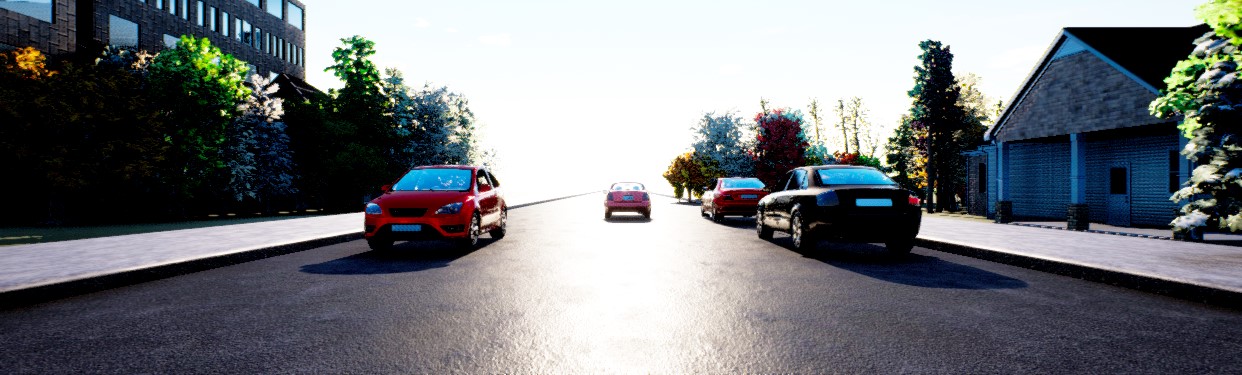} &
      \includegraphics[width=0.24\textwidth]{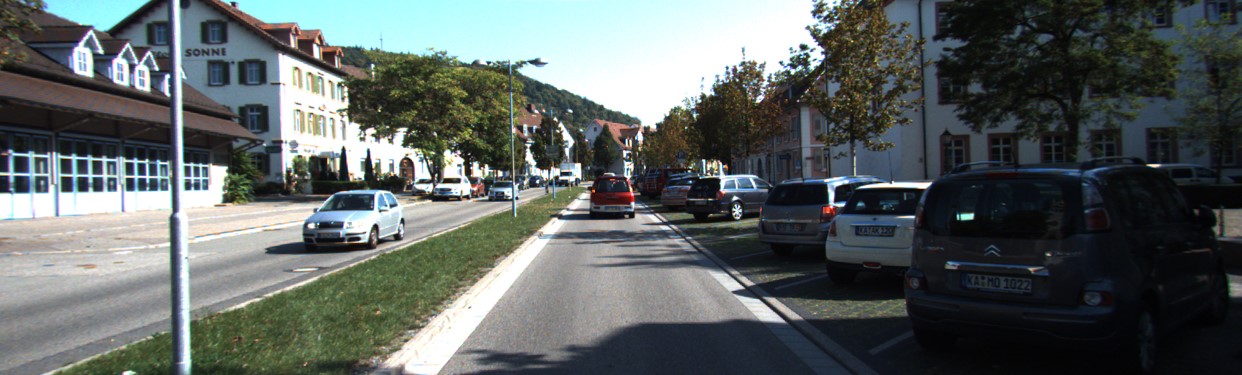} \\
	  
	  Initialization & First epoch & Third epoch & Target\\
    \end{tabular}
    \caption{As the self-learning loop progresses, synthetic data increasingly matches the content of the target image. From left to right: synthetic data initialized by SDR~\cite{sdr18}, after first and third epochs, and corresponding real KITTI target images.}
    \label{fig:data_loop}
    \vspace{-1.0em}
\end{figure*}

Figure~\ref{fig:overview} illustrates our proposed Sim2SG pipeline, which comprises two alternating steps.  
During synthesis (real-to-sim), synthetic data is generated to match the distribution of the unlabeled real data. 
During analysis (sim-to-real), a scene graph (SG) prediction network~\cite{Yang_2018} is trained using the ground truth labels of the synthetic data.
This is analogous to Expectation-Maximization, where synthesis is like the E-step, and analysis is like the M-step.
Results from different iterations of this self-learning loop, which is initialized using synthetic data generated by structured domain randomization (SDR)~\cite{sdr18}, are shown in Figure~\ref{fig:data_loop}.

We now describe the synthesis and analysis steps. Algorithm \ref{algo:train} illustrates the 
pseudocode of both steps.

\subsection{Synthesis Step}
\label{sc:content}

In this step, we aim to generate synthetic data that matches the distribution of unlabeled real data, thereby reducing the content gap.
As mentioned previously, minimizing $\epsilon^{c}$ in the current form is not tractable.
Nevertheless, a sufficient condition for 
$\epsilon^c$ to be zero is for 
$y_s = y_r$ and $h(\phi(x_r)) = h(\phi(x_s))$.
This leads to splitting the content gap into two terms:

\begin{align}
\epsilon^{c}  
        &=  \int q(z) (e_r - e_s) dz \nonumber\\
&\simeq \underbrace{\int q(z) (y_s- y_r)dz}_{\epsilon^{c,label}} 
 +  \underbrace{\int q(z)( h(\phi(x_r)) -  h(\phi(x_s))dz}_{\epsilon^{c,pred}}
\label{eq:content_full}
\end{align}
where $\epsilon^{c,label}$ refers to the gap between the ground truth labels of synthetic and real data, and where 
$\epsilon^{c,pred}$ refers to the gap between the output of the network $h$ trained on synthetic and real data.

{
\begin{algorithm}[t!]
\begin{footnotesize}
\begin{algorithmic}[1]
\State \textbf{Given:} {\tt generator}, $X_r$ \Comment{{\scriptsize Data generator, real images }}

\While{not converged} 
	\State \textcolor{darkblue}{$\triangleright$ Synthesis}
	\State $G \gets \{h\left(\phi(x_r) \right) : x_r \in X_r \}$ \Comment{{\scriptsize Scene graphs}}
    \State $X_s,Y_s \gets$ {\tt generator}$(G)$ \Comment{{\scriptsize Generate aligned synthetic data}}
\State \textcolor{darkblue}{$\triangleright$ Analysis}
\State $loss \gets$ 0
	\For{$(x_s,y_s;x_r) \in (X_s,Y_s;X_r)$}
	\State $loss$ += ${\tt \mathcal{L}^{a}}(\phi(x_s),\phi(x_r))$ \Comment{{\scriptsize Appearance gap}} 
	\State $loss$ += ${\tt \mathcal{L}^{c}}(h(\phi(x_s)),h(\phi(x_r)))$ \Comment{{\scriptsize Prediction gap}}
	\State $loss$ += ${\tt task}(x_s,y_s)$ \Comment{{\scriptsize SG prediction task loss}}
	\State $\phi$, $h \gets$  {\tt optimize}($\phi$, $h$, $loss$) \Comment{{\scriptsize SGD step}}
	\EndFor
\EndWhile
\end{algorithmic}
\end{footnotesize}
\caption{\small{Pseudocode for Sim2SG training}}
\label{algo:train}
\end{algorithm}
}



Since we do not have access to the labels $y_r$ of the target (real) domain, we propose to estimate $y_r$ through a self-supervised method that infers scene graphs from the target data.
Our scene graphs consist of nodes and edges.
Each node $o_i = \langle b_i,c_i \rangle $ consists of a bounding box $b_i = \{u_i,v_i,w_i,h_i\}$ (center, width, and height of the 2D box) and category $c_i$ (such as car, pedestrian, building, etc.).
Each edge captures a relationship $r_i=\langle o_i,p,o_j \rangle$ where $p$ is a predicate (such as behind, left, on, etc.).
Using known camera parameters and a flat ground plane assumption, the scene graph can be mapped to a full 3D scene.
Some parameters (\eg, texture or pose) and some context (ground, sky, lighting), that are not part of the scene graph are randomized.
One advantage of generating synthetic data is that we can augment the relationships of the inferred scene graph by adding new relationships, simply by reasoning in the 3D representation.
A synthetic data generator is then used to render the 3D scenes. 
Note that the weights of our networks  ($\phi$, $h$) remain unchanged during this step.





\subsection{Analysis Step}
\label{sc:style}

In this step, we train the encoder $\phi$ and predictor $h$ of our scene graph (SG) generation model on the synthetic data generated by the preceding step. The task loss function is from Yang \emph{et al.}~\cite{Yang_2018}, which includes cross entropy loss for object classification and relationship classification, and $\ell_1$ loss for bounding boxes. 




Although using synthetic data aligns the content label gap $\epsilon^{c,label}$, it does not align the appearance gap $\epsilon^{a}$ nor content prediction gap  $\epsilon^{c,pred}$.
To address the appearance gap, we align the feature distributions $p(z)$ and $q(z)$,
since $p(z) = q(z)$ is a sufficient condition for $\epsilon^{a}$ to be zero~\cite{wu2019domain}.
To do this, during training $z$ is passed through a Gradient Reversal Layer (GRL)~\cite{ganin2016domain,Chen_2018},
followed by a domain classifier $D^a$ that learns to classify the input as synthetic or real.
The GRL acts as an identity function during forward propagation, but flips the sign of the gradients during back propagation.
The classifier $D^a$ provides an additional loss term for training $\phi$.
We minimize $D^a$'s loss \wrt its own parameters while maximizing \wrt the network parameters of $\phi$.
The loss function for training is as follows:
\begin{equation}
	\mathcal{L}^{a} = -\sum_{x} \Big[ d_i \log {D}^{a}(\phi(x)) + (1-d_i) \log (1-{D}^{a}(\phi(x)))\Big]
	\label{eqn:appearance}
\end{equation}
where $x \in x_s,x_r$,  $d_i=0$ for synthetic images $x_s$ and $d_i = 1$ for real images $x_r$.

Similarly, we seek to minimize the content prediction gap $\epsilon^{c,pred}$ by matching the distributions of the outputs $h(\phi(x_s))$ and $h(\phi(x_r))$ using the same GRL-based technique.
This is based on the observation that the output of the scene graph generation model should be similar for corresponding inputs from the synthetic and real domains. 
During training, the output $h(z)$ is passed through a GRL, followed by a domain classifier $D^c$. 
The classifier provides an additional loss term for training both $\phi$ and $h$. 
The loss function is computed as:
\begin{equation}
	\mathcal{L}^{c} = -\sum_{z} \Big[ d_i \log {D}^{c}(h(z)) + (1-d_i) \log (1-{D}^{c}(h(z)))\Big]
	\label{eqn:content}
\end{equation}
where $z \in \phi(x_s),\phi(x_r)$.
Fully matching the source and target appearance distributions without regard to content may cause detrimental results~\cite{Saito_2019,wu2019domain}. We introduce a warm-up period during which content is aligned without regard to appearance or content prediction.


\section{Experiments}
\label{sc:experiments}


We evaluate Sim2SG in three different environments with increasing complexity:
CLEVR \cite{johnson2017clevr}, our own Dining-Sim using ShapeNet~\cite{chang2015shapenet}, and KITTI~\cite{kitti}. 
In each environment we have a fully labeled synthetic domain and unlabeled target domain with labeled test data.
We use scene graph (SG) generation as the downstream task and implement the encoder
$\phi$ using ResNet-101~\cite{He_2016} 
and the predictor $h$ using Graph R-CNN~\cite{Yang_2018}.

Quantitative evaluation metrics include detection mAP (mean average precision) @0.5 IoU (Intersection over Union in 2D) and relationship triplet recall @20 or @50~\cite{visual_genome}. Note that relationship triplet recall implicitly includes object detection recall. All the mean and standard deviations are based on five runs, and all results are reported after saturation.
Details of the environments, training parameters, and hyperparameters are in the appendix.

For notational simplicity, we use $\sigma^{c,label}$ to refer to our synthesis step (whose purpose is to reduce the content label gap $\epsilon^{c,label}$), $\sigma^{a}$ to refer to the first GRL (whose purpose is to reduce the appearance gap $\epsilon^{a}$), 
and $\sigma^{c,pred}$ to refer to the second GRL (whose purpose is to reduce the content prediction gap $\epsilon^{c,pred}$).
Similarly, we use $\sigma^c$ to refer to the combination of $\sigma^{c,label}$ and $\sigma^{c,pred}$, whose collective purpose is to reduce the content gap.


\begin{figure*}
\addtolength{\tabcolsep}{-4.6pt}
    \centering
    \begin{tabular}{ccc}
    
      \includegraphics[width=0.40\textwidth]{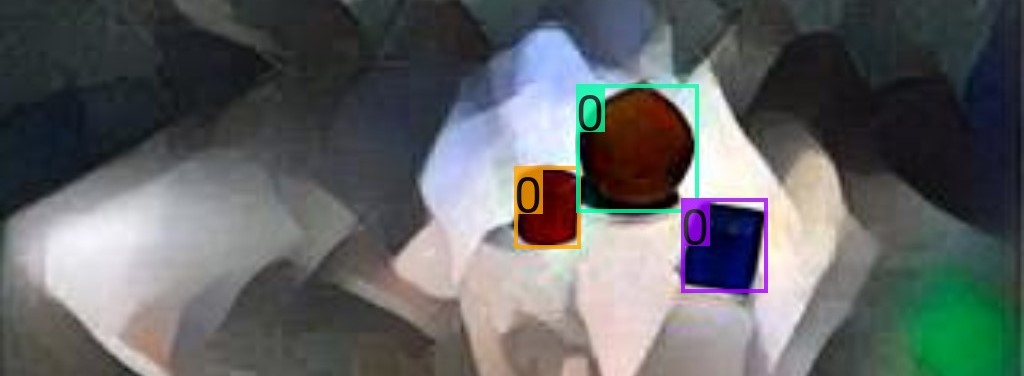} &
      \includegraphics[height=0.16\textwidth]{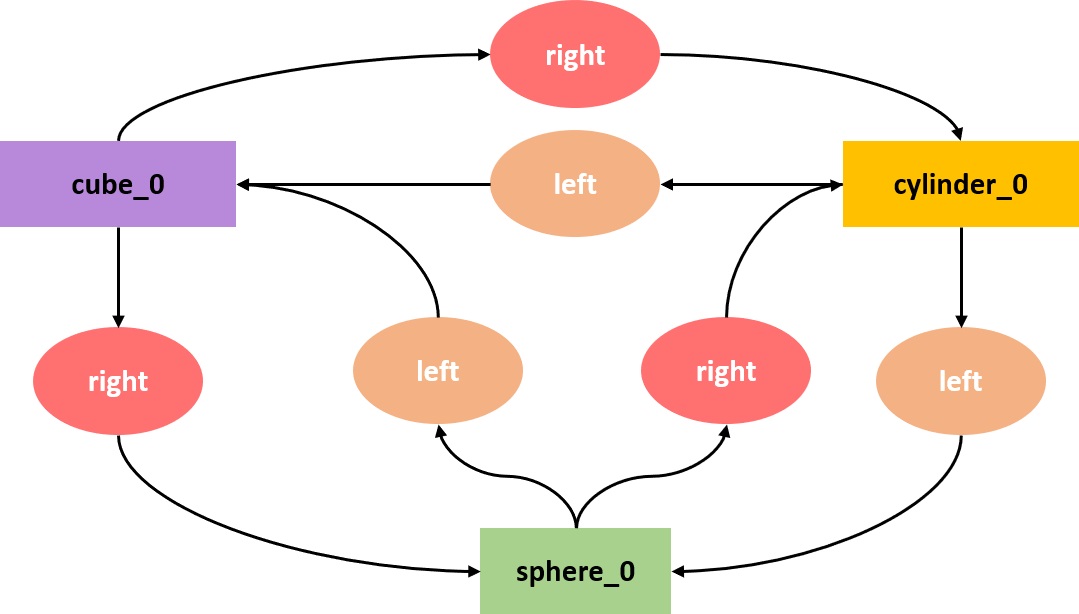} \\

      \includegraphics[width=0.40\textwidth]{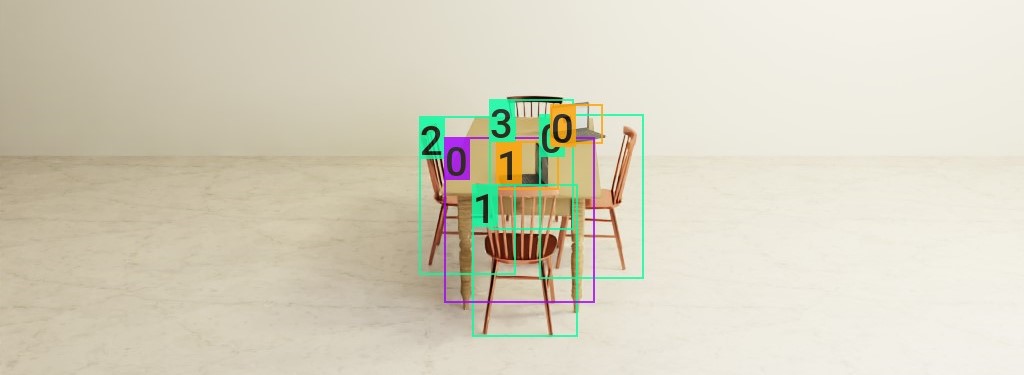} &
      \includegraphics[height=0.15\textwidth]{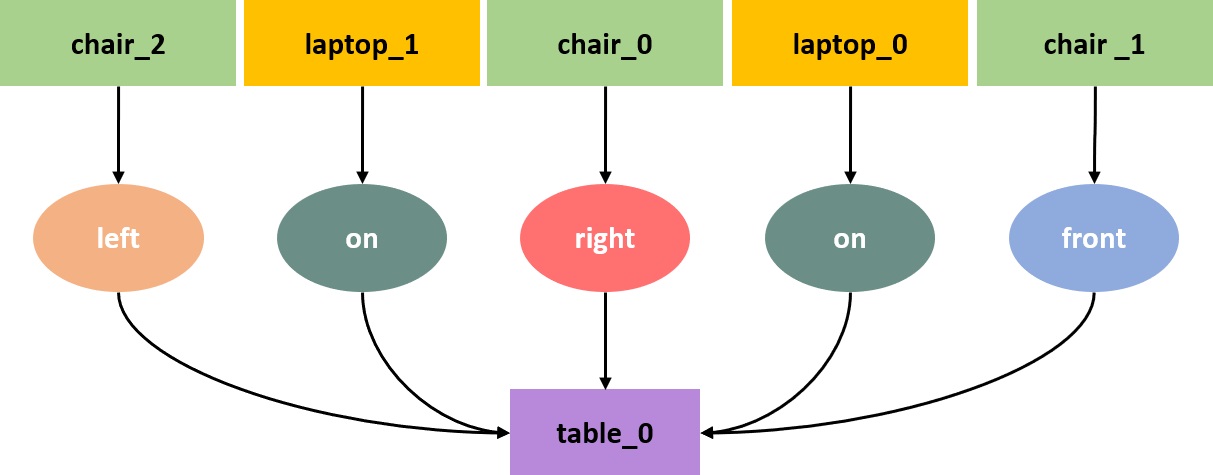} \\

      \includegraphics[width=0.40\textwidth]{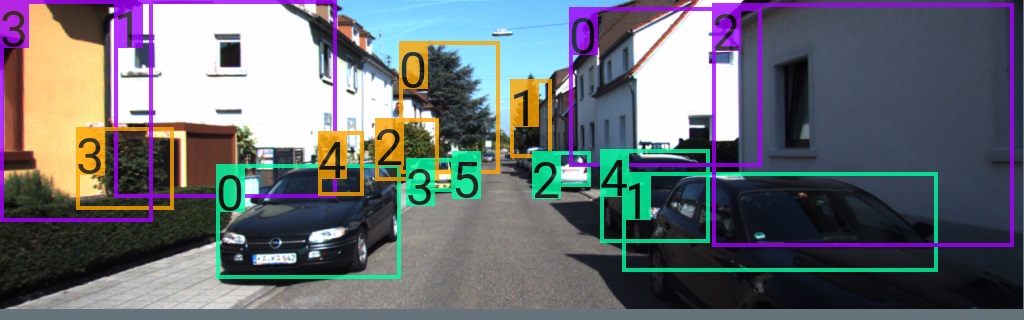} &
      \includegraphics[width=0.55\textwidth]{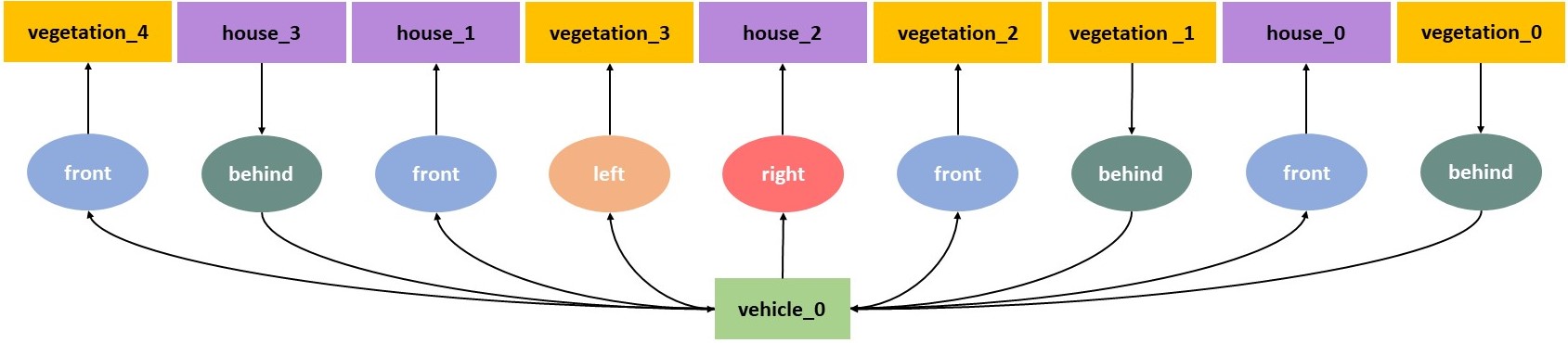} \\
      
    \end{tabular}
    \caption{Qualitative results of Sim2SG on the target domain for CLEVR (top row), Dining Sim (middle row) and KITTI environments (bottom row).
    Objects are color coded. For better visibility, we only show partial scene graph for KITTI.}
    \label{fig:qualitative}
\end{figure*}

\subsection{Experiments on the CLEVR dataset}
\label{subsc:clevr}

The goal of the experiments on the CLEVR environment~\cite{johnson2017clevr} is to show that Sim2SG can address the domain gap in a simple controlled environment. 
There are three classes of objects: cube, sphere and cylinder; and four relationships: front, behind, left and right. 
The source and target domains were generated using disjoint object colors, object materials, margin between objects, and number of objects, to ensure a significant domain gap.
Furthermore, the target images have more complex texture via application of a style transfer network.
We use 1000 labeled images for source domain, 1000 unlabeled images for target domain for training, and 200 labeled source and target images for evaluation. 



\begin{table}[!t]
\centering
\begin{small}
\begin{tabular}{l c c}
\toprule
 Method  
 & mAP@0.5 IoU & Recall@20   \\
 \midrule
SDR~\cite{sdr18}
 &0.723 $\pm{ 0.053}$ & 0.356 $\pm{ 0.047}$ \\
 Ours ($\sigma^{c,label}$) 
 & 0.832 $\pm{ 0.046}$ & 0.493 $\pm{ 0.064}$ \\
 Ours ($\sigma^a$) 
 & 0.821 $\pm{ 0.048}$ & 0.815 $\pm{ 0.026}$ \\
  Ours ($\sigma^a$, $\sigma^{c,label}$)  
  & \textbf{0.892} $\pm{ 0.024}$ & \textbf{0.888} $\pm{ 0.018}$ \\
 \bottomrule
\end{tabular}
\end{small}
\vspace{1mm}
\caption{Results of Sim2SG on the CLEVR target domain.  Aligning both appearance and content yields the best results.}
\label{exp:clevr_3}
\end{table}

\textbf{Results:} 
Quantitative evaluation of Sim2SG is reported in Table~\ref{exp:clevr_3}.
Compared with the baseline using SDR~\cite{sdr18}, our techniques for label alignment $\sigma^{c,label}$ and appearance alignment $\sigma^a$ \emph{drastically reduce} the domain gap.
The best results are achieved by combining these two ideas.
With such a simple environment, however, we found $\sigma^{c,pred}$ to have negligible effect, and therefore it is not reported.
Qualitative results of scene graph recall are shown in first row of Figure~\ref{fig:qualitative}.

\textbf{Ablations:} 
We conduct two ablation experiments on the CLEVR dataset to confirm that appearance alignment  $\sigma^a$ and label alignment $\sigma^{c,label}$ work as intended.
In the first experiment, the source and target domains are created with the same color and texture but different number and placement of objects.
Thus, there is a content gap, but no appearance gap.
We observe that label alignment $\sigma^{c,label}$ closes the domain gap from 0.76 to 0.996 for Recall@20, whereas $\sigma^a$ leads to performance degradation. In the second experiment, the source and target domains have the same number of objects and placement but use different color and texture. 
Thus, there is an appearance gap, but no content gap.
We observe that appearance alignment $\sigma^a$ reduces the domain gap, achieving 0.938 versus 0.339 for Recall@20 for the baseline, whereas label alignment $\sigma^{c,label}$ fails to have significant improvement on relationship triplet recall.
These experiments show that our label alignment procedure $\sigma^{c,label}$ reduces the content gap, while appearance alignment $\sigma^a$ addresses the appearance gap.

\begin{table}
\centering
\begin{small}
\begin{tabular}{l c c}
\toprule
 Method 
 & mAP@0.5 IoU & Recall@50   \\
 \midrule
 SDR~\cite{sdr18}  
 & 0.584 $\pm{0.049}$ & 0.331 $\pm{0.064}$ \\
 Ours ($\sigma^{c,label}$)  
 & 0.713 $\pm{0.038}$ & 0.501 $\pm{0.044}$ \\
Ours ($\sigma^{c}$,  $\sigma^{a}$) 
& \textbf{0.729 $\pm{0.015}$} & \textbf{0.547 $\pm{0.015}$} \\
 \bottomrule
\end{tabular}
\end{small}
\vspace{1mm}
\caption{Results of Sim2SG on the Dining-Sim target domain.  In the last row, $\sigma^c$ includes both $\sigma^{c,label}$ and $\sigma^{c,pred}$.}
\label{exp:dining-sim}
\end{table}

\subsection{Experiments on Dining-Sim}
\label{subsc:shapenet}

We created a dataset that we call Dining-Sim by placing objects created from ShapeNet objects~\cite{chang2015shapenet} in realistic arrangements, with more complex textures and realistic lighting.
This dataset has three classes of objects: chair, table and laptop; and there are five relationships: front, behind, left, right, and on.
Quantitative results, shown in Table~\ref{exp:dining-sim}, agree with the findings of the previous section.
Since this data is more complex, the warm-up period mentioned earlier is necessary, and therefore the label alignment $\sigma^{c,label}$ must be performed first.
This label alignment drastically improves performance on the target domain,
and the combination of label alignment $\sigma^{c,label}$, appearance alignment $\sigma^a$, and prediction alignment  $\sigma^{c,pred}$ achieve the best results.
For reference, the oracle performance on the target domain is 0.904 mAP@0.5 IoU and 0.846 Recall@50.
Qualitative results are illustrated in the second row of Figure~\ref{fig:qualitative}.

\begin{figure*}
\addtolength{\tabcolsep}{-4.6pt}
    \centering
    \begin{tabular}{ccc}
    \multicolumn{1}{c}{\textbf{SDR~\cite{sdr18}}} & 
    \multicolumn{1}{c}{\textbf{Meta-Sim~\cite{meta-sim}}} &
    \multicolumn{1}{c}{\textbf{Sim2SG (ours)}}  \\

	\includegraphics[width=0.3\textwidth]{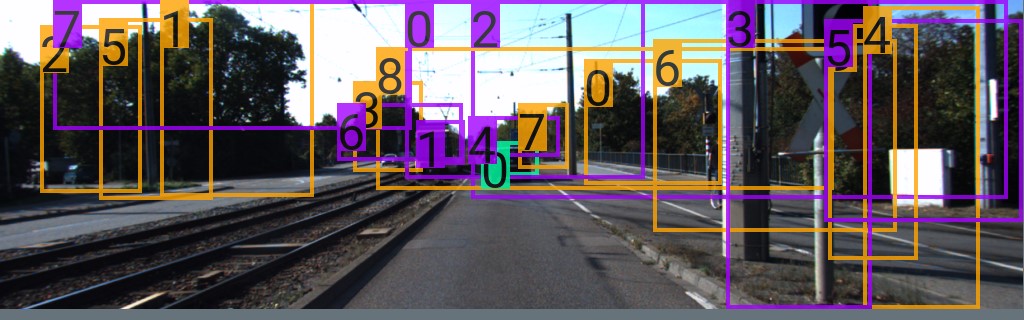} &
	 \includegraphics[width=0.30\textwidth]{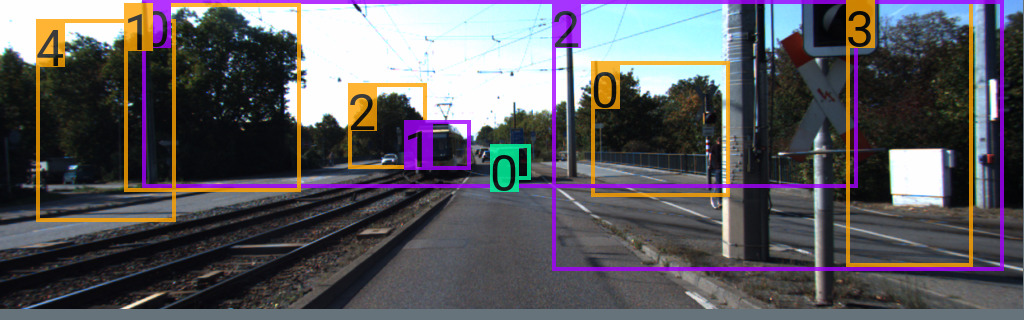} &
      \includegraphics[width=0.3\textwidth]{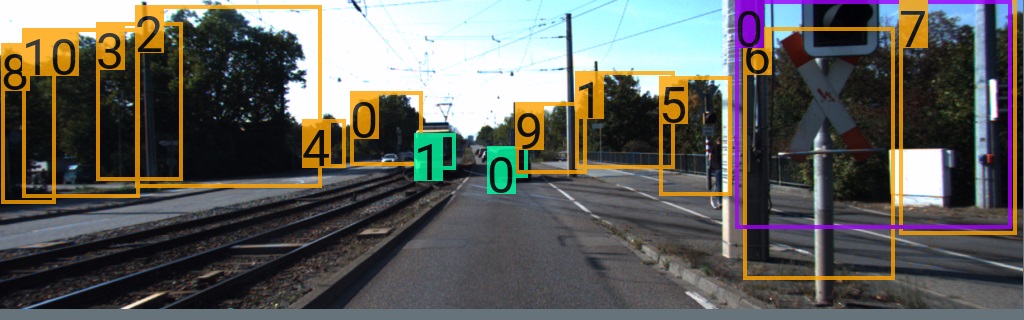} \\
    
    \includegraphics[width=0.3\textwidth]{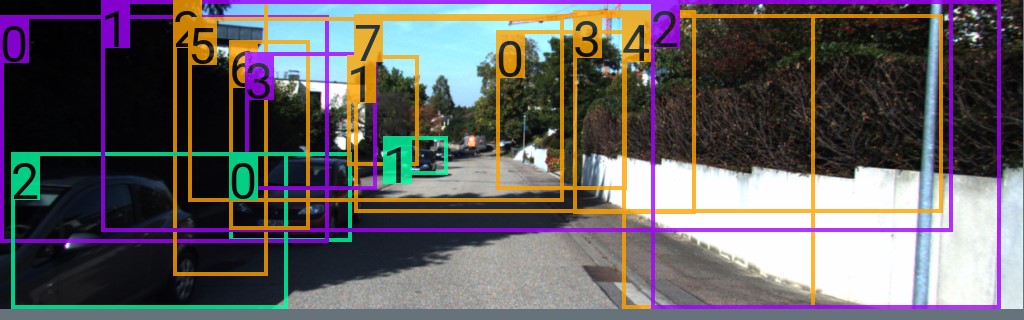} &
        \includegraphics[width=0.3\textwidth]{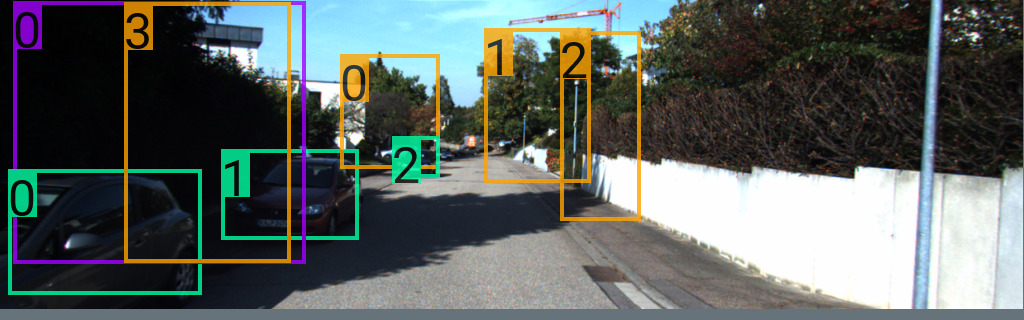} &
      \includegraphics[width=0.3\textwidth]{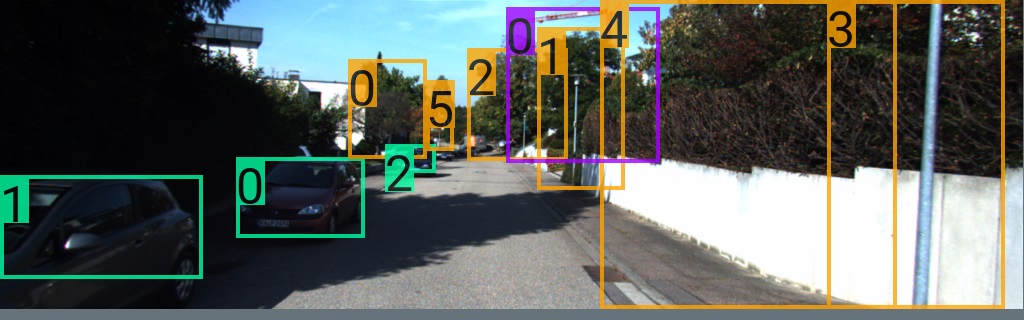} \\
    
       \includegraphics[width=0.3\textwidth]{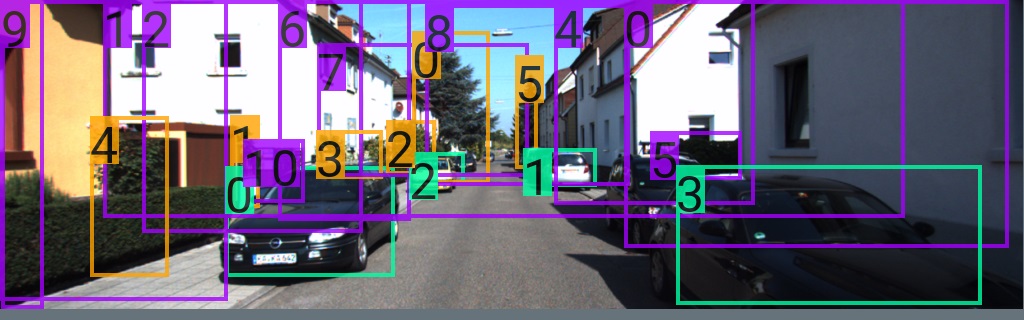} &
       \includegraphics[width=0.3\textwidth]{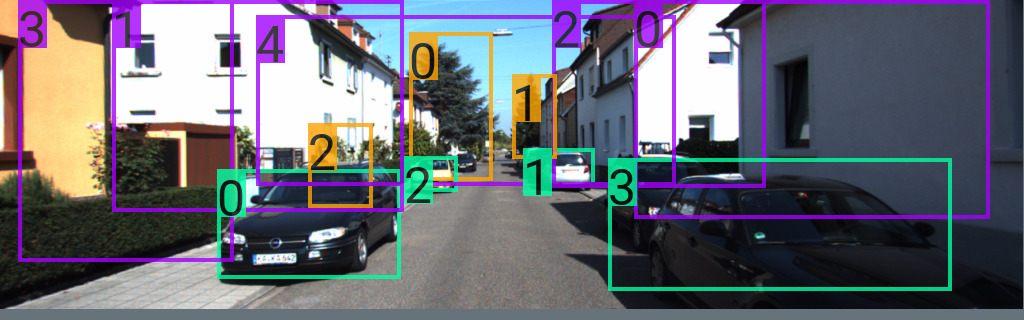} &
      \includegraphics[width=0.3\textwidth]{figures/results/drivesim/label_style_pred/007411.jpg} \\
      
    \end{tabular}
    \caption{Qualitative results of objects detected on three different KITTI images.
    Left: SDR fails to detect many objects and yields a large number of false positives (mislabels), leading to poor scene graphs (not shown). Middle:  Meta-Sim improves on false-positives, but still fails to detect some objects. 
    Right:  Our method detects objects correctly with fewer false positives, thus generating more accurate scene graphs. 
    (Cars in green, vegetation in yellow, buildings in purple.)}
    \label{fig:qualitative_downstream}
\end{figure*}

\subsection{Real-world experiments on KITTI}
\label{subsc:drivesim}
In this section, we validate our approach on the real-world KITTI dataset~\cite{kitti}.
For the synthetic domain, we implemented from scratch a simplified version of SDR~\cite{sdr18}, with a fixed camera, a subset of object classes, no post processing, and without curved road splines, all for faster data generation. 
The number of lanes, sidewalks, and various objects, along with their positions, pose, color, texture, as well as lighting settings are randomly selected.
Note that, unlike SDR, heuristic rules are not used to avoid collisions or ensure realistic placement, which further demonstrates the power of our proposed method in automatically generating useful training data.
We use four object classes: car, pedestrian, vegetation, house, and four relationships: front, left, right, behind. Relationships are constrained so that at least one node is a car, e.g., \textit{car behind car}, \textit{vegetation left car}, and so on.
Although we have shown the `on' relationship to work in the Dining-Sim environment, such relationships are trivial to predict in this driving environment because they are always true: cars are always \emph{on} the road, pedestrians are \emph{on} the sidewalk, and so forth. Therefore, we did not include them in the experiments.

A subset of the real KITTI data~\cite{kitti} from the 2D object detection suite are used as the target domain. 
We use 6000 unlabeled KITTI images for training, and 550 labeled KITTI images for evaluation. 
The latter images include not only the provided annotations for cars and pedestrians, but also annotations that we added for vegetation and houses, along with relationships.\footnote{Our annotations are publicly available at \url{https://research.nvidia.com/publication/2021-08_Sim2SG}}
For training, we used 6000 labeled synthetic images, and another 1000 labeled synthetic images for validation. 


\begin{table*}[!t]
\centering
\footnotesize
\begin{tabular}{l c c c c c c c c}
\toprule
  Method & Car  & Pedestrian  & House  & Vegetation  &  mAP@0.5IoU & Recall@50  \\
\midrule
 SDR~\cite{sdr18} & 
  0.382  $\pm{0.029}$ &	0.168 $\pm{0.017}$ &	0.211 $\pm{0.023}$ & 	0.174 $\pm{ 0.010}$	 & 0.234 $\pm{0.006}$ &	0.070 $\pm{0.007}$ \\
  Meta-Sim~\cite{meta-sim} &  0.413  $\pm{0.009}$ &	0.197 $\pm{0.027}$ &	0.236 $\pm{0.009}$ & 	0.164 $\pm{ 0.023}$	 & 0.253 $\pm{0.003}$ &	0.075 $\pm{0.005}$ \\
 Self-learning~\cite{pseudolabelyang} & 0.312  $\pm{0.006}$ &	0.167 $\pm{0.015}$ &	0.191 $\pm{0.003}$ & 	0.263 $\pm{ 0.006}$	 & 0.233 $\pm{0.004}$ &	0.062 $\pm{0.003}$ \\
 DA Faster R-CNN~\cite{Chen_2018} &  0.424  $\pm{0.028}$ &	0.170 $\pm{0.029}$ &	0.200 $\pm{0.041}$ & 	0.169 $\pm{ 0.014}$	 & 0.241 $\pm{0.014}$ &	0.074 $\pm{0.015}$ \\
 GPA~\cite{Xu_2020} & 0.174  $\pm{0.040}$ &	0.011 $\pm{0.016}$ &	0.106 $\pm{0.031}$ & 	0.059 $\pm{ 0.027}$	 & 0.087 $\pm{0.020}$ &	0.015 $\pm{0.005}$ \\
  SAPNet~\cite{li2020spatial} & 0.362  $\pm{0.054}$ &	0.085 $\pm{0.051}$ &	0.116 $\pm{0.021}$ & 	0.067 $\pm{ 0.022}$	 & 0.157 $\pm{0.024}$ &	-- \\
 Ours ($\sigma^{c,label}$) &  0.410  $\pm{0.009}$ &	\textbf{0.262 $\pm{0.025}$} &	0.240 $\pm{0.010}$ & 	0.229 $\pm{ 0.036}$	 & 0.285 $\pm{0.003}$ &	0.104 $\pm{0.006}$ \\
 Ours ($\sigma^{c,label}$, $\sigma^a$) &  0.493  $\pm{0.004}$ &	0.252 $\pm{0.014}$ &	0.247 $\pm{0.012}$ & 	0.253 $\pm{ 0.020}$	 & 0.311 $\pm{0.311}$ &	0.127 $\pm{0.004}$ \\
 Ours ($\sigma^{c,label}$,  $\sigma^a$, $\sigma^{c,pred}$) & \textbf{0.501  $\pm{0.006}$} &	0.241 $\pm{0.018}$ & \textbf{0.254 $\pm{0.010}$} & 	\textbf{0.269 $\pm{ 0.014}$} & \textbf{0.316 $\pm{0.004}$} & \textbf{0.139 $\pm{0.004}$} \\
 \bottomrule
\end{tabular}
\vspace{1mm}
\caption{Results on KITTI hard after training on labeled synthetic data and unlabeled real data. The class specific AP values for 2D object detection are reported at 0.5 IoU.  The last column shows relationship triplet recall for scene graph generation.}
\label{exp:drive3d}
\end{table*}

\begin{figure*}
\centering
\includegraphics[width=1.0\textwidth]{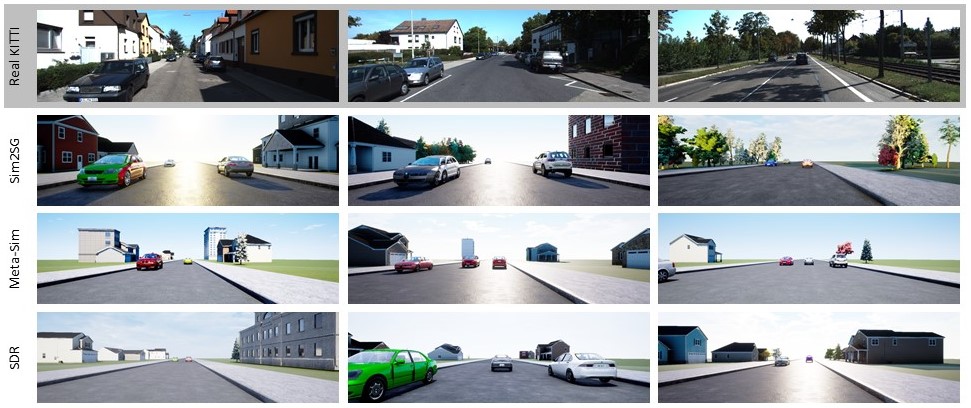} 

    \caption{Synthetic images generated by Sim2SG (ours), Meta-Sim~\cite{meta-sim}, and SDR~\cite{sdr18} with the corresponding KITTI samples. Our method aligns the number and placement of both cars and context (vegetation, houses) better than other methods.}
    \label{fig:synthetic-data}
\end{figure*}

\textbf{Baselines}:
We compare against methods for which code is publicly available.
In particular, we compare against unsupervised scene-generation methods, including structured domain randomization (SDR)~\cite{sdr18} and Meta-Sim~\cite{meta-sim}.
We also compare against a popular self-learning method based on pseudo labels extracted by training directly on the real KITTI data after pretraining on SDR synthetic data~\cite{pseudolabelyang}.
Furthermore, we compare against unsupervised domain adaptation methods for object detection, which are based on aligning features from the source and target domains: DA Faster R-CNN~\cite{Chen_2018}, GPA~\cite{Xu_2020}, and SAPNet~\cite{li2020spatial}. 
According to a recent survey paper~\cite{li2020deep}, the latter two are state-of-the-art leading techniques.
All baselines are trained on 6000 images from SDR, using the hyperparameters provided by the original authors. 



\textbf{Results:} We evaluate these baselines, along with our method, on three KITTI evaluation modes for 2D object detection: easy, moderate and hard, which are based on object size, occlusion and truncation.
For all three modes, our method yields significantly improved results over the baselines. 
Table~\ref{exp:drive3d} shows the object detection and scene graph generation results for KITTI hard; other results are in the appendix.
The last three rows of the table show that most improvements come from label alignment $\sigma^{c,label}$ and appearance alignment  $\sigma^{a}$.
The combination of all three, $\sigma^{c,label}$, $\sigma^{a}$ and $\sigma^{c,pred}$, achieves the best results overall.  
Note that $\sigma^{c,label}$ alone is able to beat most baselines, thus demonstrating the efficacy of our synthesis step at automatically generating realistic training data without heuristics.
Qualitative results are illustrated in the third row of Figure~\ref{fig:qualitative}.

Insight as to why our method outperforms SDR~\cite{sdr18} and Meta-Sim~\cite{meta-sim} can be gained by viewing the images generated by the various methods, shown in Figure~\ref{fig:synthetic-data}.
Our method generates synthetic data that better matches the distribution of the real data, because SDR lacks access to the object distributions in KITTI, and because Meta-Sim cannot align the structure of the scenes (e.g., the number of objects) and lacks the notion of relationships in its scene graphs.
(Note that Meta-Sim2~\cite{devaranjan2020metasim2} also lacks relationships, but code is not publicly available.)
Furthermore, our method scales better with scene complexity compared with Meta-Sim, which requires passing expensive numerical gradients through a renderer.
As a result, our training time is 12 hours on a single NVIDIA V100 GPU, compared with 72 hours for Meta-Sim.
Figure~\ref{fig:qualitative_downstream} compares our method on the downstream task with SDR and Meta-Sim.


The reason our method outperforms domain adaptation techniques~\cite{Chen_2018, Xu_2020,li2020spatial} is because feature alignment without content alignment is not effective, as we discussed briefly at the end of Section~\ref{sec:domgap}.
Nevertheless, it may be possible to combine our label alignment technique $\sigma^{c,label}$ with domain adaptation methods, which we leave for future work.
Our method always produces proper ground truth for the scene via synthetic data generator, even if the scene is not generated exactly according to the corresponding real image.
Self-learning~\cite{pseudolabelyang} methods based on pseudolabels from KITTI, on the other hand, can have errors in the ground truth due to incorrectly classified objects or imprecise bounding boxes, as explained in \cite{zheng2020rectifying}.




Note in Table~\ref{exp:drive3d} that the detection accuracy of the pedestrian category does not improve with $\sigma^{a}$ and $\sigma^{c,pred}$. 
The reason for this limitation is that pedestrians are an under-represented class in KITTI, not to mention  small and hard to detect. 
While our Sim2SG method can align the label distribution, it cannot address class imbalance in the target domain. 
Nevertheless, our method is not restricted to the types or number of relations. Given a simulator to generate scenarios such that objects and their relations are detectable, our method should extend to handle them.

\textbf{Ablations:}
As briefly discussed at the end of Section~\ref{sec:domgap}, we run the label alignment $\sigma^{c,label}$ before appearance alignment  $\sigma^{a}$ and prediction alignment $\sigma^{c,pred}$ to address the fact that feature alignment can be detrimental if the content of both domains are not already aligned. 
We indeed found that performance drops significantly when we train Sim2SG without $\sigma^{c,label}$ and evaluate in the same setting as Table~\ref{exp:drive3d}. Sim2SG with $\sigma^{a}$ and $\sigma^{c,pred}$ gives a 0.246 mAP@0.5 IoU for detection and 0.076 Recall@50 for relationship triplets, while adding $\sigma^{c,label}$ yields a significant boost of 0.316 mAP@0.5 IoU for detection and 0.139 Recall@50 for relationship triplets on KITTI Hard.
These results confirm the effectiveness of $\sigma^{c,label}$ and therefore the importance of the entire Sim2SG framework.

\section{Conclusion}

In this work, we have proposed Sim2SG, a self-supervised real-to-sim automatic scene generation technique that matches the distribution of real data, for the purpose of training a network to infer scene graphs. The method bridges both the content and appearance gap with real data without requiring any costly supervision on the real-world dataset. 
The approach achieves significant improvements over baselines in all three environments for which we tested:  CLEVR, a Dining-Sim dataset, and real KITTI data.
The latter of these demonstrates the ability of our method to perform sim-to-real transfer.


{\small
\bibliographystyle{ieee_fullname}
\bibliography{iccvreferences}
}

\newpage
\appendix
\normalsize
    

\section{Appendix}

\subsection{Architecture details}
\label{appendix:arch}
\paragraph{Encoder $\phi$ and SG Predictor $h$}
We use ResNet 101~\cite{He_2016} pretrained on ImageNet as the encoder neural network. We use the Faster R-CNN~\cite{fasterrcnn} and Graph Convolution Network based architecture from Graph R-CNN~\cite{Yang_2018} to implement the SG Predictor $h$.

\paragraph{GRL}
For appearance alignment $\sigma^a$, we use a 2-layer 2D convolution neural network based discriminator with ReLU activation. For prediction alignment $\sigma^{c,pred}$ we use 2-layer MLP based discriminator. We also scale the gradients to the encoder network $\phi$ from the discriminator by a factor of 4 in above cases. 

\subsection{Experiments}

\subsubsection{CLEVR}
\label{appnd:clevr}
\begin{figure*}
    \centering
    \addtolength{\tabcolsep}{-4.6pt}
    \begin{tabular}{ccccc}
				\raisebox{5ex}{\rotatebox[origin=c]{90}{ Source}}  &
			 \includegraphics[width=0.23\textwidth]{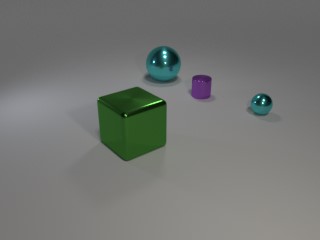} &
			 \includegraphics[width=0.23\textwidth]{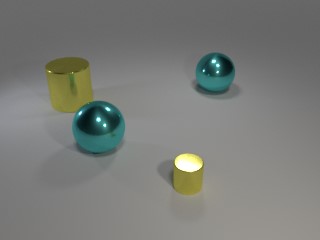} &
      \includegraphics[width=0.23\textwidth]{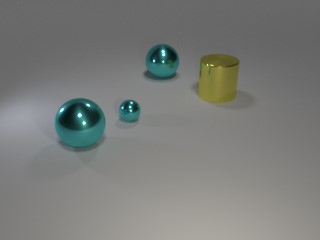} &
      \includegraphics[width=0.23\textwidth]{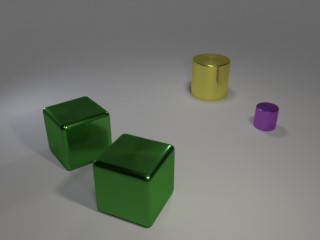} \\
			
			 \raisebox{5ex}{\rotatebox[origin=c]{90}{ Target}} &
      \includegraphics[width=0.23\textwidth]{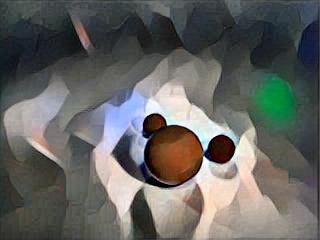} &
			 \includegraphics[width=0.23\textwidth]{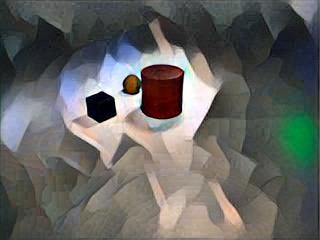} &
      \includegraphics[width=0.23\textwidth]{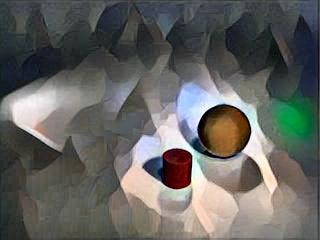} &
      \includegraphics[width=0.23\textwidth]{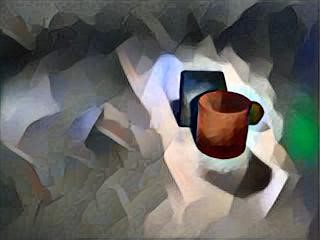} \\
      
      		\raisebox{5ex}{\rotatebox[origin=c]{90}{ Source}}  &
			 \includegraphics[width=0.23\textwidth]{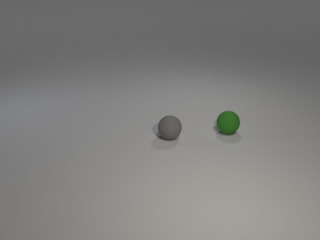} &
			 \includegraphics[width=0.23\textwidth]{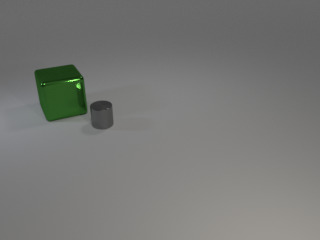} &
      \includegraphics[width=0.23\textwidth]{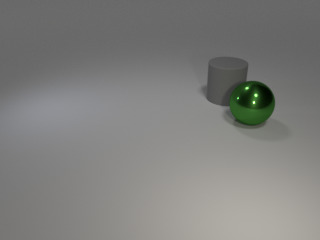} &
      \includegraphics[width=0.23\textwidth]{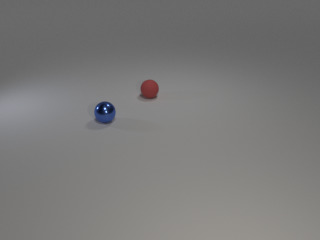} \\
			
			 \raisebox{5ex}{\rotatebox[origin=c]{90}{ Target}} &
      \includegraphics[width=0.23\textwidth]{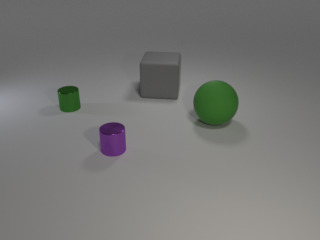} &
			 \includegraphics[width=0.23\textwidth]{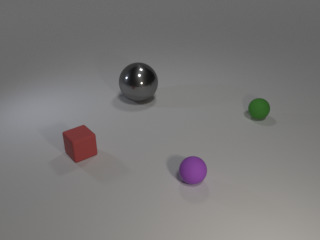} &
      \includegraphics[width=0.23\textwidth]{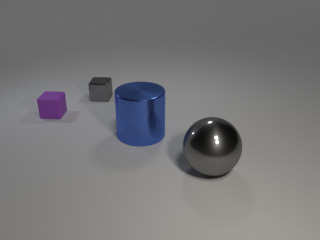} &
      \includegraphics[width=0.23\textwidth]{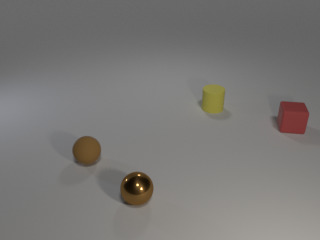} \\
      
      \raisebox{5ex}{\rotatebox[origin=c]{90}{ Source}}  &
			 \includegraphics[width=0.23\textwidth]{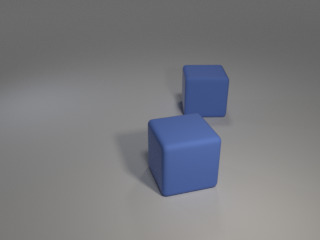} &
			 \includegraphics[width=0.23\textwidth]{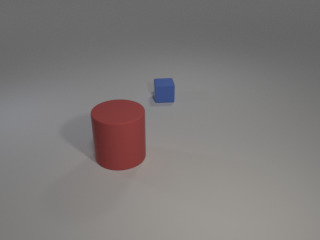} &
      \includegraphics[width=0.23\textwidth]{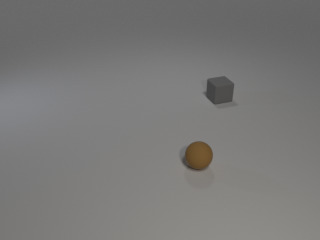} &
      \includegraphics[width=0.23\textwidth]{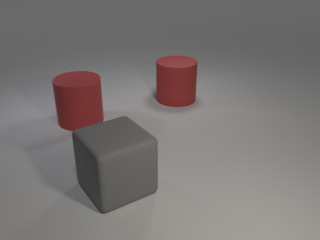} \\
			
			 \raisebox{5ex}{\rotatebox[origin=c]{90}{ Target}} &
      \includegraphics[width=0.23\textwidth]{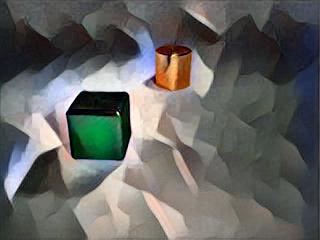} &
			 \includegraphics[width=0.23\textwidth]{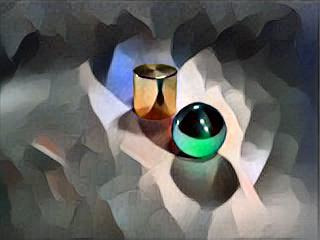} &
      \includegraphics[width=0.23\textwidth]{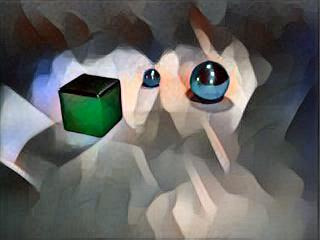} &
      \includegraphics[width=0.23\textwidth]{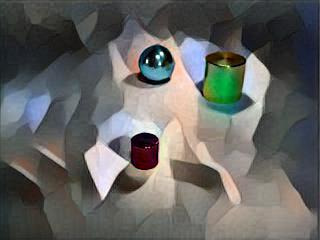} \\
     
    \end{tabular}
    \caption{Samples from source and target distributions from Clevr environment. Row 1-2: Source and Target differ in both appearance and content. Row 3-4: Source and Target differ in content but have same appearance. Row 5-6: Source and Target differ in appearance but have same content.
    }
    \label{fig:clevrdata}
\end{figure*}

\paragraph{Setup}
The source and target domains of the CLEVR~\cite{johnson2017clevr} environment leverage Blender~\cite{blender} to render 320 x 240 images and corresponding ground truth scene graphs.
The source and target domains were generated using disjoint object colors, object materials, margin between objects, and number of objects, to ensure a significant domain gap.
We use 4 objects of colors (blue, green, magenta, yellow) and material (metal) for source domain.
We use 2--3 objects of colors (pink, brown, white) and material (rubber) for target domain. Additionally, we transform the target by using a style transfer network
\footnote{{\scriptsize\url{https://github.com/pytorch/examples/tree/master/fast_neural_style}}}.
For both domains, we sample each class and their size (small, medium \& large) with equal probability.
The environment has three lights and a fixed camera. We add a small random jitter to their initial positions during the rendering process. 
Some samples of source and target domain are shown in  Figure~\ref{fig:clevrdata}.

\paragraph{Training Details}
We train our model for $10^5$ iterations with label alignment $\sigma^{c,label}$ and appearance alignments $\sigma^a$.
We optimize the model using a SGD optimizer with learning rate of $10^{-4}$ and momentum of 0.9. We train our model using a batch size 4 on NVIDIA DGX workstations. We report saturation peak performance in all our tables.
We give equal regularization weights to source task loss $\sigma_s$, appearance alignment $\sigma^a$ and label alignment $\sigma^{c,label}$. 

\begin{figure*}
\addtolength{\tabcolsep}{-4.6pt}
    \centering
    \begin{tabular}{ccc}
    
    \includegraphics[height=0.18\textwidth]{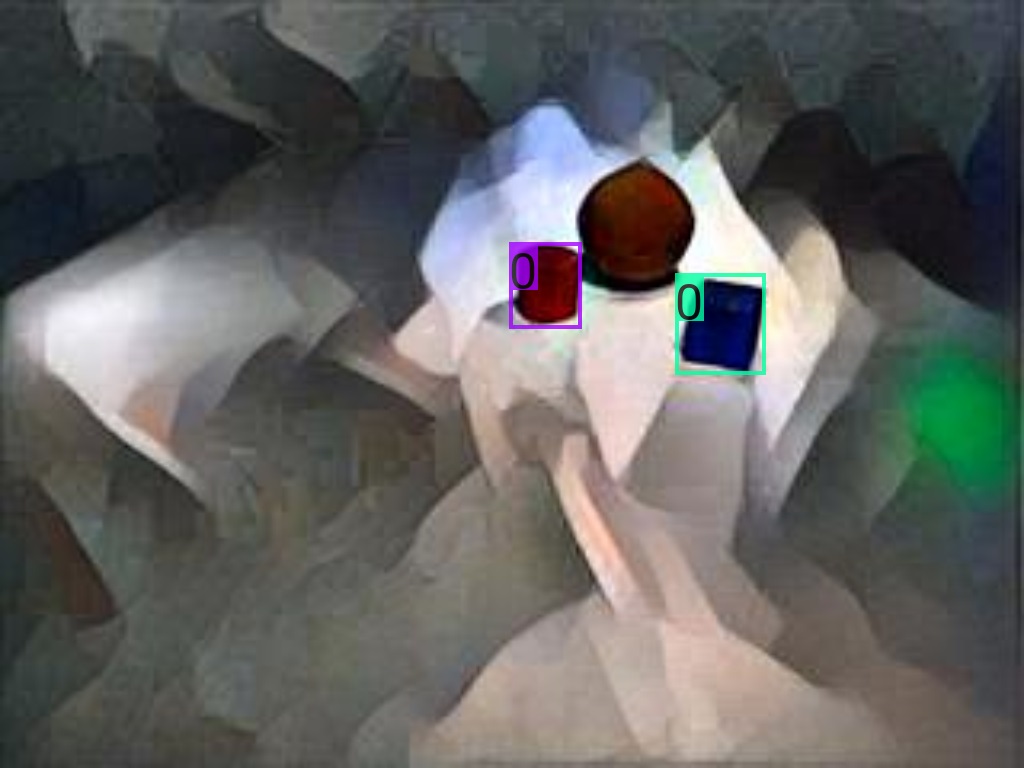} &
      \includegraphics[height=0.18\textwidth]{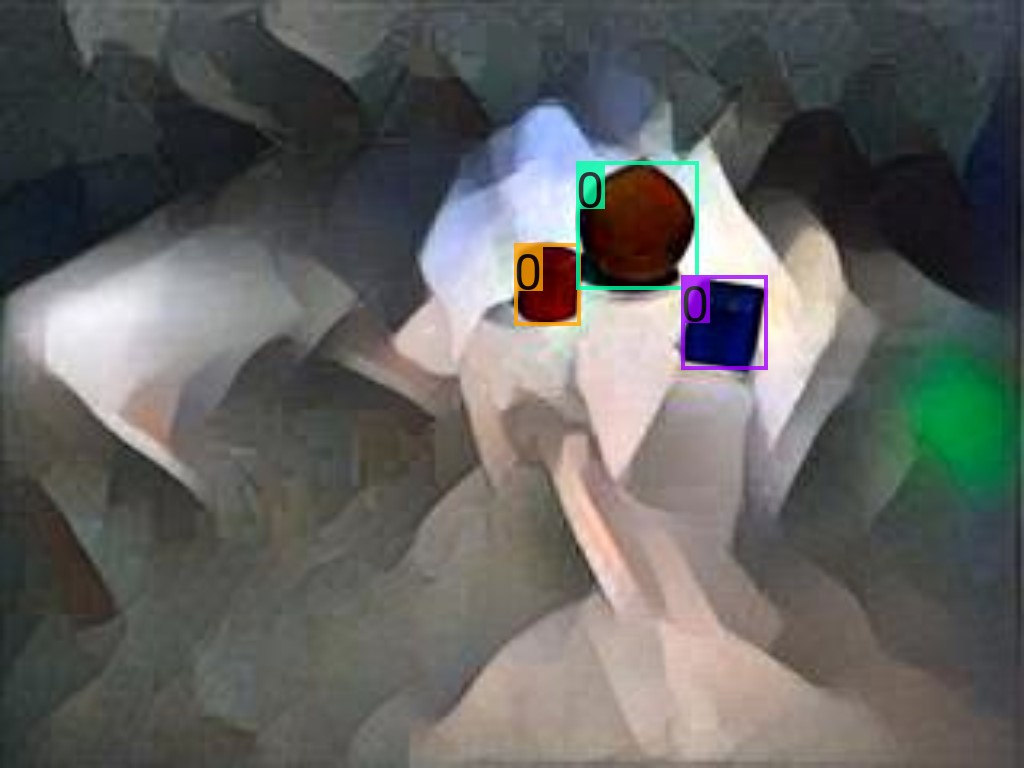} &
      \includegraphics[height=0.18\textwidth]{figures/results/clevr/content_style/clevr_hor_1.jpg} \\
      
    \includegraphics[height=0.18\textwidth]{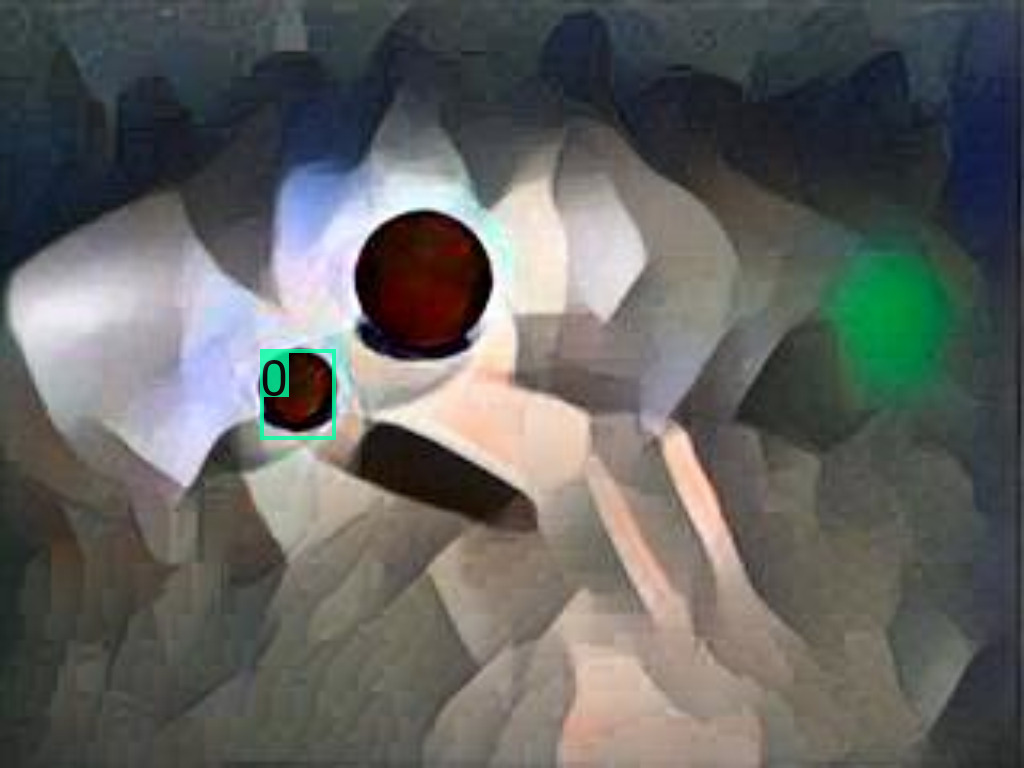} &
      \includegraphics[height=0.18\textwidth]{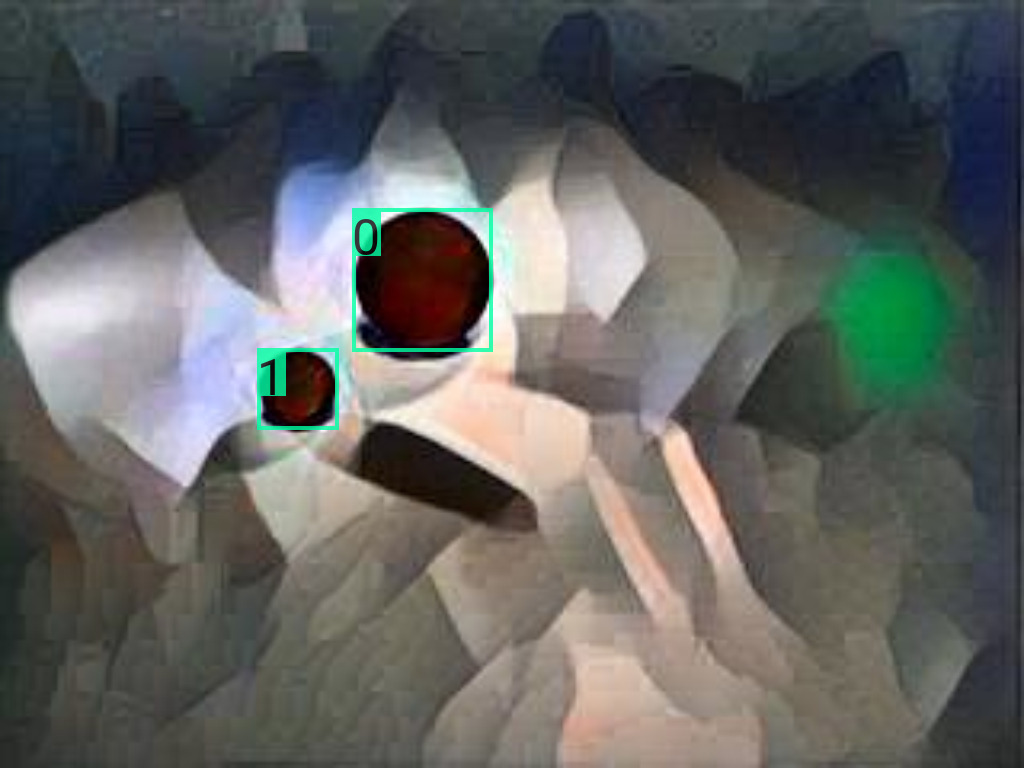} &
      \includegraphics[height=0.10\textwidth]{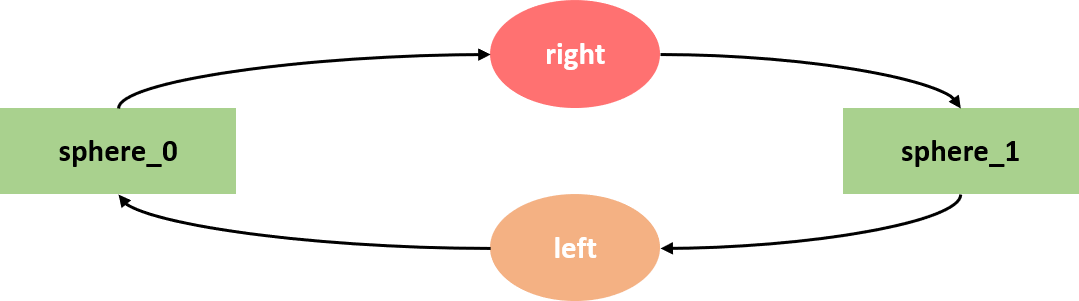} \\
    
    \includegraphics[height=0.18\textwidth]{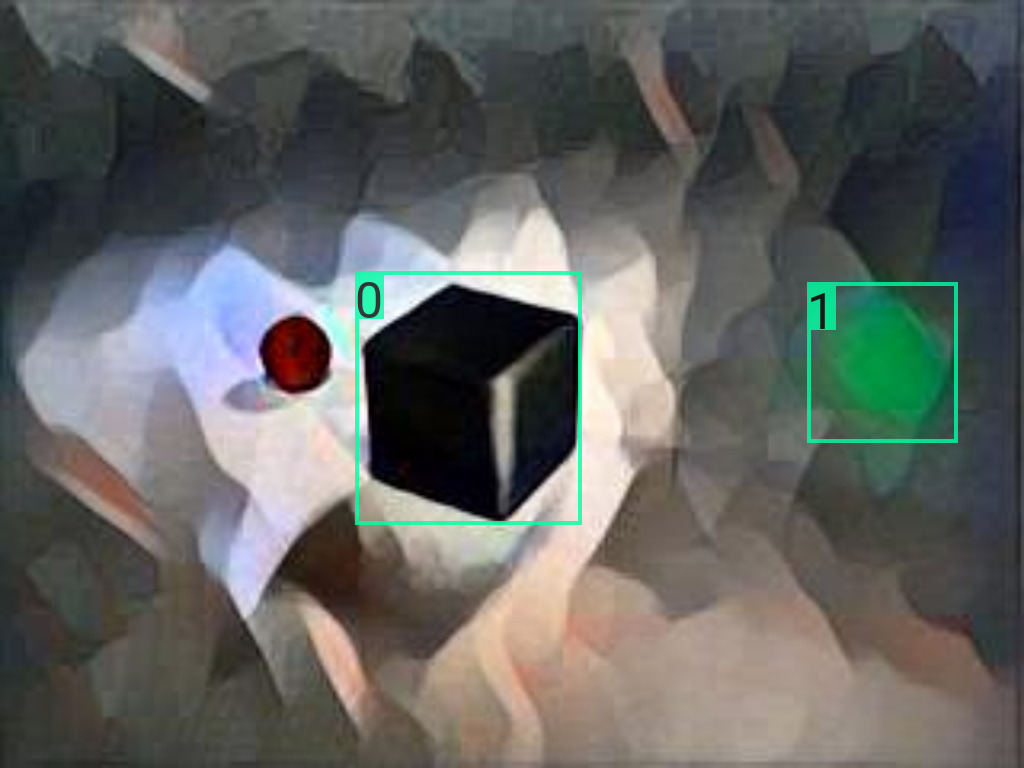} &
      \includegraphics[height=0.18\textwidth]{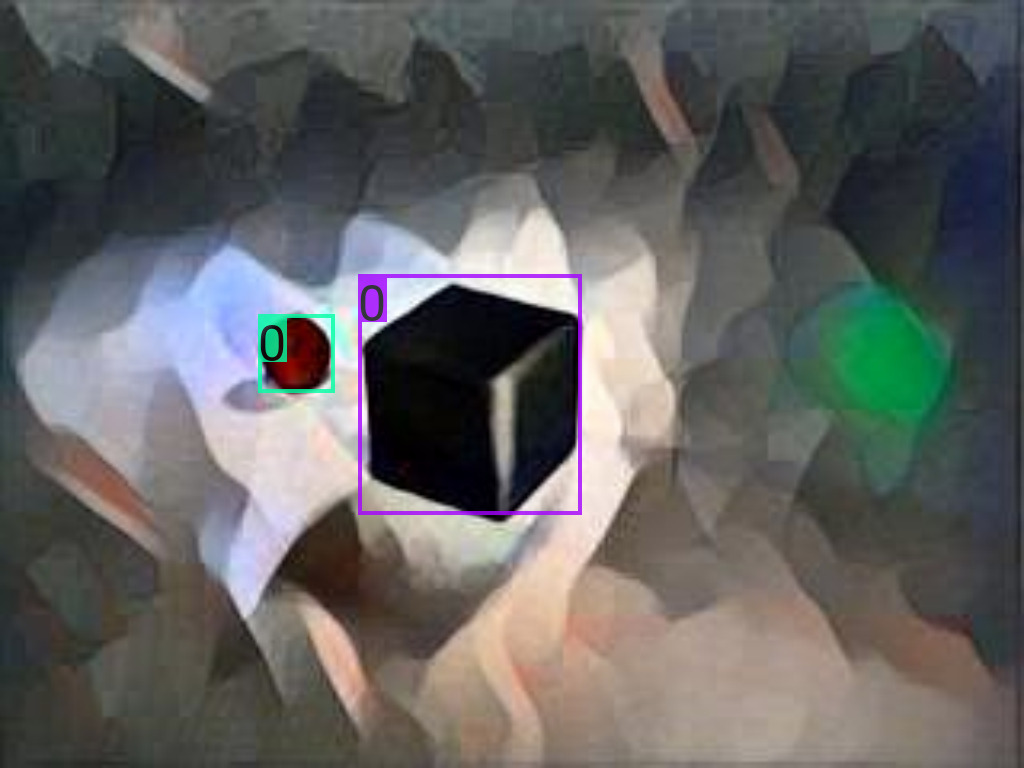} &
      \includegraphics[height=0.10\textwidth]{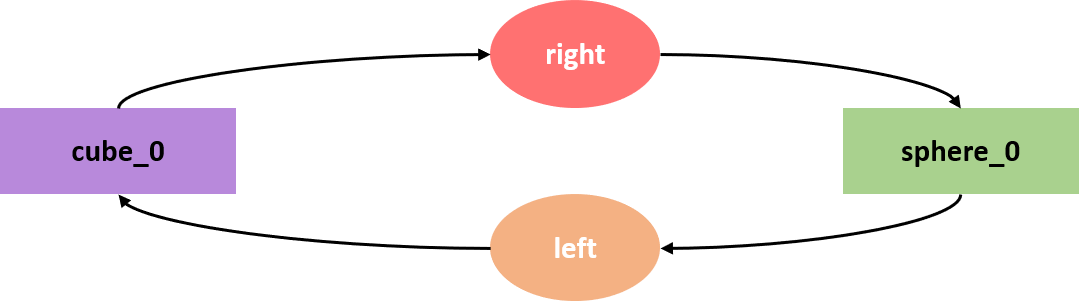} \\
      
    \includegraphics[height=0.18\textwidth]{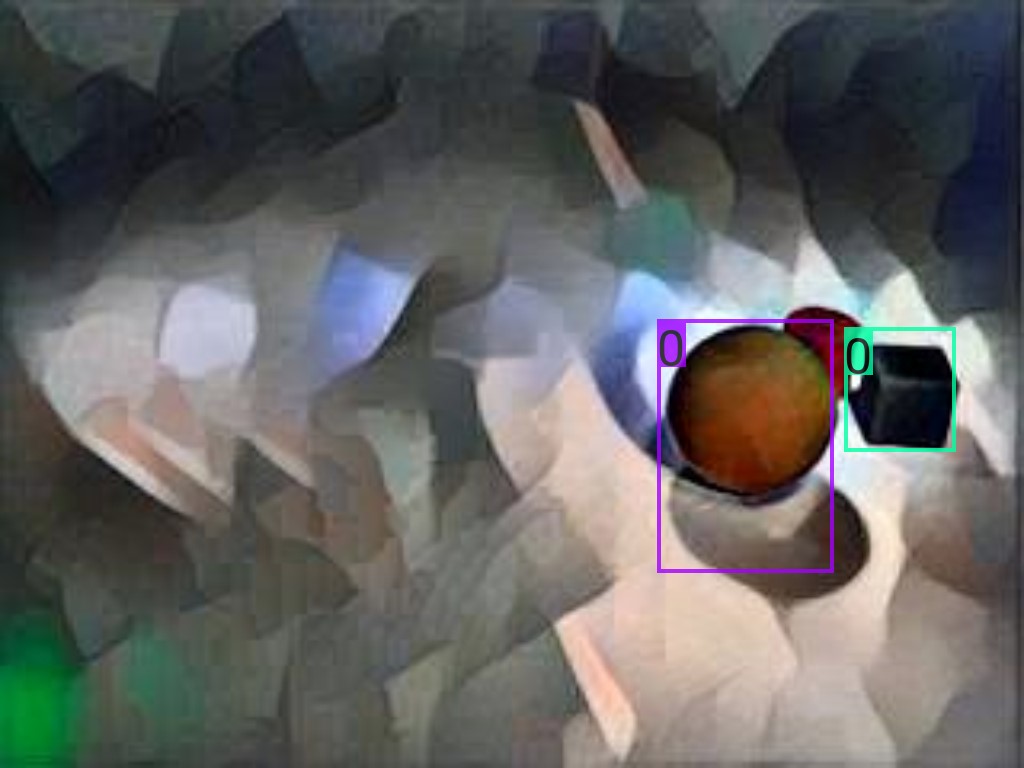} &
      \includegraphics[height=0.18\textwidth]{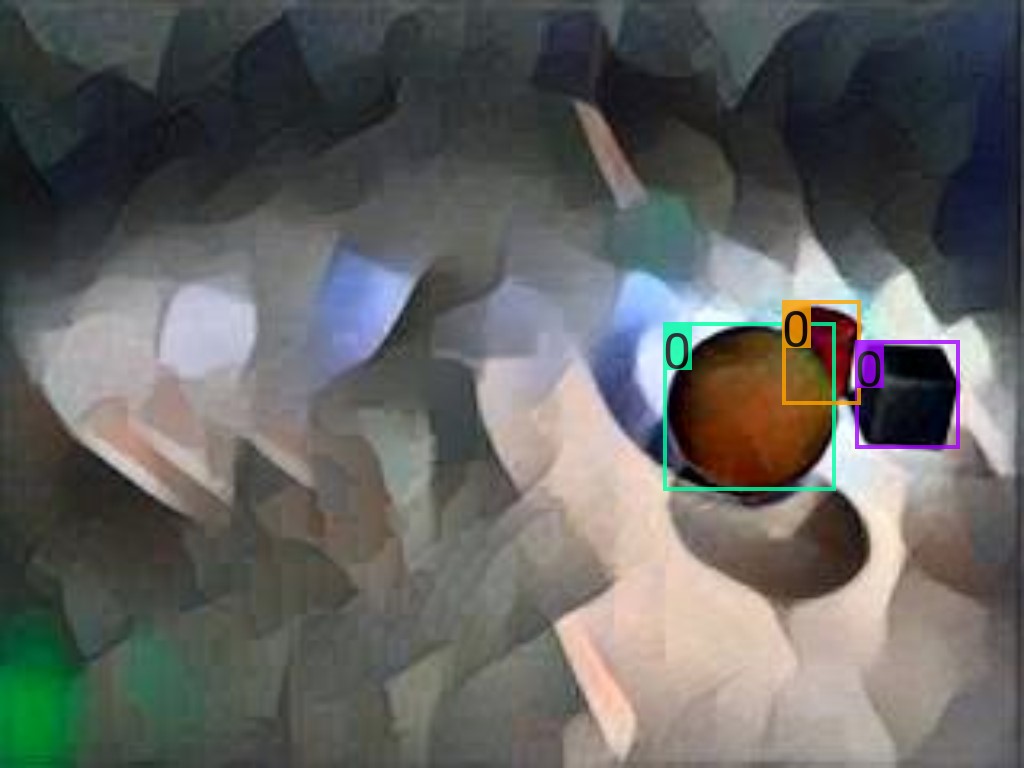} &
      \includegraphics[height=0.18\textwidth]{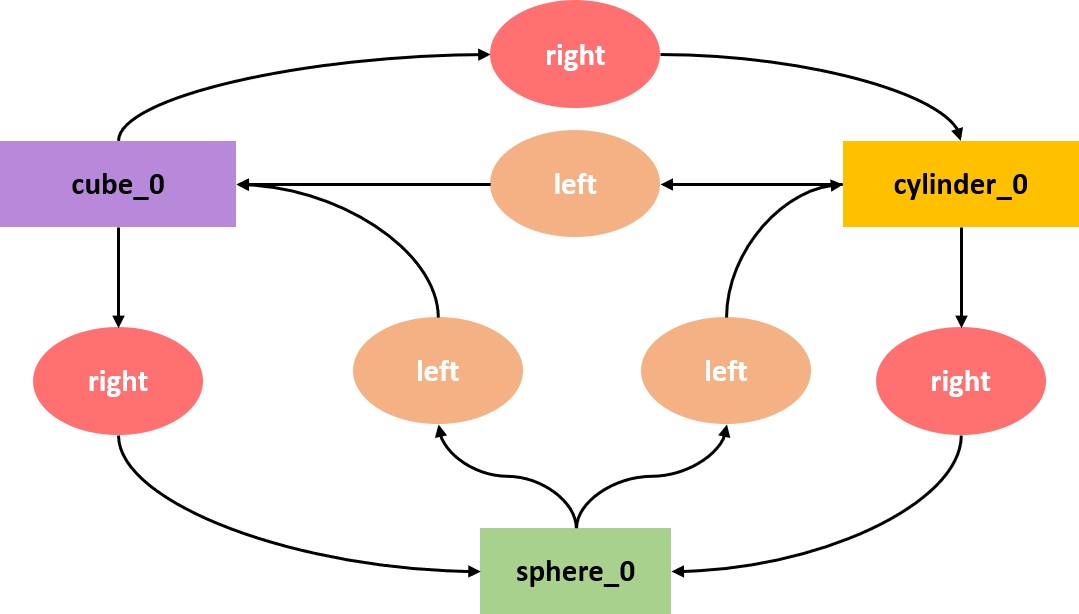} \\
      
    \end{tabular}
    \caption{Qualitative results of Sim2SG on the target domain for CLEVR. First column shows that the SDR~\cite{sdr18} fails to either detect objects or have high number of false positives (mislabels) leading to poor scene graph. Our method detects objects better, has fewer false positives and ultimately generates more accurate scene graphs as shown in second and third column respectively. Objects are color coded.}
    \label{fig:clevr_qualitative}
\end{figure*}

\begin{figure*}
\vspace{-2mm}
    \centering
    \begin{tabular}{cccc}
			 \includegraphics[width=0.21\textwidth]{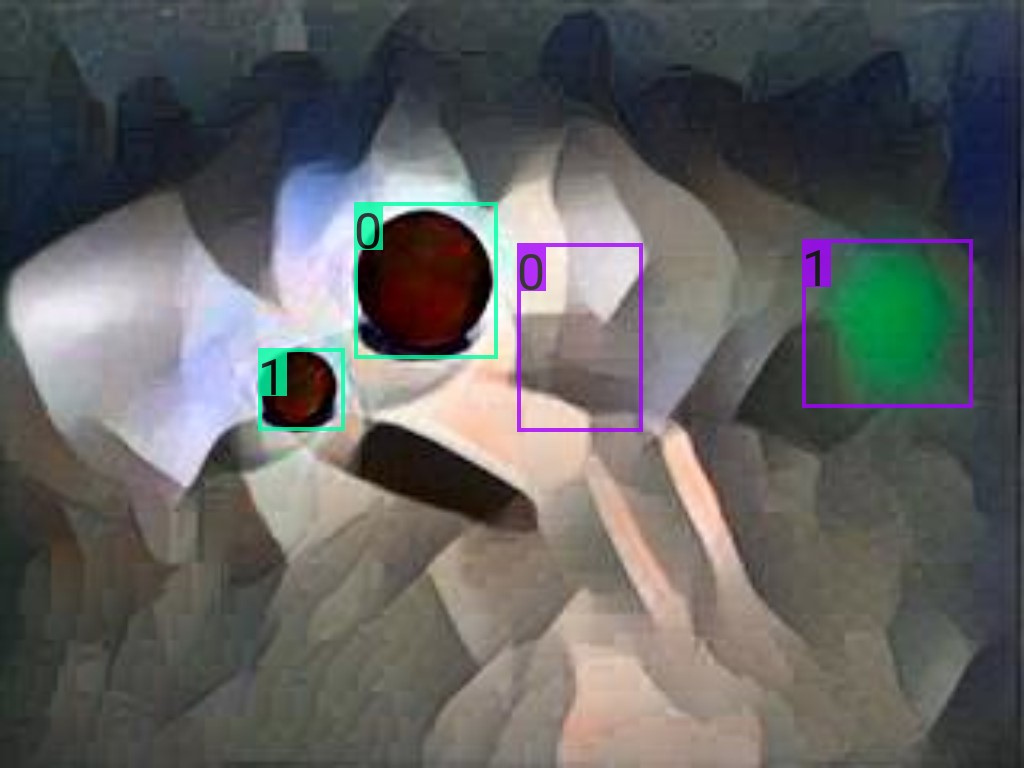} &
			 \includegraphics[width=0.21\textwidth]{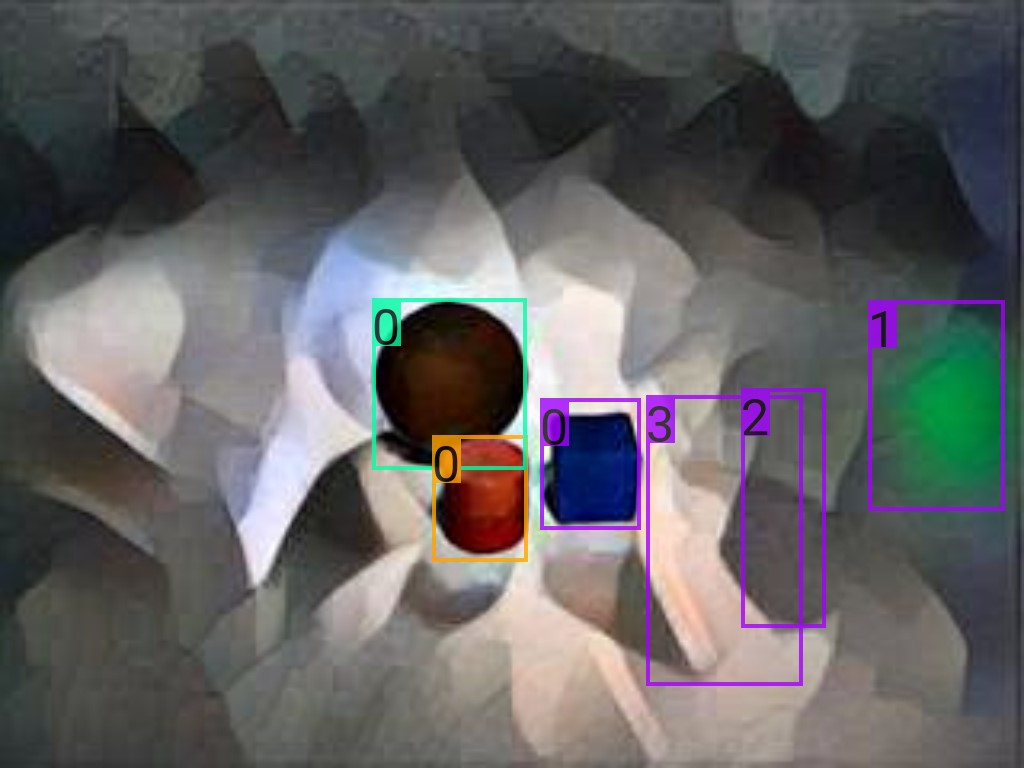} &
			 \includegraphics[width=0.21\textwidth]{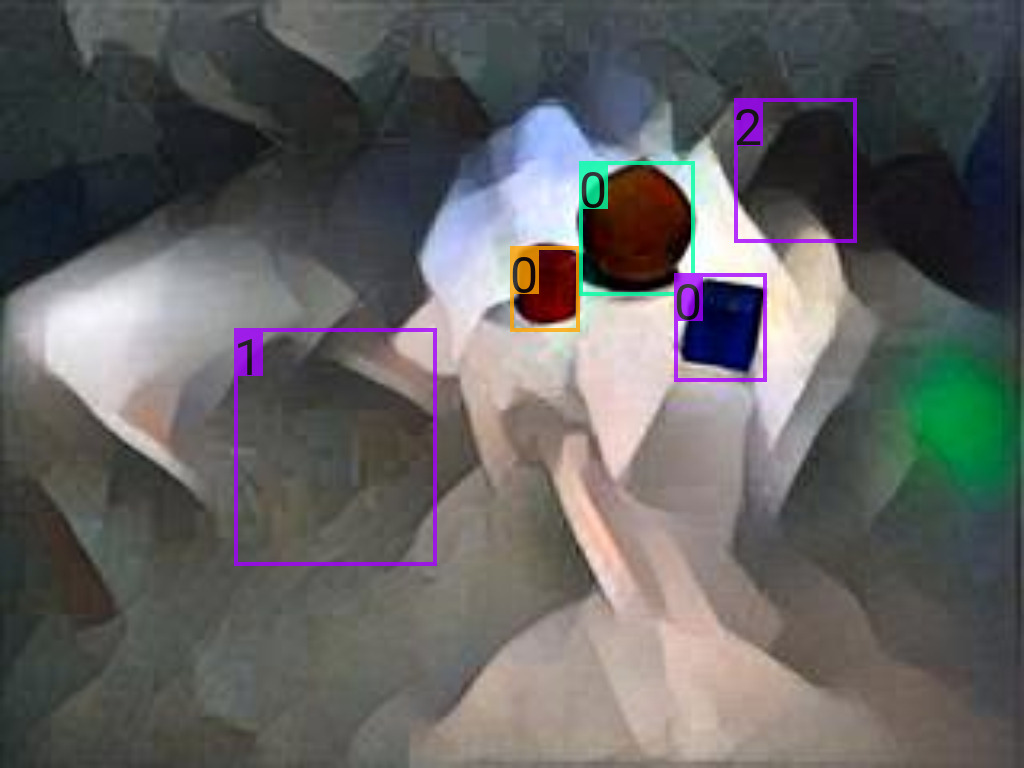} &
			 \includegraphics[width=0.21\textwidth]{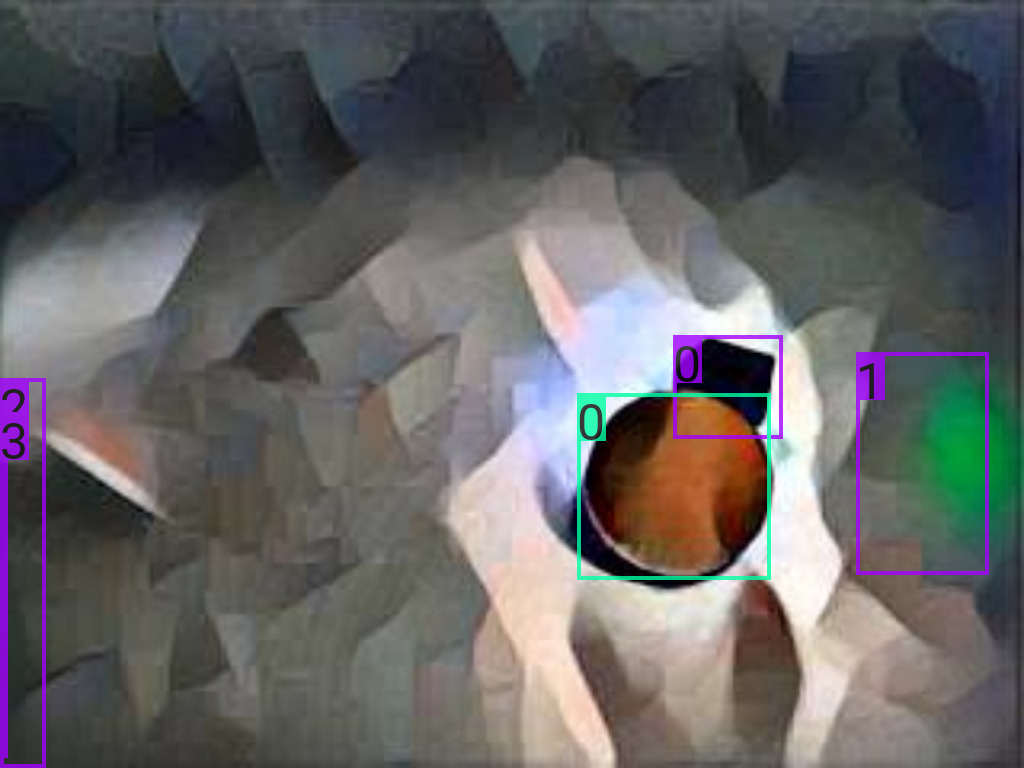} \\  

      \includegraphics[width=0.21\textwidth]{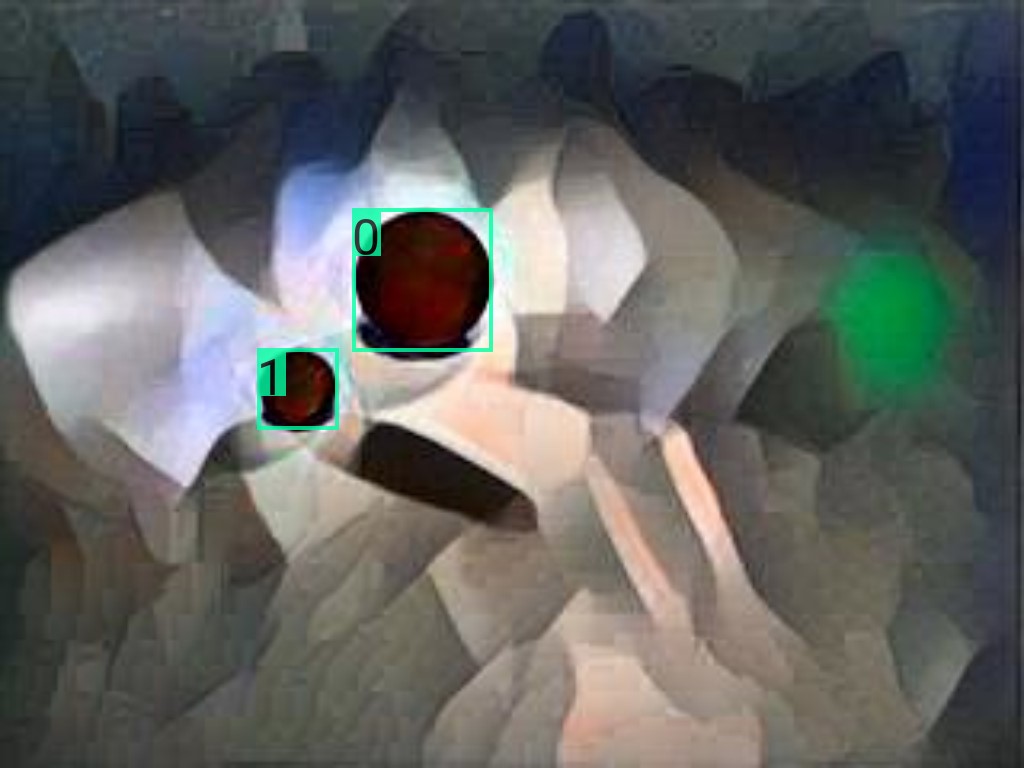} &
			 \includegraphics[width=0.21\textwidth]{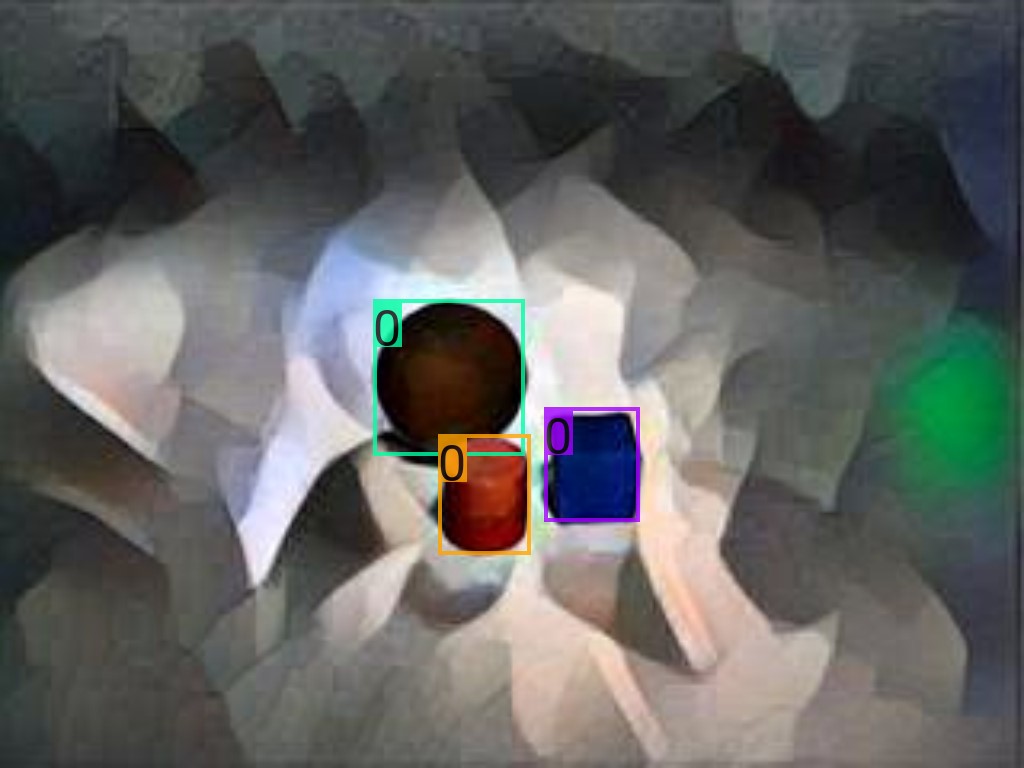} &
			 \includegraphics[width=0.21\textwidth]{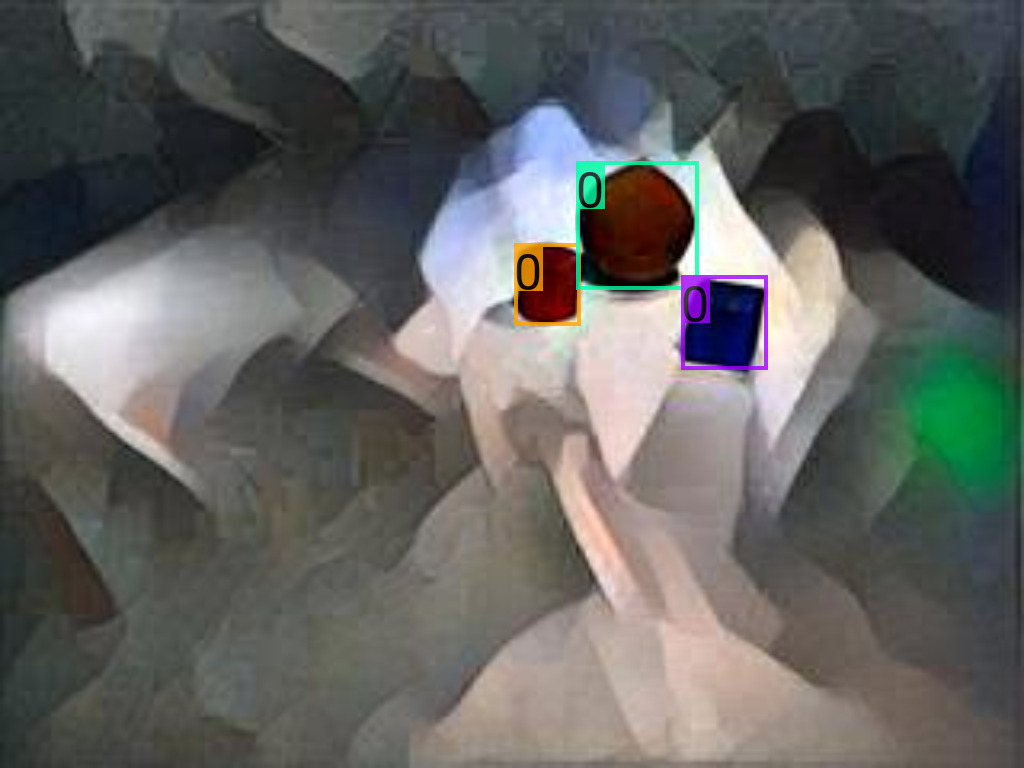} &
			 \includegraphics[width=0.21\textwidth]{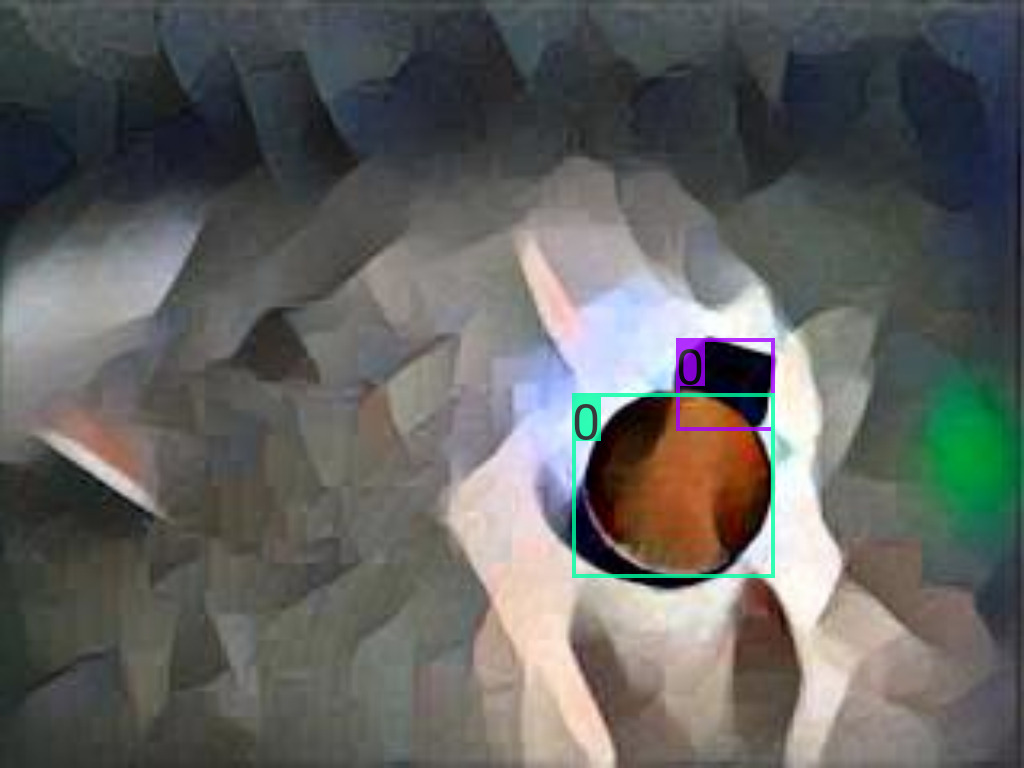} \\ 

    \end{tabular}
    \vspace{-0.5em}
    \caption{ Appearance alignment $\sigma^{a}$ reducing false positive. Top row: $\sigma^{c,label}$,  bottom row: $\sigma^{c,label}$ + $\sigma^{a}$ }
    \label{fig:clevr_false_positive}
\end{figure*}

\paragraph{Results}
More qualitative results of Sim2SG evaluated on the target domain for CLEVR are shown in Figure~\ref{fig:clevr_qualitative}. We see better recall and fewer false positive object detections leading to more accurate scene graphs.
Label alignment $\sigma^{c,label}$ improves object recall, but occasionally introduces some false positive detections. Our appearance alignment $\sigma^a$ helps in reducing such false positives as shown in Figure~\ref{fig:clevr_false_positive}. The full quantitative results of ablation are present in Table~\ref{expclevr2}.

\begin{table*}
\centering
\begin{small}
\begin{tabular}{l c c }
\toprule
  Method  &  mAP@0.5 IoU & Recall@20   \\
 \midrule
  SDR~\cite{sdr18} & 0.675 & 0.339 \\
 Ours ($\sigma^{c,label}$) & 0.923 & 0.646 \\
 Ours ($\sigma^a$)  & \textbf{0.938} & \textbf{0.938} \\
 \bottomrule
\end{tabular}
 \end{small}
\begin{small}
\begin{tabular}{l c c }

\toprule
 Method &  mAP@0.5 IoU & Recall@20   \\
 \midrule
 SDR~\cite{sdr18} & \textbf{1.000} & 0.760 \\
 Ours ($\sigma^a$) & 0.970 & 0.722 \\
 Ours ($\sigma^{c,label})$ & \textbf{1.000} & \textbf{0.996} \\
 \bottomrule
\end{tabular}
 \end{small}
 \caption{ 
\label{expclevr2} Left (resp. right): Source and target domains have different (resp. similar) appearance but similar (resp. different) content distribution. All the evaluations are on the target domain.}
\end{table*}

\subsubsection{Dining-Sim}
\label{appnd:diningsim}
\paragraph{Setup}
The Dining-Sim environment is written using Pixar's USD API\footnote{{\scriptsize\url{https://graphics.pixar.com/usd/docs/index.html}}} and rendered with a proprietary renderer.
The source domain is rendered with 2 spp (samples per pixels) followed by denoiser. We select 1 chair (cantilever chair), 1 table (workshop table) and 1 laptop (PC). We randomly place chair and table on the floor and laptop on the floor as well as on the table with a random orientation.
The asset for each subcategory is randomly chosen from a list of subcategory specific ShapeNet~\cite{chang2015shapenet} assets.
We also ensure that objects do not overlap by applying collision avoidance with simple box collision volumes. 
A subset of 4 to 5 simple materials that vary only in diffuse colour is created for each of the walls, floor, chair and table. Laptops use the original asset texture. 

The target domain is rendered  using path tracing with 20 spp (samples per pixels) followed by denoiser. We use 4 chairs (Windsor chair), 1 table (kitchen table) and 2 laptops (MacBook).  We first place the table with a random orientation and position on the floor.
We then place the four chairs at each side of the table, oriented towards the table centre. Two laptops are then placed randomly on the table surface with a random rotation.
The asset for each subcategory is randomly chosen from a list of subcategory specific ShapeNet~\cite{chang2015shapenet} assets.
For materials, we use a subset of 4 to 6 physically based, highly detailed materials for each of the walls, floor, chair and table. As with the source domain, laptops use the original asset texture. 

\begin{figure*}
    \centering
    \addtolength{\tabcolsep}{-4.6pt}
    \begin{tabular}{ccccc}
				\raisebox{5ex}{\rotatebox[origin=c]{90}{ Source}}  &
			 \includegraphics[width=0.23\textwidth]{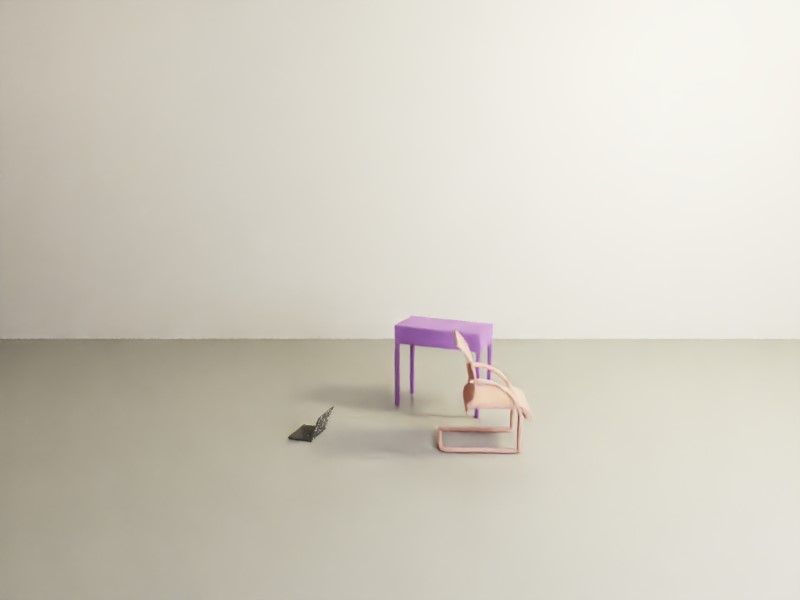} &
			 \includegraphics[width=0.23\textwidth]{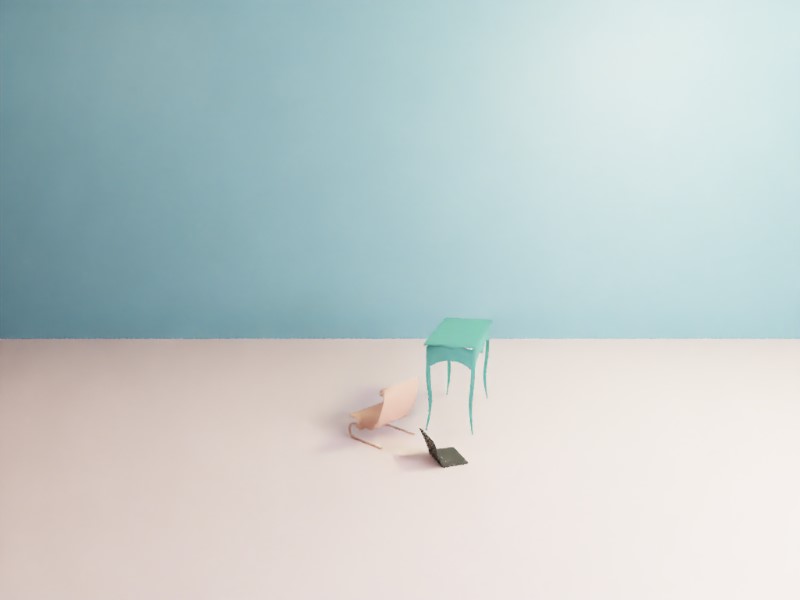} &
      \includegraphics[width=0.23\textwidth]{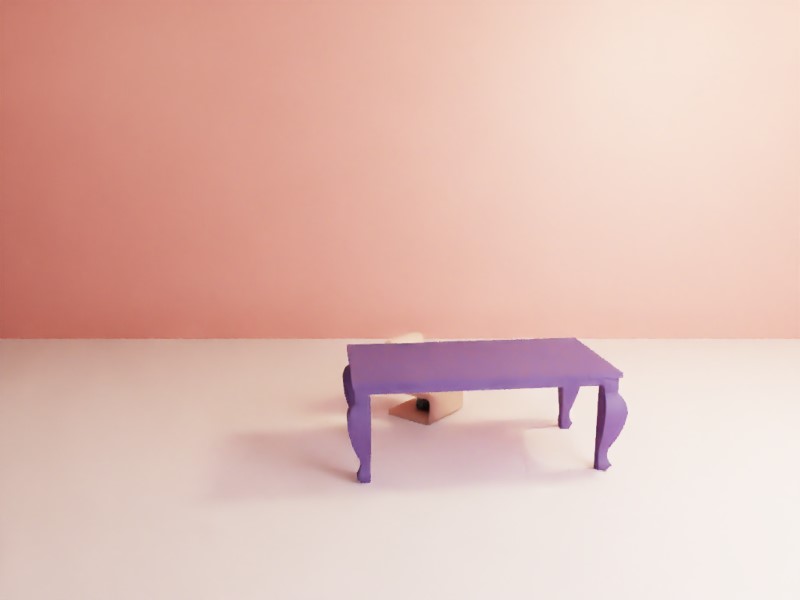} &
      \includegraphics[width=0.23\textwidth]{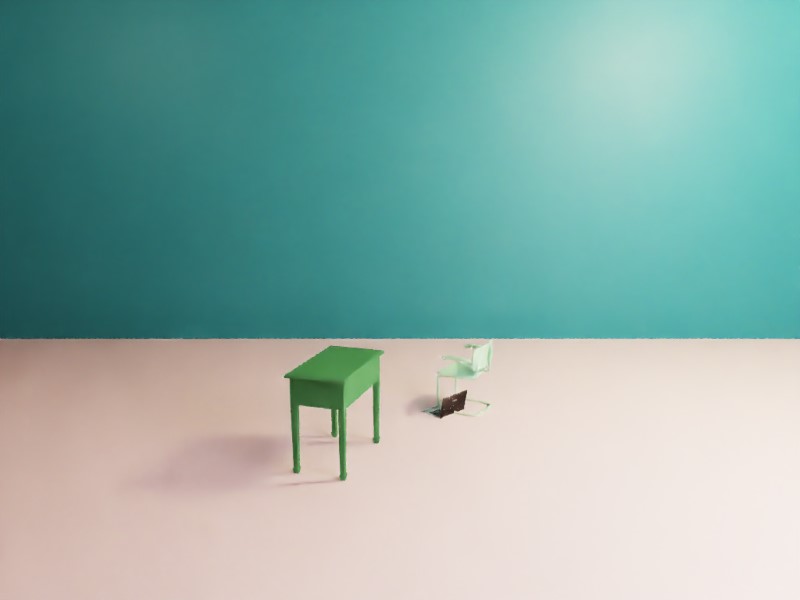} \\
			
			 \raisebox{5ex}{\rotatebox[origin=c]{90}{ Target}} &
      \includegraphics[width=0.23\textwidth]{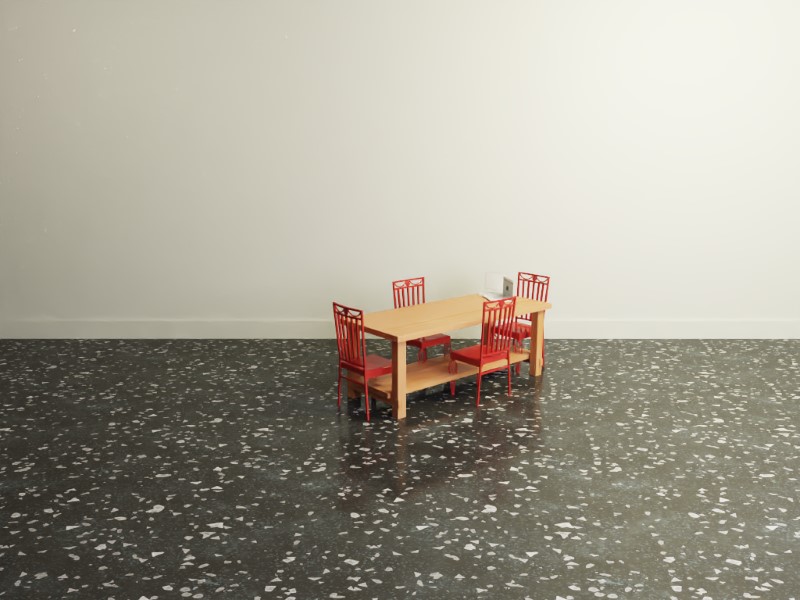} &
			 \includegraphics[width=0.23\textwidth]{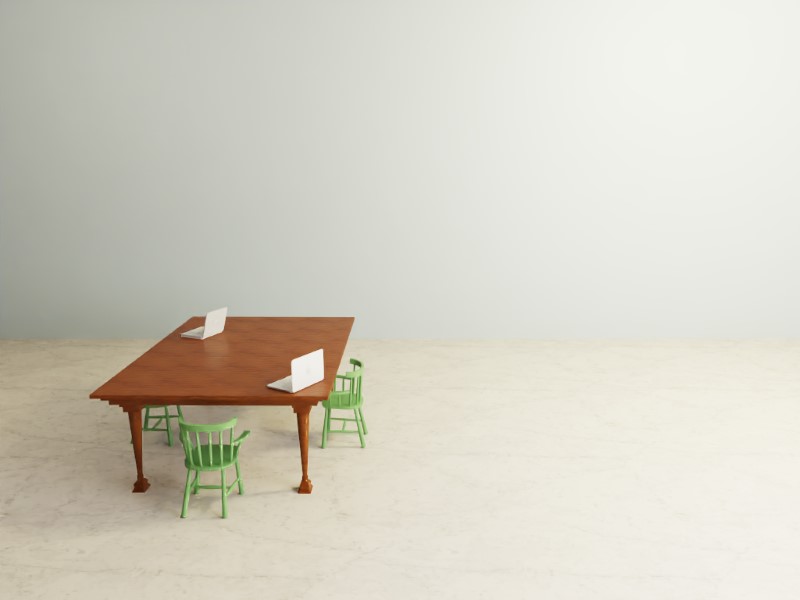} &
      \includegraphics[width=0.23\textwidth]{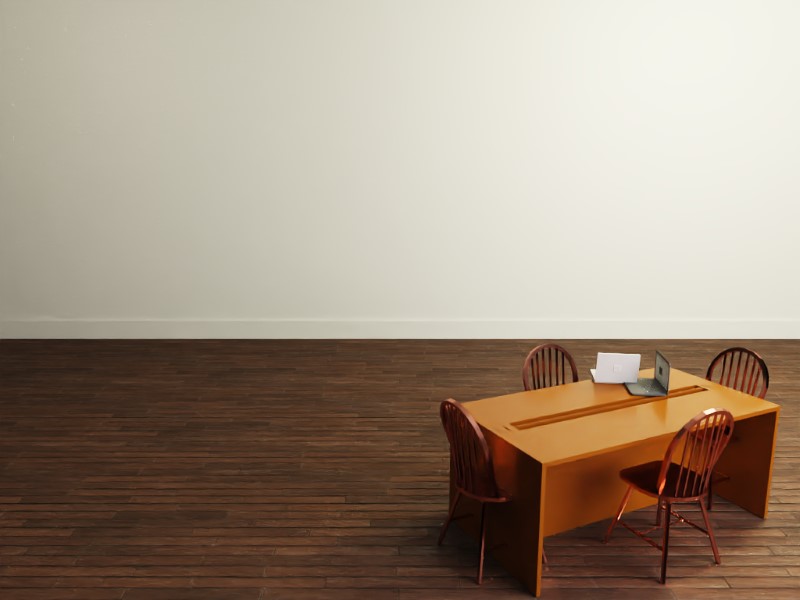} &
      \includegraphics[width=0.23\textwidth]{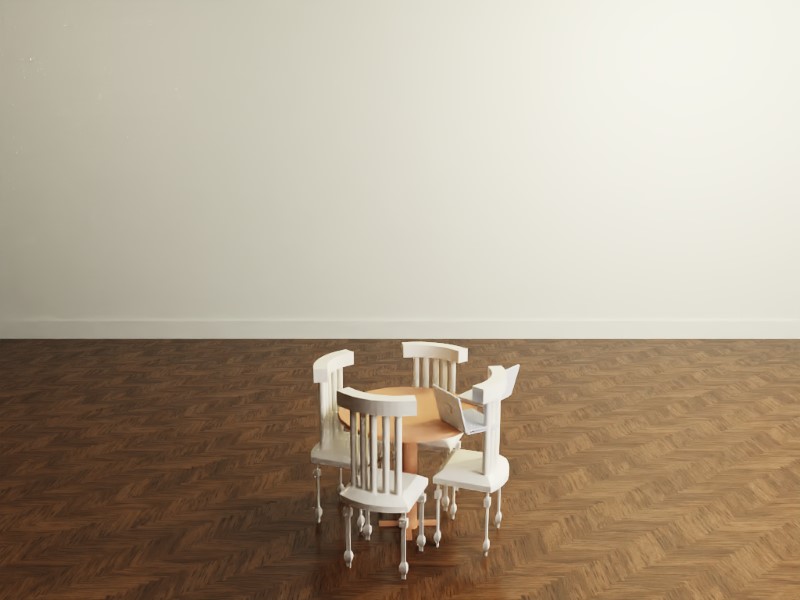} \\
      
      		\raisebox{5ex}{\rotatebox[origin=c]{90}{ Source}}  &
			 \includegraphics[width=0.23\textwidth]{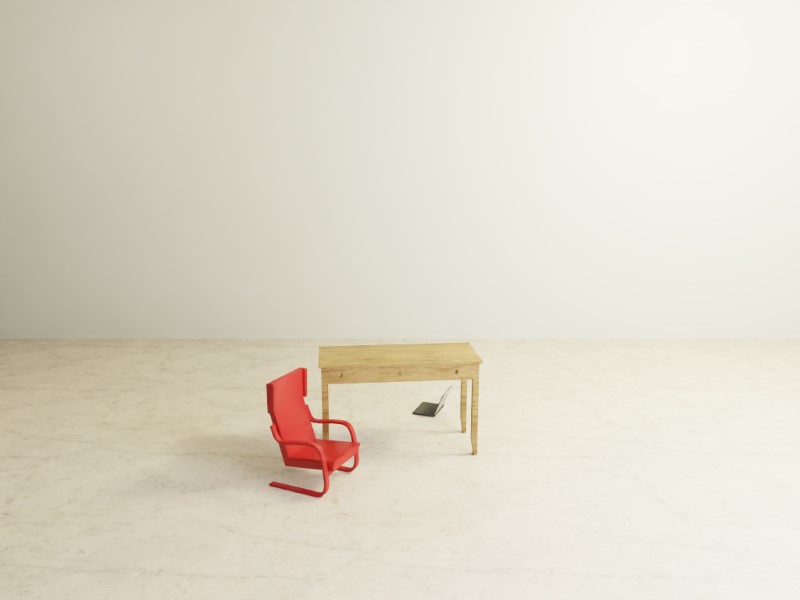} &
			 \includegraphics[width=0.23\textwidth]{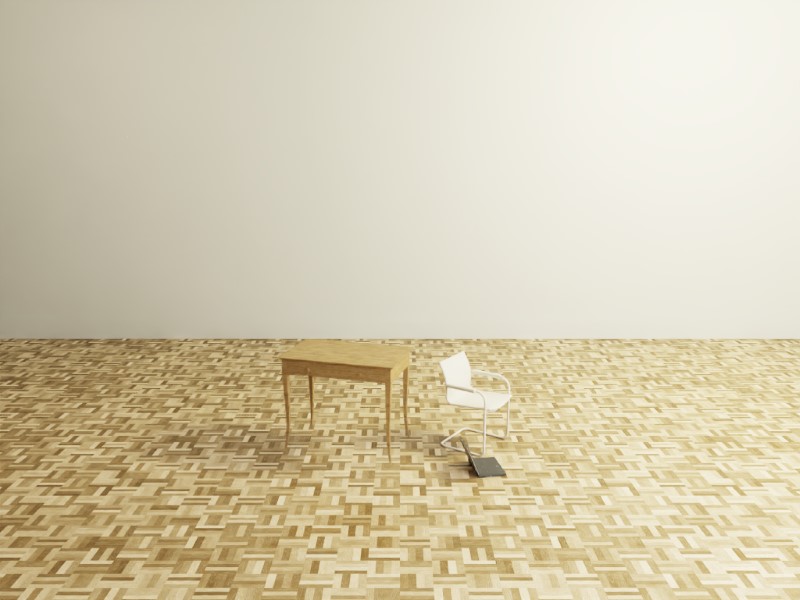} &
      \includegraphics[width=0.23\textwidth]{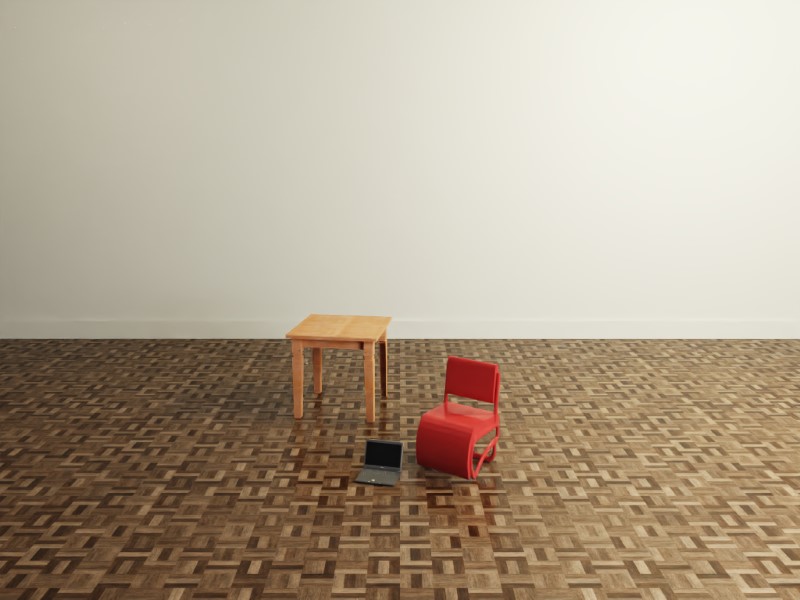} &
      \includegraphics[width=0.23\textwidth]{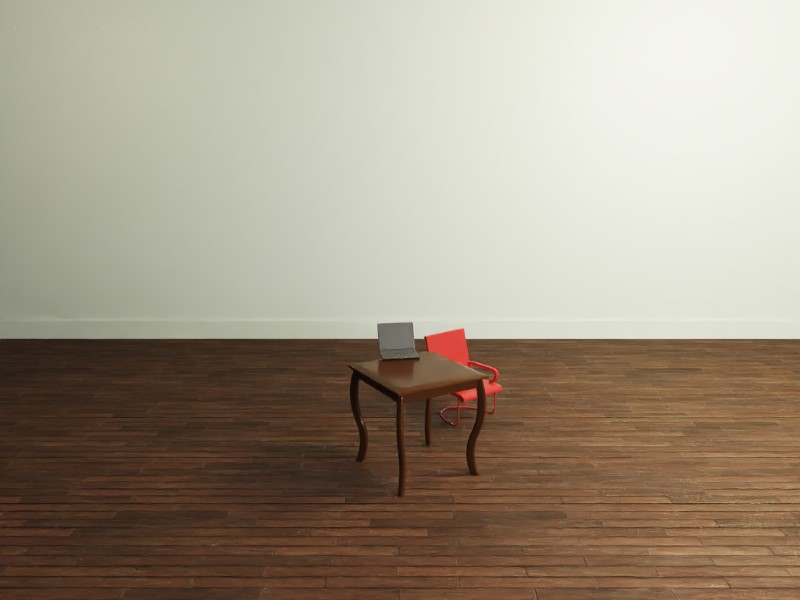} \\
			
			 \raisebox{5ex}{\rotatebox[origin=c]{90}{ Target}} &
      \includegraphics[width=0.23\textwidth]{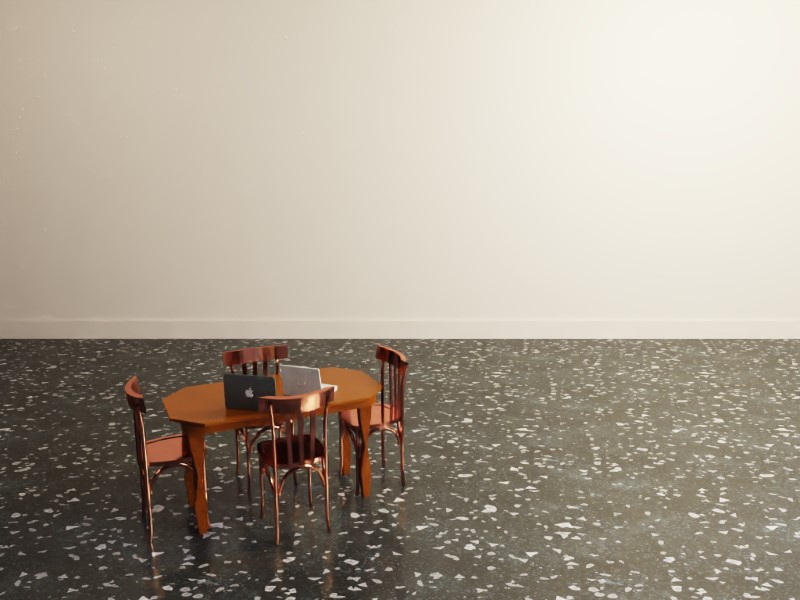} &
			 \includegraphics[width=0.23\textwidth]{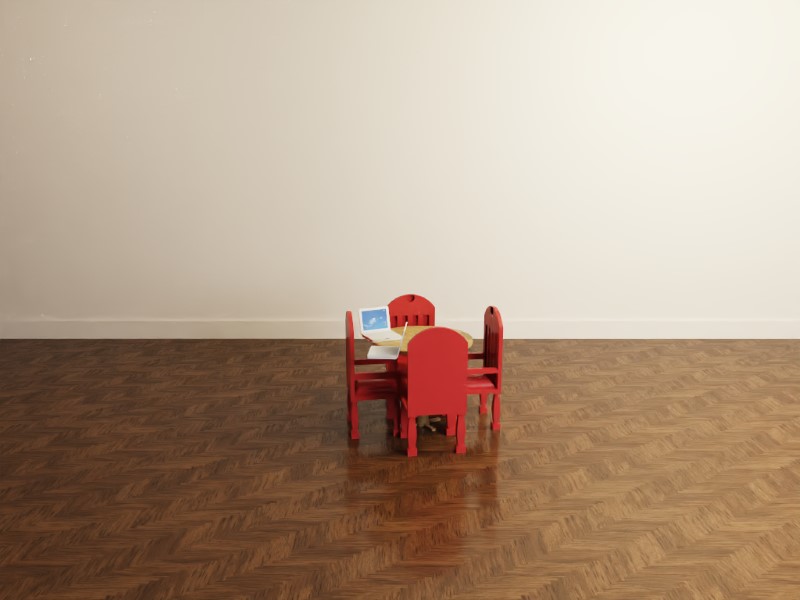} &
      \includegraphics[width=0.23\textwidth]{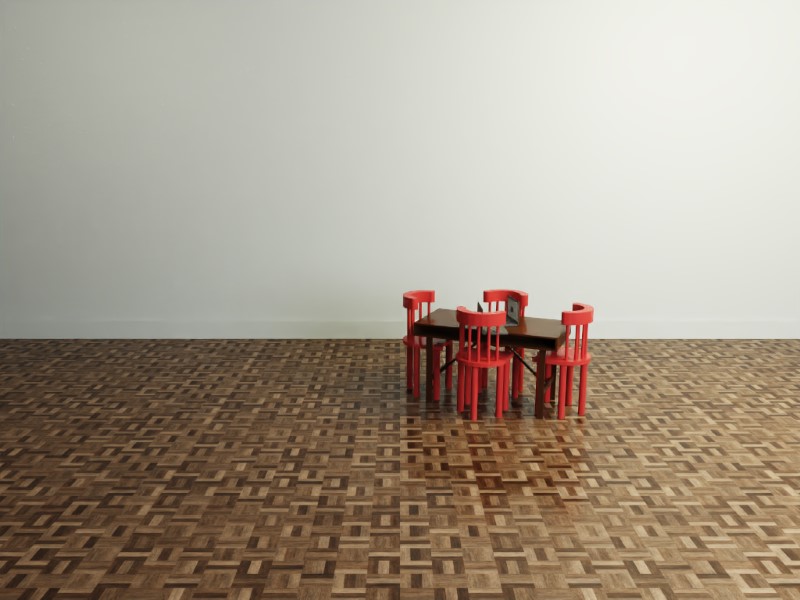} &
      \includegraphics[width=0.23\textwidth]{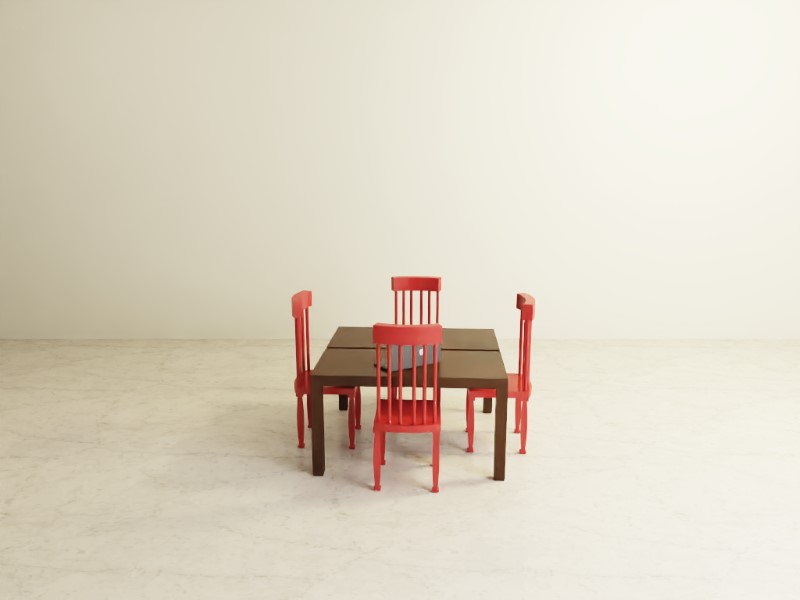} \\
      
      \raisebox{5ex}{\rotatebox[origin=c]{90}{ Source}}  &
			 \includegraphics[width=0.23\textwidth]{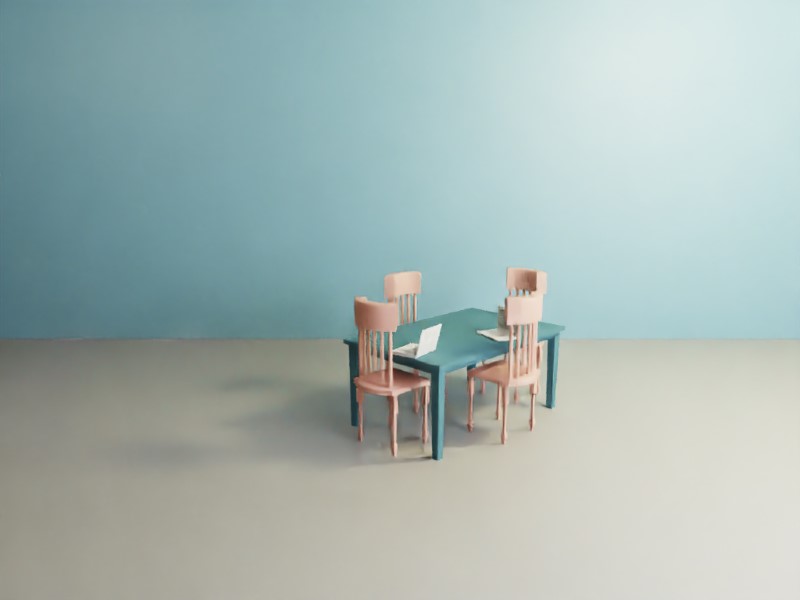} &
			 \includegraphics[width=0.23\textwidth]{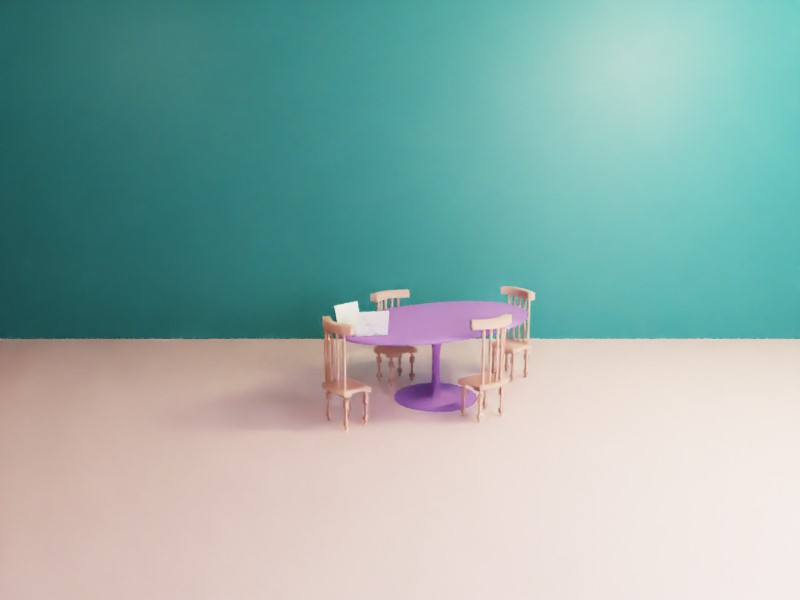} &
      \includegraphics[width=0.23\textwidth]{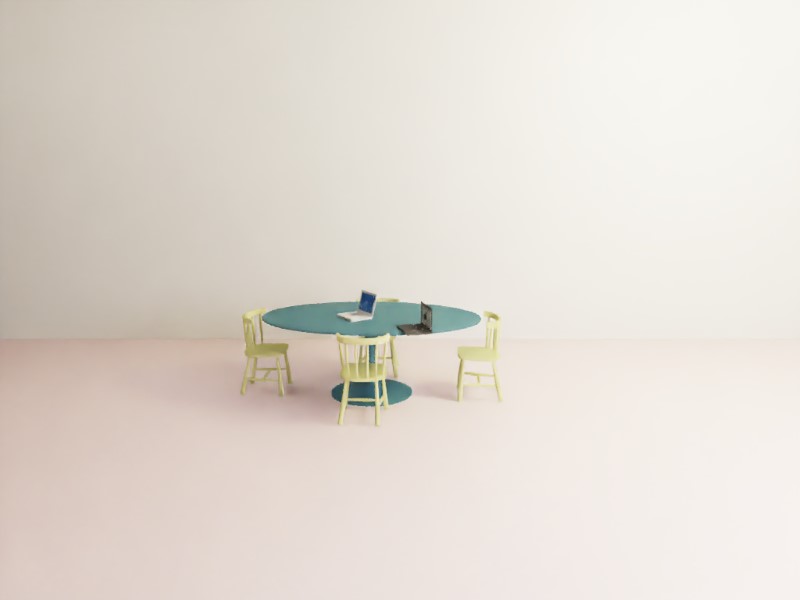} &
      \includegraphics[width=0.23\textwidth]{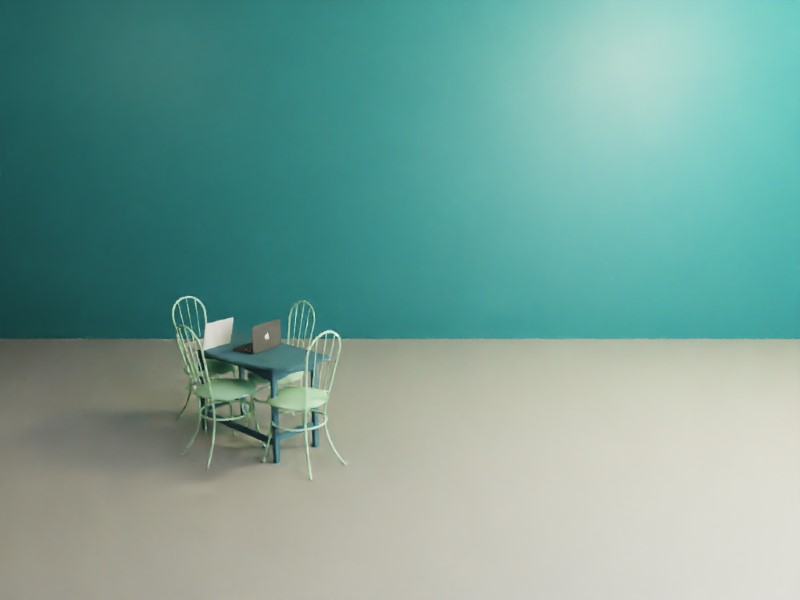} \\
			
			 \raisebox{5ex}{\rotatebox[origin=c]{90}{ Target}} &
      \includegraphics[width=0.23\textwidth]{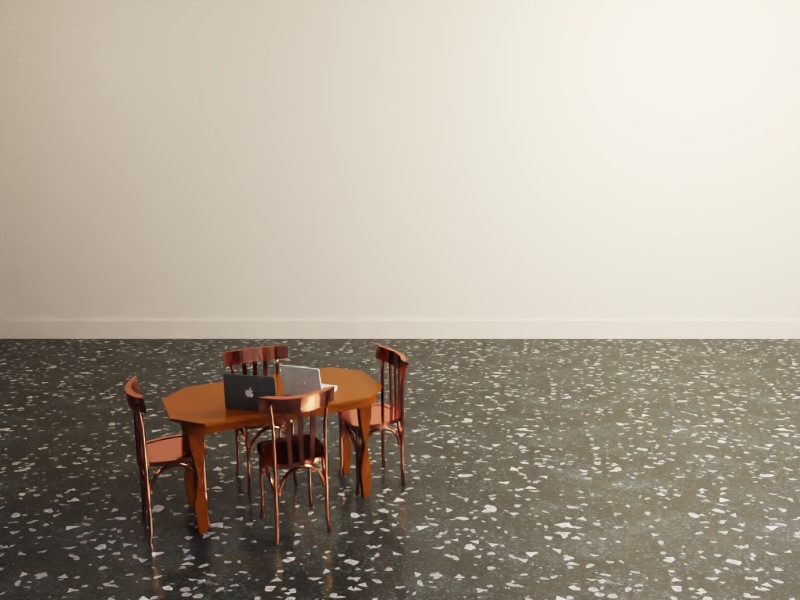} &
			 \includegraphics[width=0.23\textwidth]{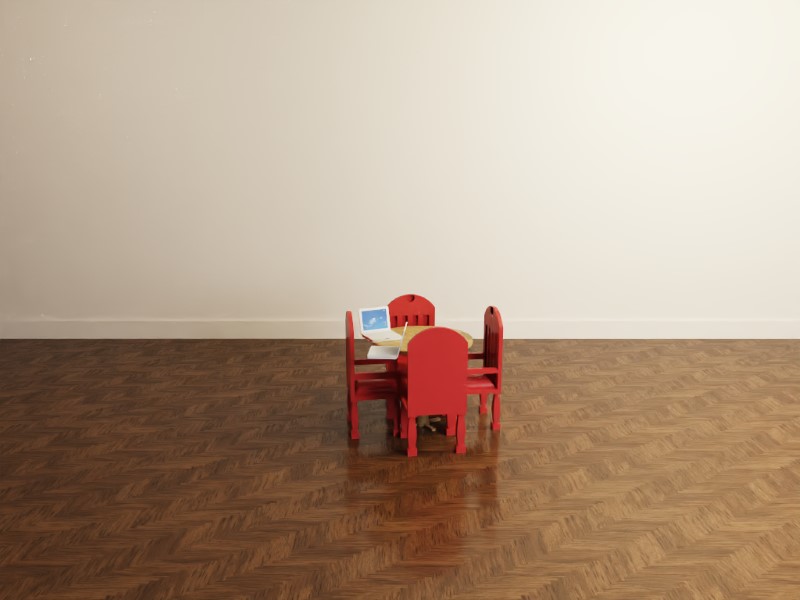} &
      \includegraphics[width=0.23\textwidth]{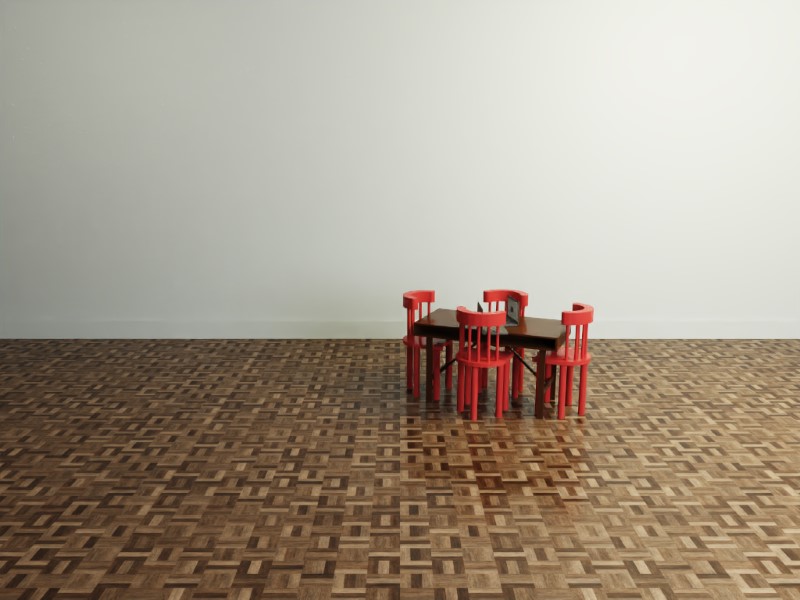} &
      \includegraphics[width=0.23\textwidth]{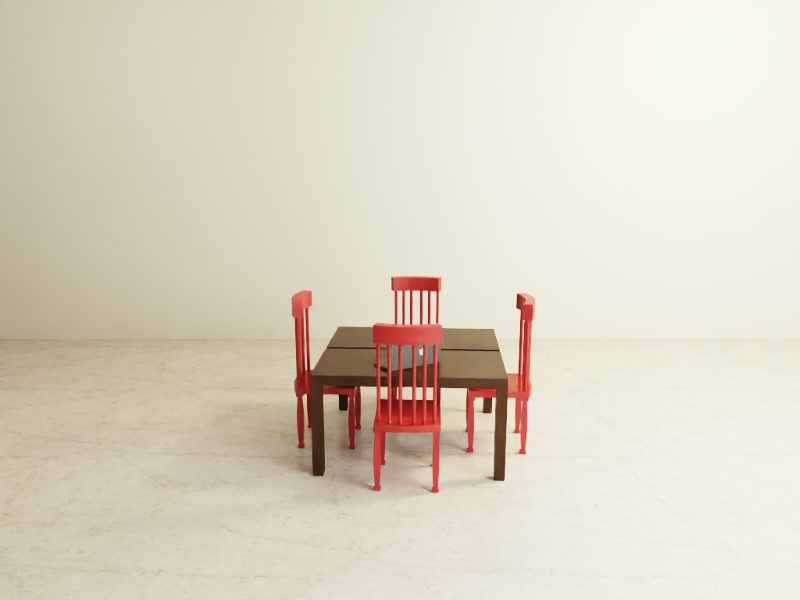} \\
     
    \end{tabular}
    \caption{Samples from source and target distributions for Dining-Sim. Row 1-2: Source and Target domains differ in both appearance and content. Row 3-4: Source and Target differ in content but have same appearance. Row 5-6: Source and Target differ in appearance but have same content.
    }
    \label{fig:diningsim}
\end{figure*}

Both domains share room parameters: a fixed camera (60 degree field of view, positioned at far side of the room)
and 3 fixed spherical lights.
 Samples from the source and the target domains are shown in Figure~\ref{fig:diningsim}.
There are five kinds of relationships - front, behind, left, right, and on with table as subject.
We use 5000 labeled images from source, 5000 unlabeled images from target for training and 1000 labeled images from both source \& target domains for evaluation. We use 1024 x 768 image resolution for training and evaluation.

\paragraph{Details of Synthesis Step}
We describe how we generate synthetic data by inferring scene graphs from target domain in detail.
We filter the objects and relationships among them using an adaptive threshold (details in the next paragraph) for the generation. 
We assume access to camera parameters (intrinsic and extrinsic both). 
We project a ray from the camera through the pixel corresponding to the bounding box bottom-centre. The 3D coordinate of the object is then the intersection of the ray and the ground plane, which we assume to be flat at elevation 0.
We place each object in the 3D scene by picking a random 3D asset according its type (class) and assigning random pose in the range $0^{\circ}$--$360^{\circ}$. 
We assume context like ground, wall as described in the previous paragraph. We refine the 3D scene further according to the predicted relationships among objects.
For example, we use ``on'' relationship to refine object placements by
adjusting the object (laptop or chair) elevation to match the table top.
We then render the 3D scene.


\paragraph{Training Details}
We optimize the model using a SGD optimizer with learning rate of $10^{-4}$ and momentum of 0.9. We train our model using a batch size 2 on NVIDIA DGX workstations. We report saturation peak performance in all our tables.
We give equal weights to source task loss $\sigma_s$, appearance alignment $\sigma^a$, prediction alignment $\sigma^{c,pred}$ and label alignment $\sigma^{c,label}$.  


We first train the model with label alignment $\sigma^{c,label}$) for 6 epochs each with $10^{4}$ iterations and score threshold of 0.5.
Then we add appearance alignment $\sigma^a$ and prediction alignment $\sigma^{c,pred}$ and train for an additional $2 \times 10^4 $ iterations.
It takes 12 hours for full training including rendering time.

\begin{figure*}
\addtolength{\tabcolsep}{-4.6pt}
    \centering
    \begin{tabular}{ccc}
    \includegraphics[height=0.17\textwidth]{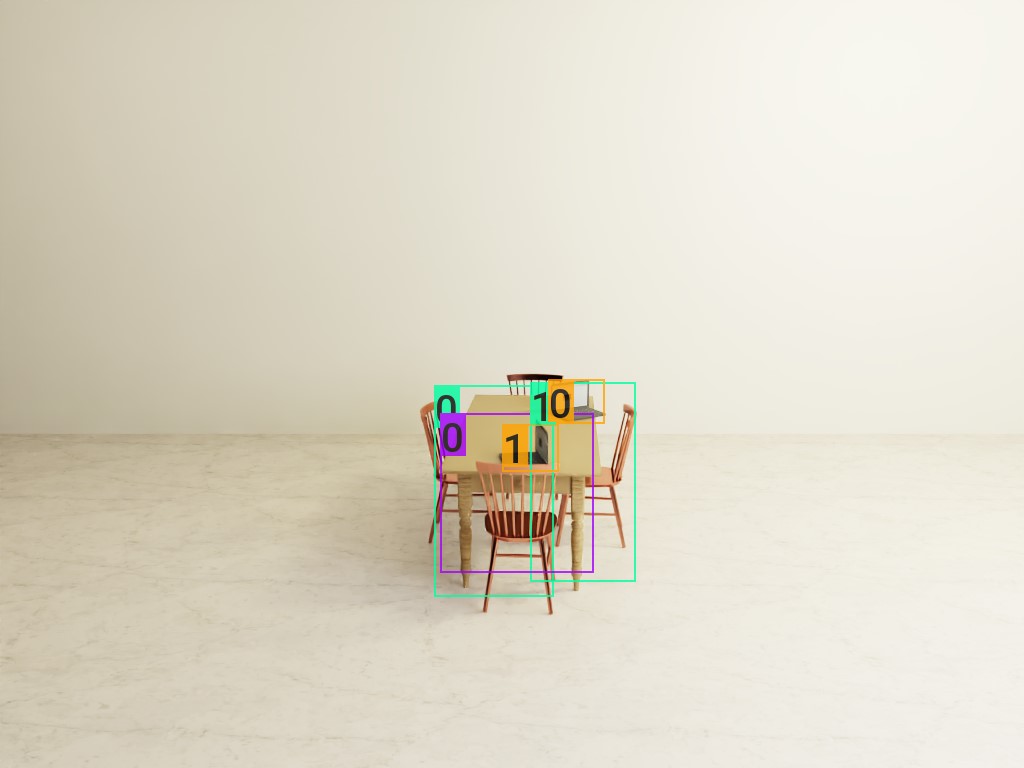} &
      \includegraphics[height=0.17\textwidth]{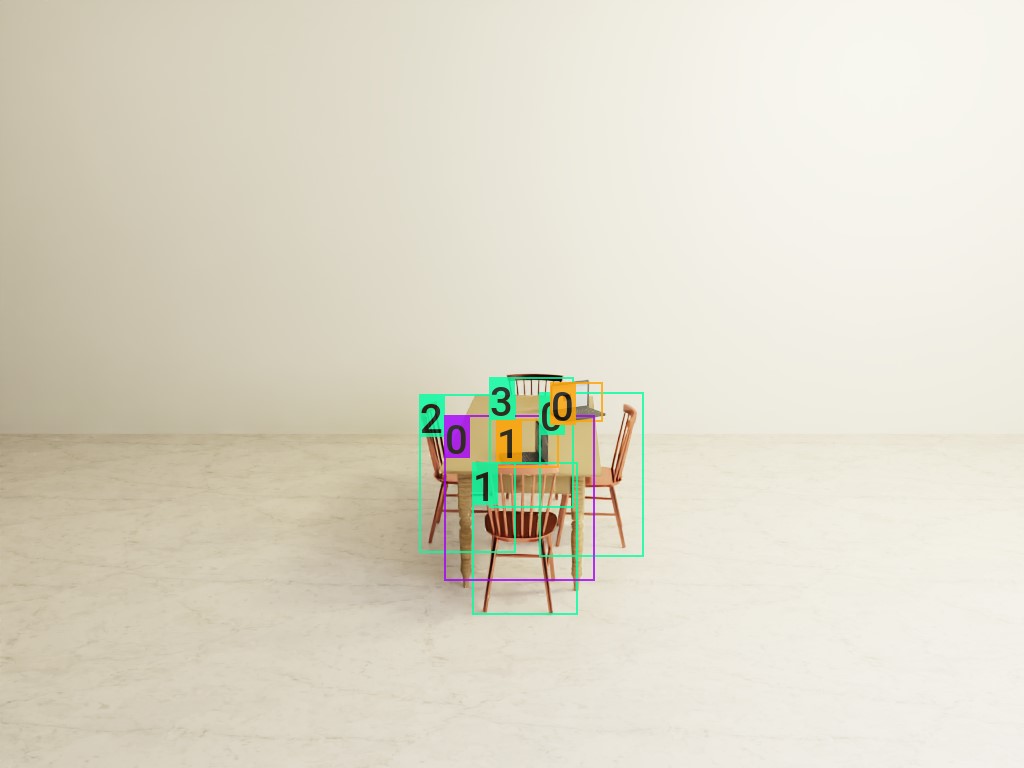} &
      \includegraphics[height=0.17\textwidth]{figures/results/diningsim/label_style_pred/dinning_hor_1.jpg} \\
     \includegraphics[height=0.17\textwidth]{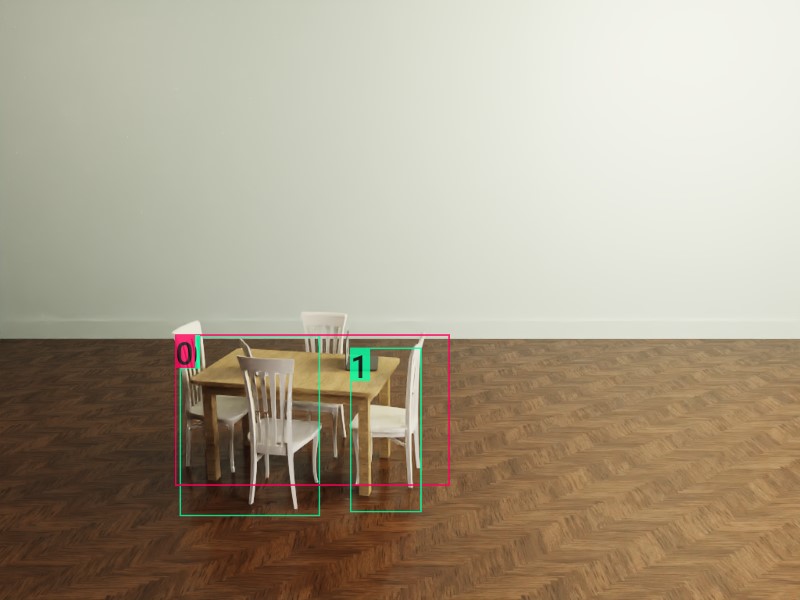} &
      \includegraphics[height=0.17\textwidth]{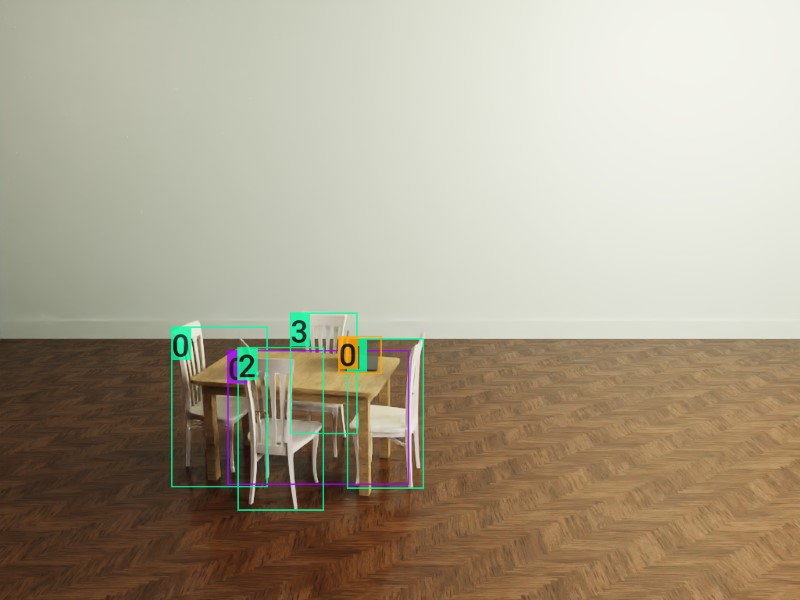} &
      \includegraphics[height=0.17\textwidth]{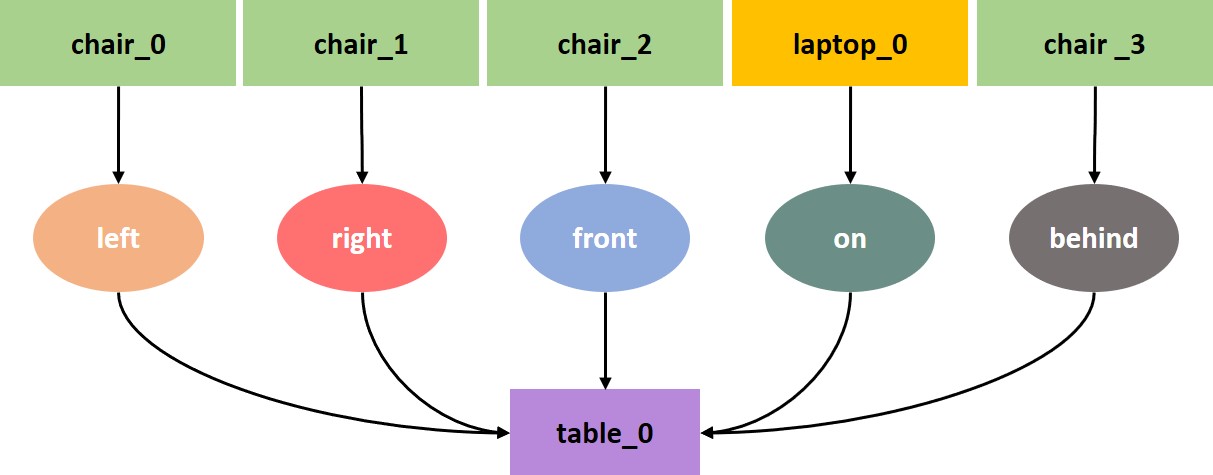} \\
     \includegraphics[height=0.17\textwidth]{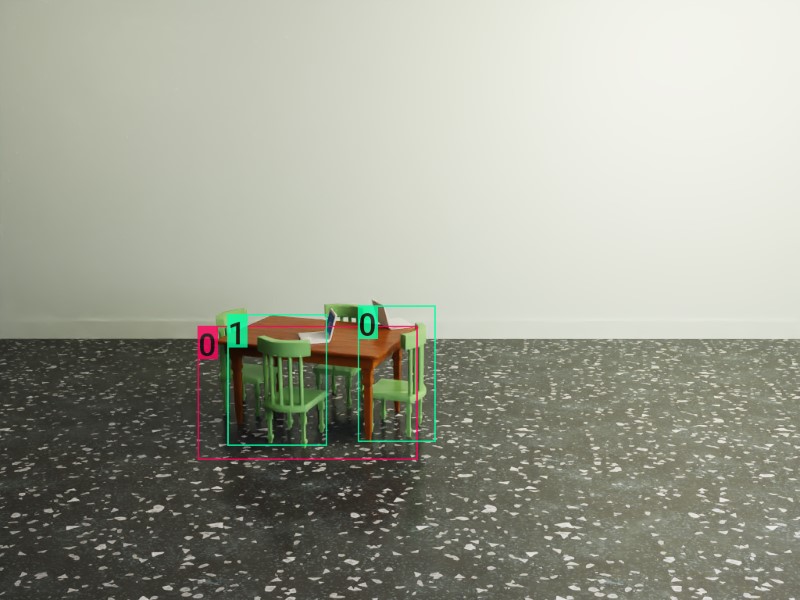} &
      \includegraphics[height=0.17\textwidth]{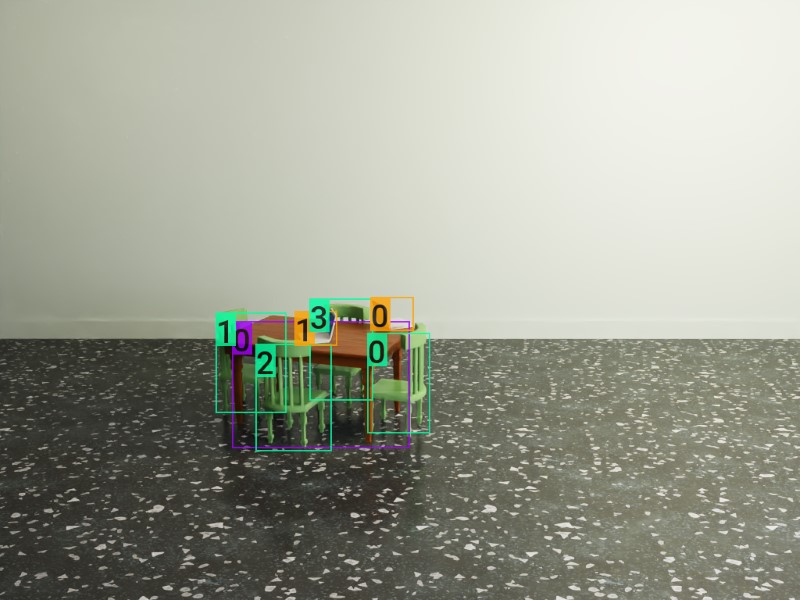} &
      \includegraphics[height=0.17\textwidth]{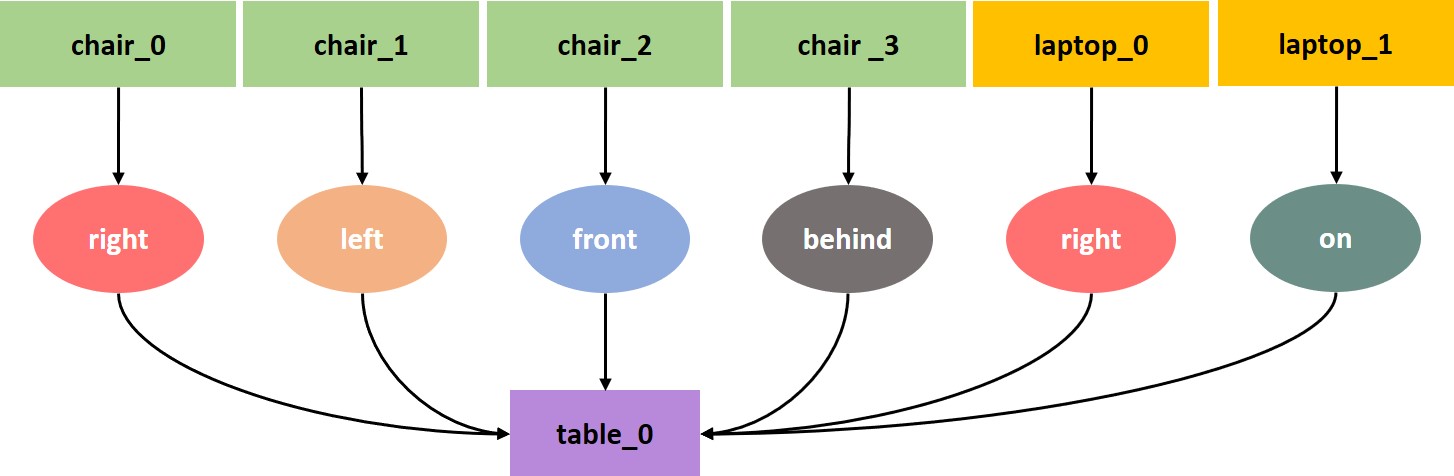} \\
     \includegraphics[height=0.17\textwidth]{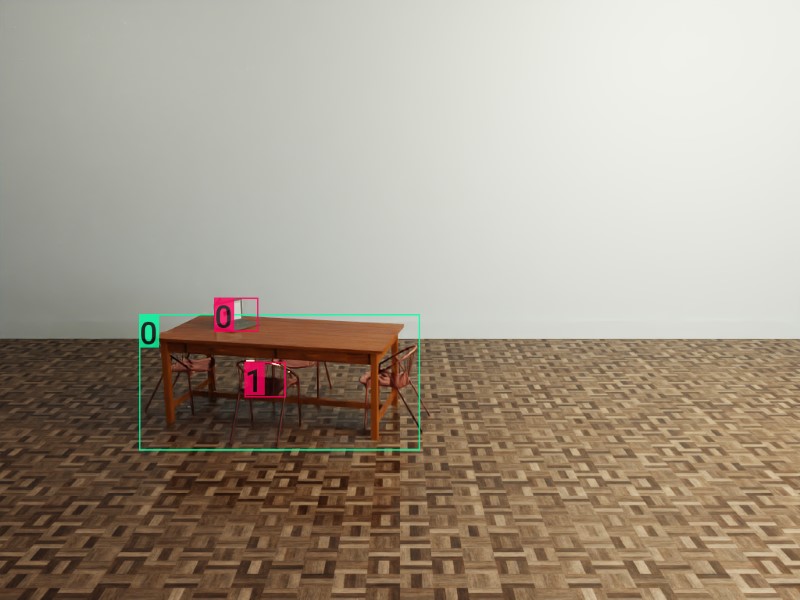} &
      \includegraphics[height=0.17\textwidth]{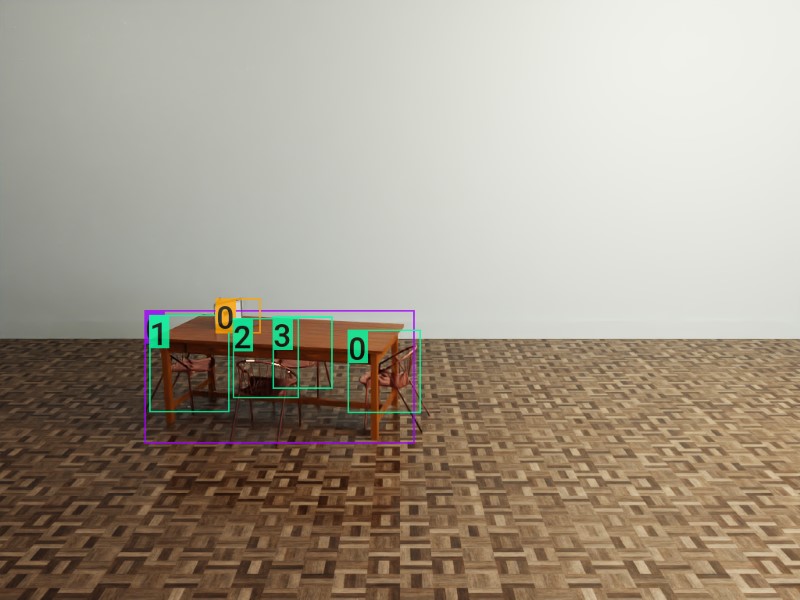} &
      \includegraphics[height=0.17\textwidth]{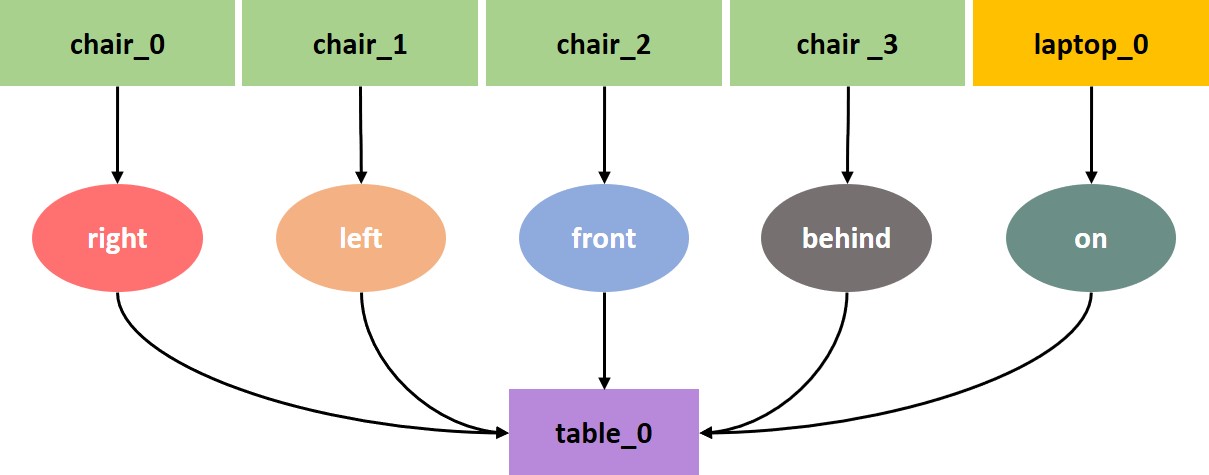} \\
    \end{tabular}
    \caption{Qualitative results of Sim2SG on the target domain for Dining-Sim. First column shows that the SDR~\cite{sdr18} fails to either detect objects or have high number of false positives (mislabels) leading to poor scene graph. Our method detects objects better, has fewer false positives and ultimately generates more accurate scene graphs as shown in second and third column respectively. Objects are color coded.}
    \label{fig:shapenet3d:qualitative}
\end{figure*}
\begin{figure*}
\vspace{-2mm}
    \centering
    \addtolength{\tabcolsep}{-4.6pt}
    \begin{tabular}{cccc}
			 \includegraphics[width=0.22\textwidth]{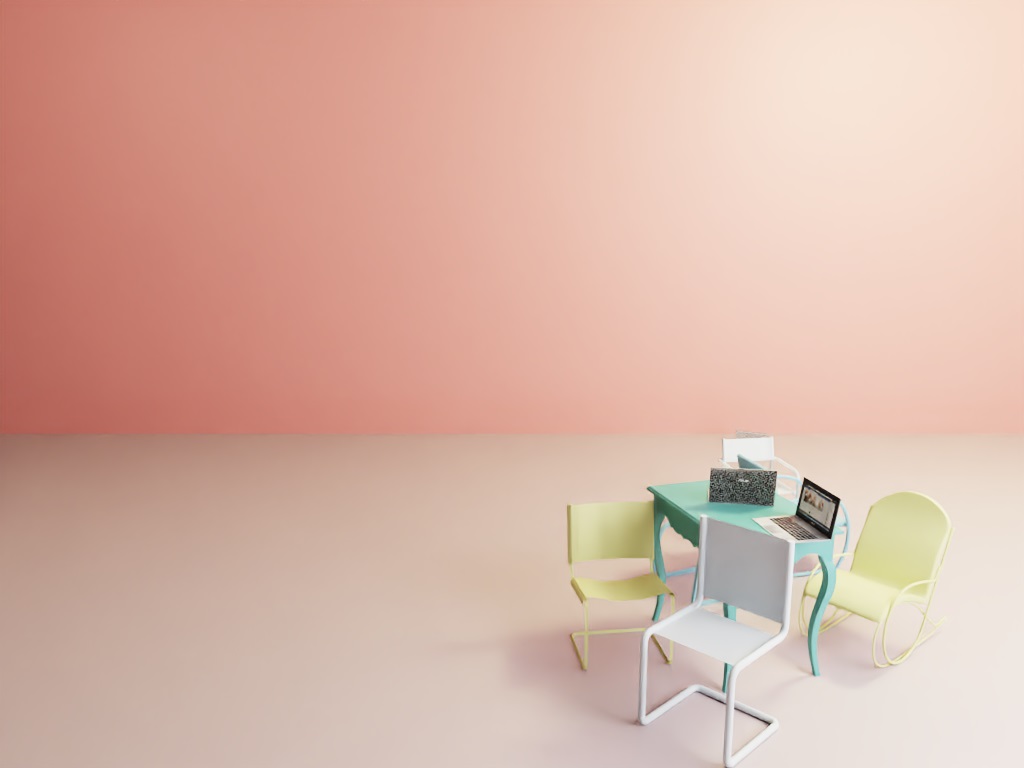} &
      \includegraphics[width=0.22\textwidth]{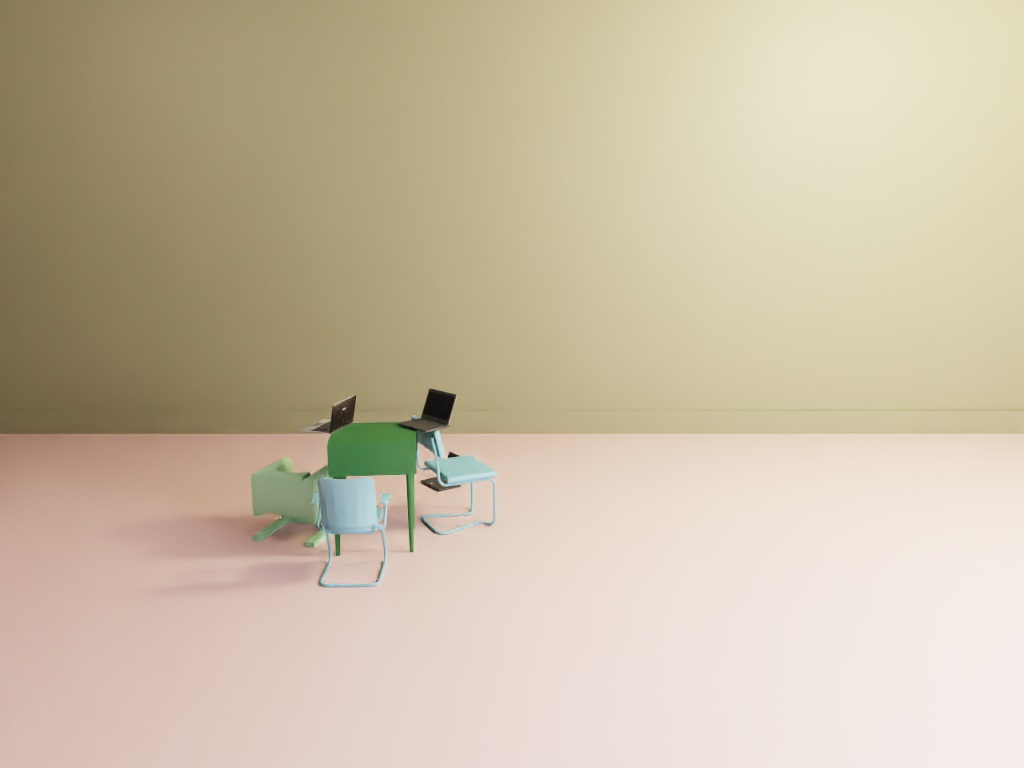} &
      \includegraphics[width=0.22\textwidth]{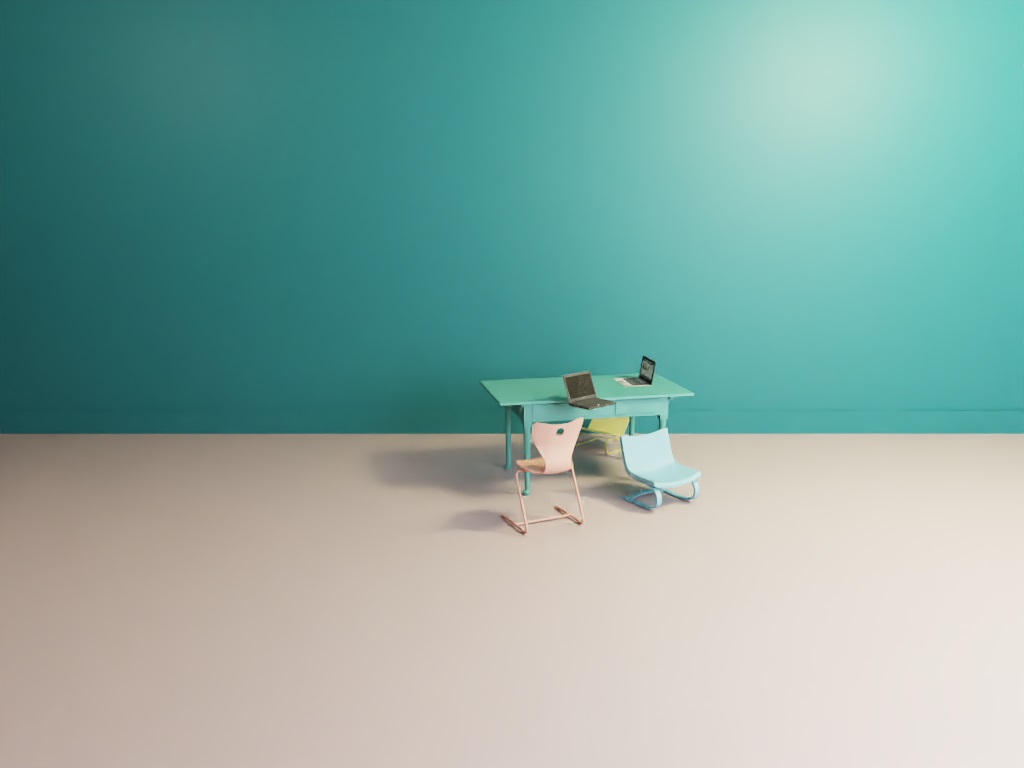} &
      \includegraphics[width=0.22\textwidth]{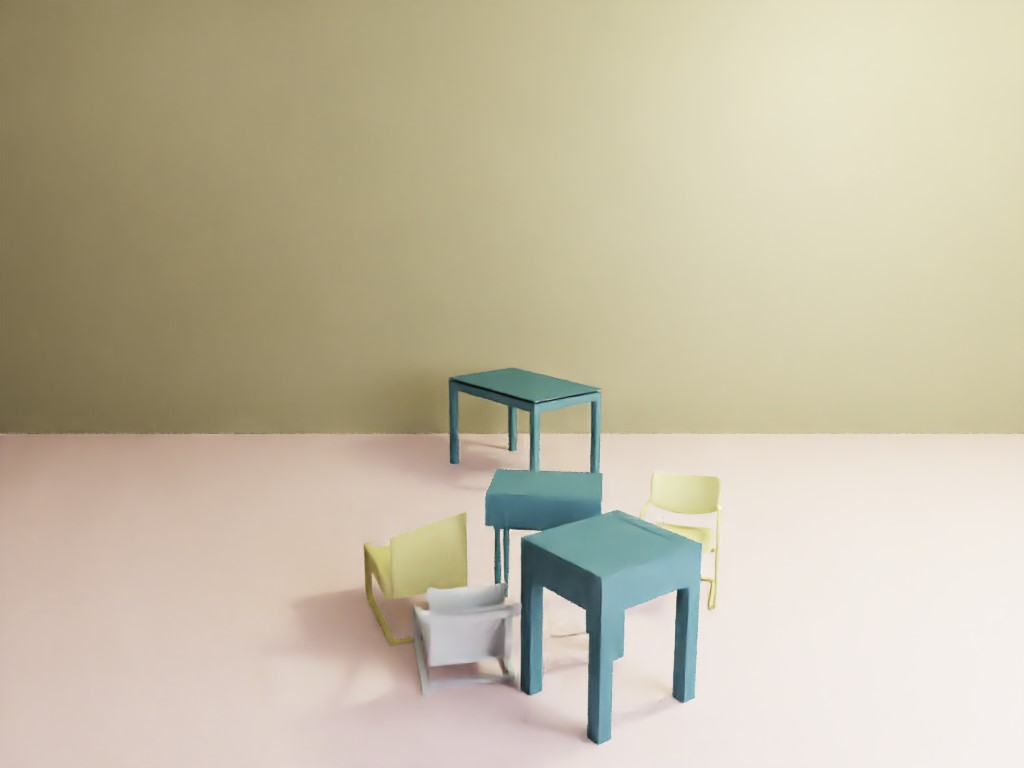} \\
			
			 \includegraphics[width=0.22\textwidth]{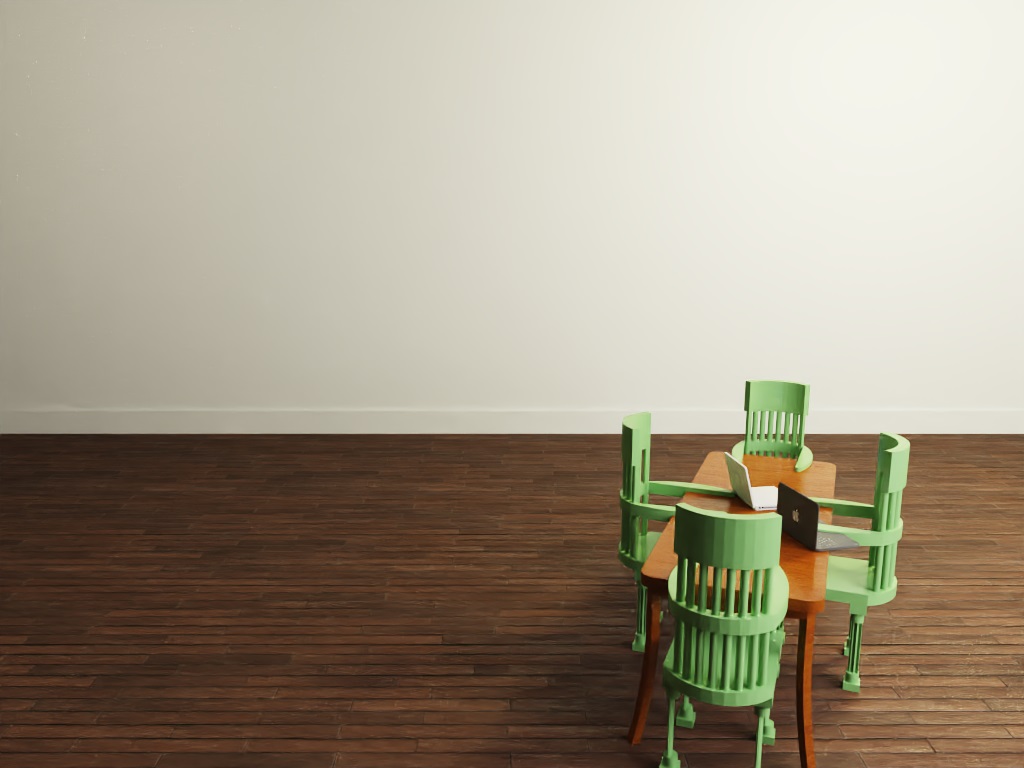} &
      \includegraphics[width=0.22\textwidth]{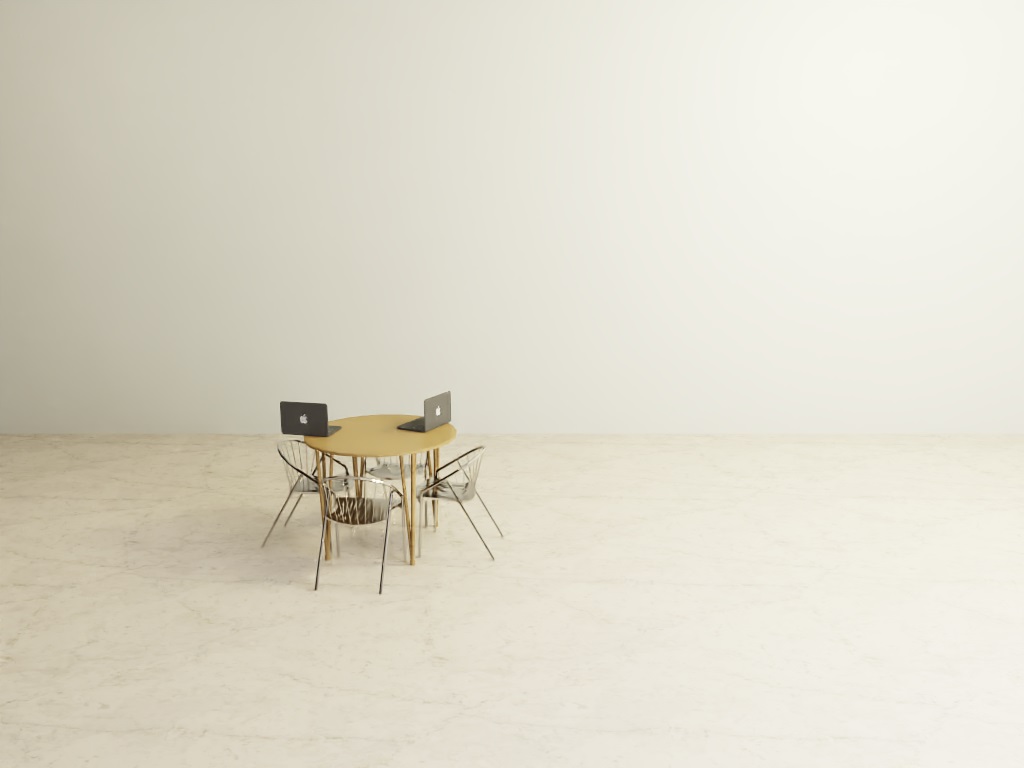} &
      \includegraphics[width=0.22\textwidth]{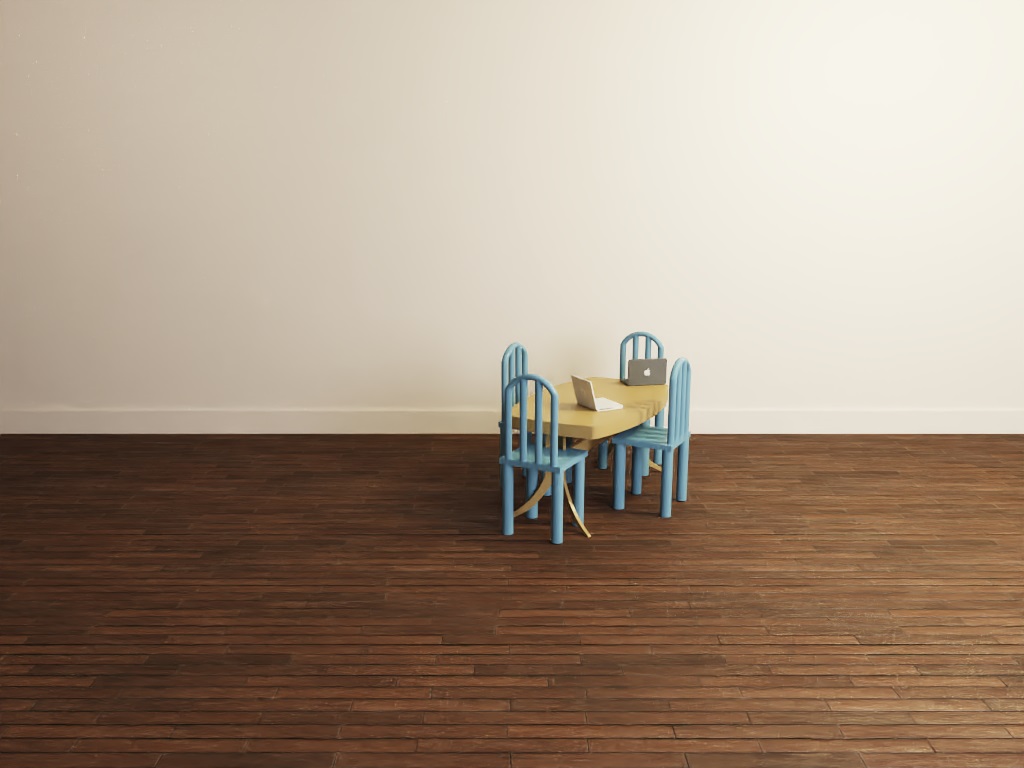} &
      \includegraphics[width=0.22\textwidth]{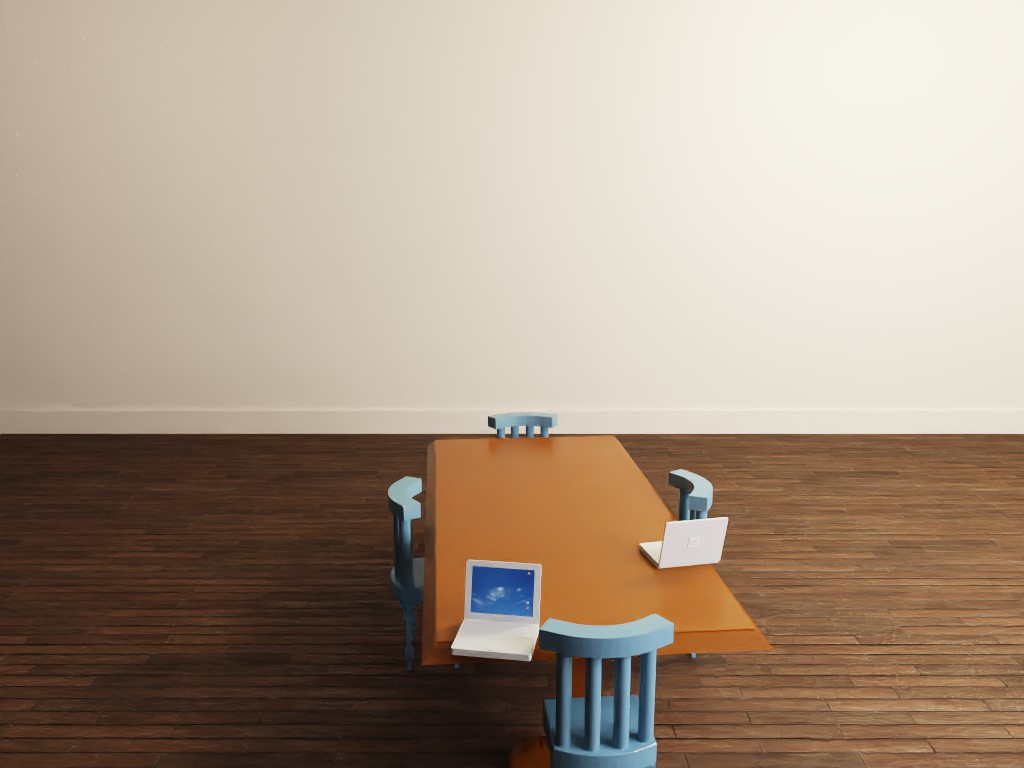} \\
\end{tabular}
\caption{Scenes generated by our method (top) for target samples (bottom) in Dining-Sim environment.}
    \label{fig:shapene3d:reconstruction}
\end{figure*}
\begin{figure*}
\vspace{-2mm}
    \centering
    \addtolength{\tabcolsep}{-4.6pt}
    \begin{tabular}{cccc}
			 \includegraphics[width=0.22\textwidth]{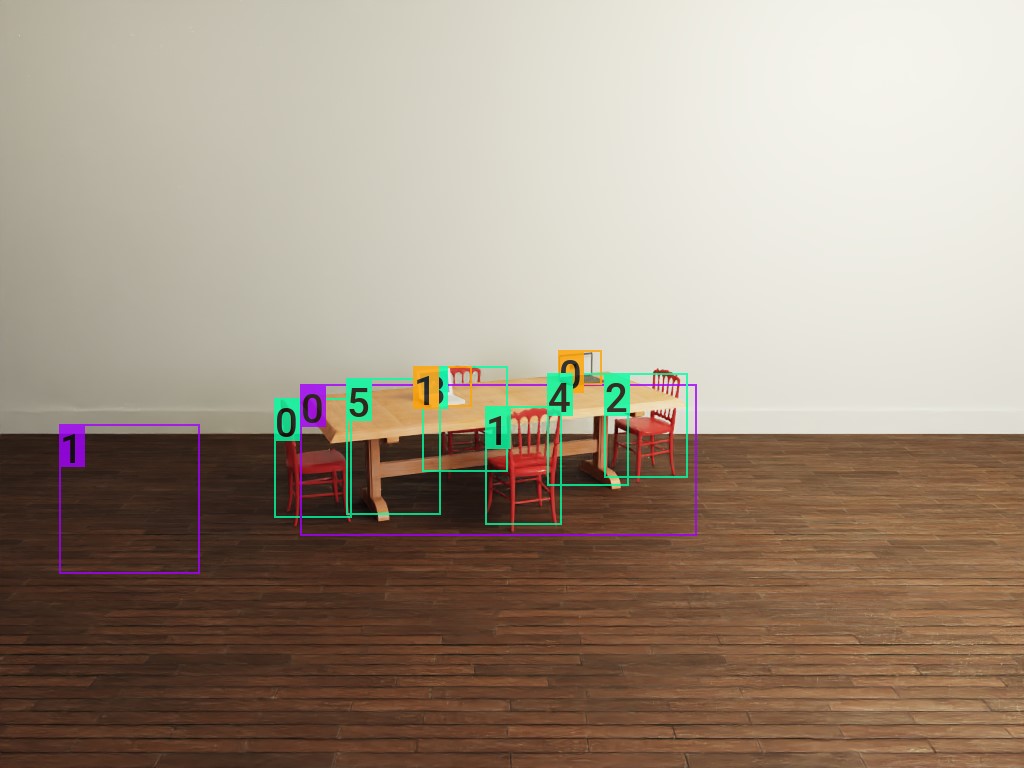} &
			 \includegraphics[width=0.22\textwidth]{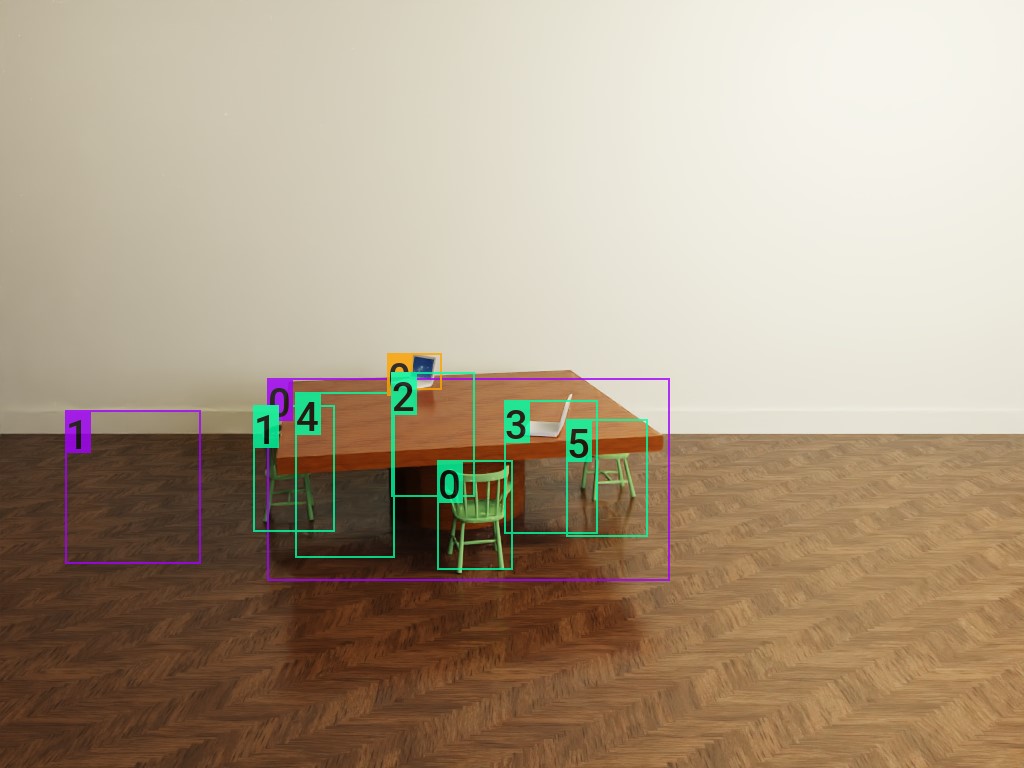} &
			 \includegraphics[width=0.22\textwidth]{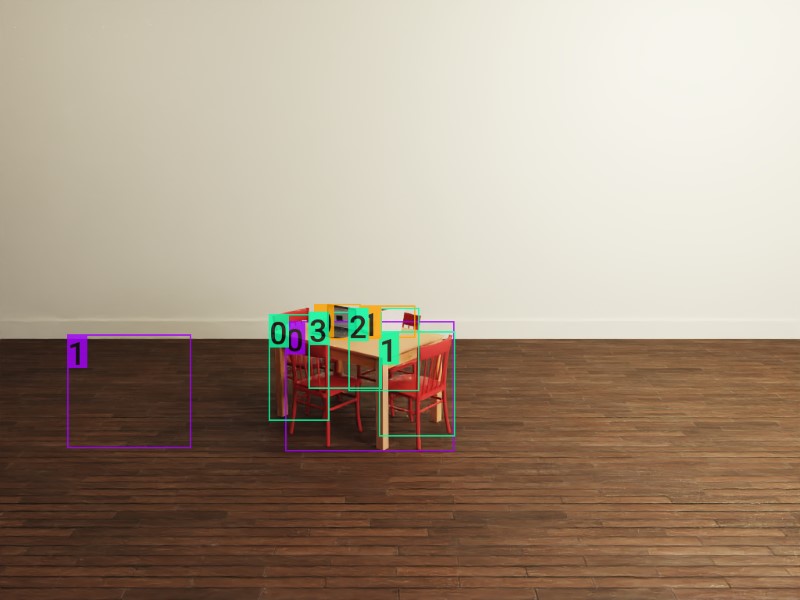} &
			 \includegraphics[width=0.22\textwidth]{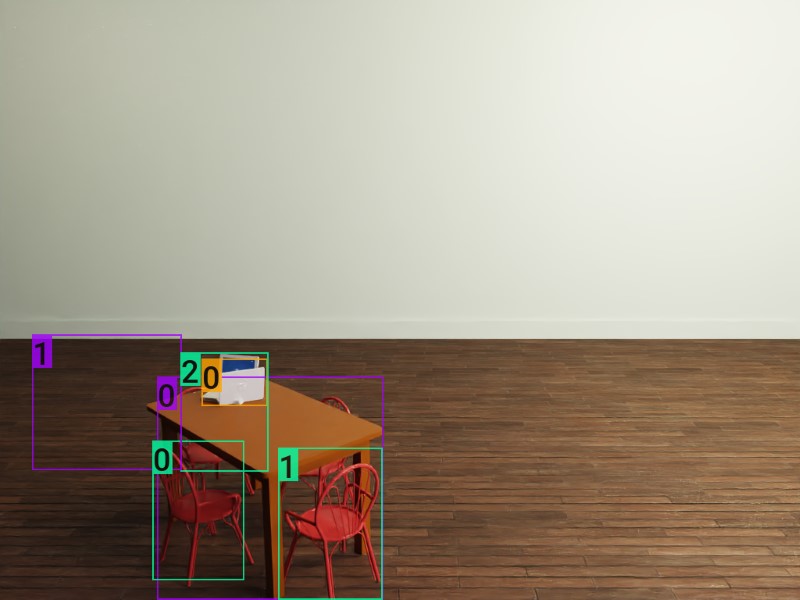} \\
			
			 \includegraphics[width=0.22\textwidth]{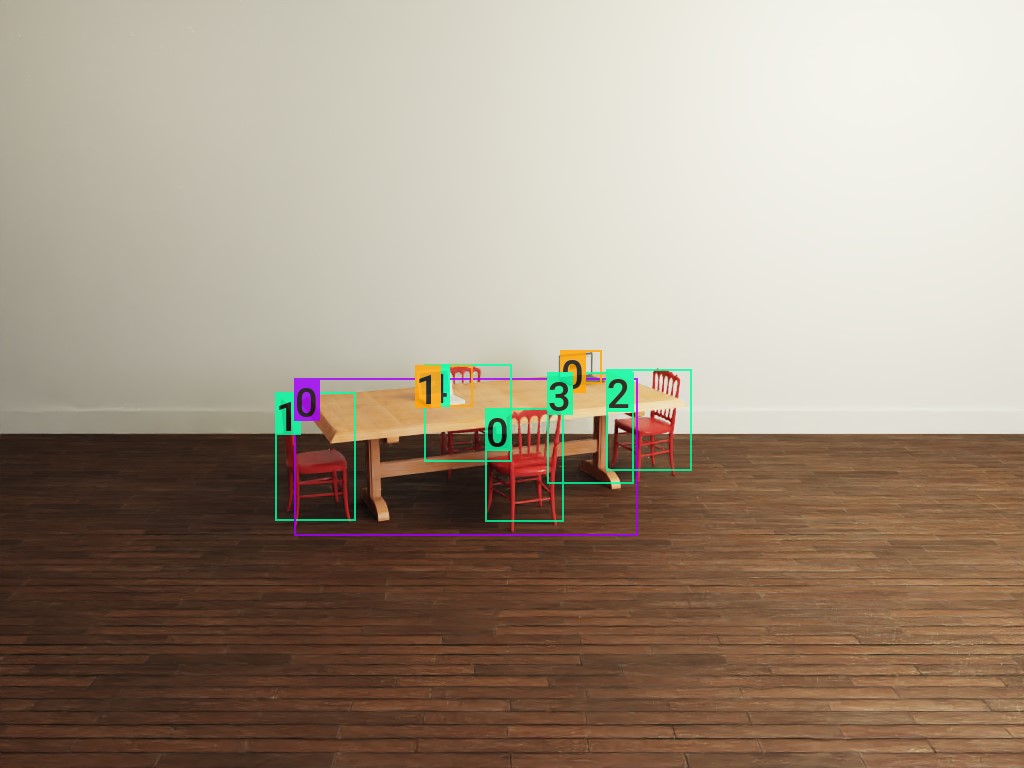} &
			 \includegraphics[width=0.22\textwidth]{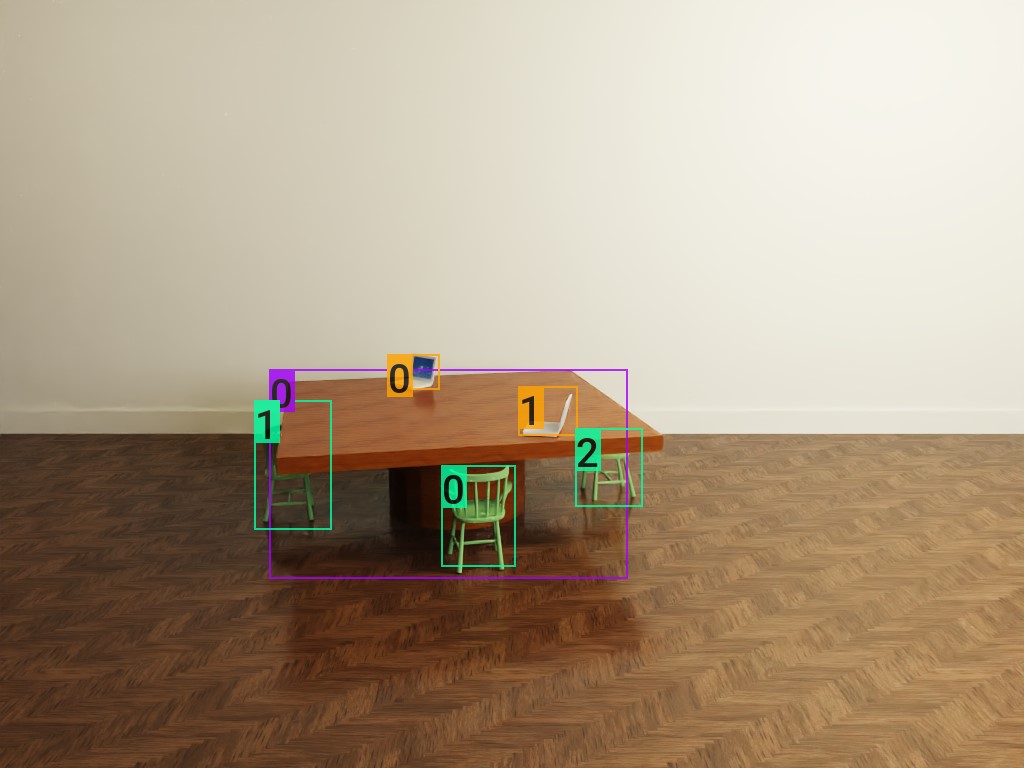} &
			 \includegraphics[width=0.22\textwidth]{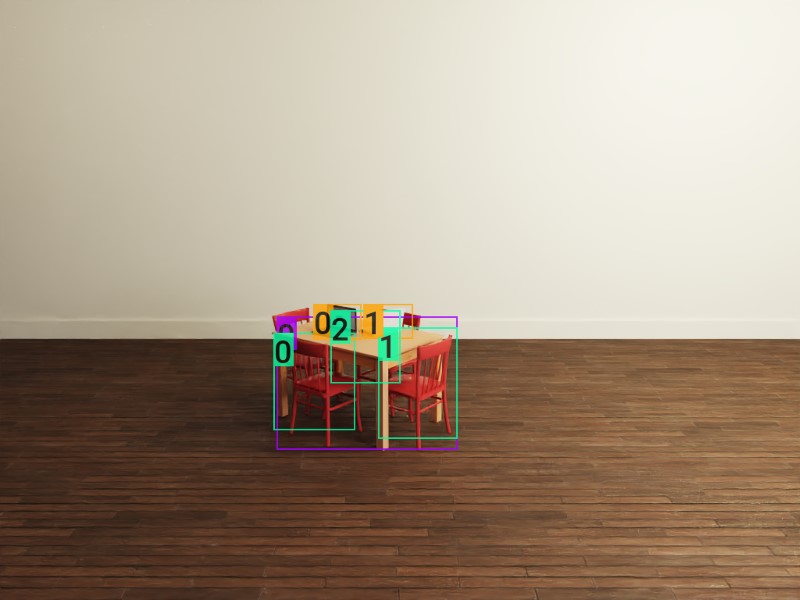} &
			 \includegraphics[width=0.22\textwidth]{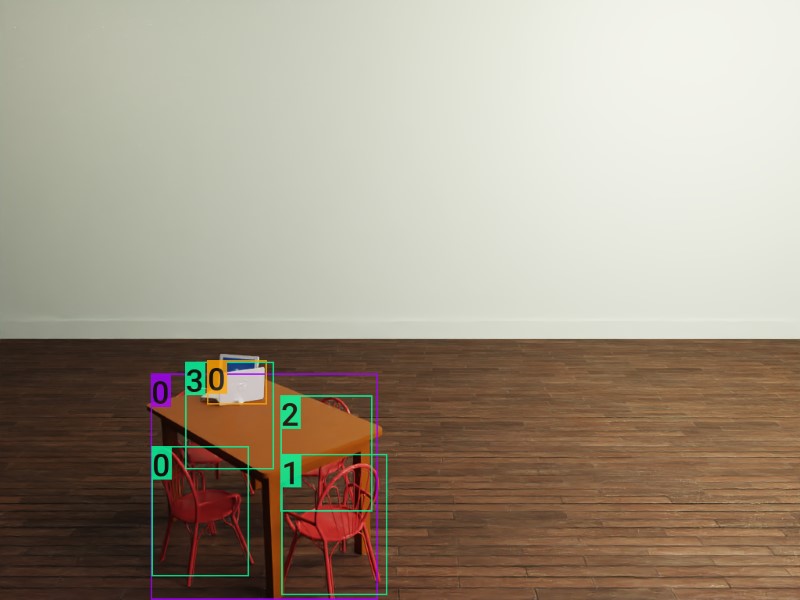} \\

    \end{tabular}
    \vspace{-0.5em}
    \caption{Appearance alignment $\sigma^{a}$ reducing false positive. Top row: $\sigma^{c,label}$,  bottom row: $\sigma^{c,label}$ + $\sigma^{a}$ }
    \label{fig:dining-sim-fp}
\end{figure*}

\paragraph{Results}
We present the full quantitative results in Table~\ref{exp:dining-sim-main} and qualitative results in the Figure~\ref{fig:shapenet3d:qualitative}. We observe that the combination of all alignment terms $\sigma^{c,label}$, $\sigma^a$ and $\sigma^{c,pred}$ gives the best relationship triplet recall of 0.547@50. In order to keep our approach as general as possible, we do not enforce strict rules on object placements and prefer to randomize parameters that are not predicted such as orientation as illustrated in the qualitative results of label alignment $\sigma^{c,label}$ in Figure~\ref{fig:shapene3d:reconstruction}.
When target domain assets are too dissimilar from the assets in the source domain, it often results in incorrect reconstructions as shown in Figure~\ref{fig:shapene3d:reconstruction} (last column).
We also observe that after label alignment $\sigma^{c,label}$, the model occasionally has false positive detections, particularly in areas of the floor that have intricate patterns. We qualitatively show that these false positives disappear with the addition of appearance alignment $\sigma^a$ term
(Figure~\ref{fig:dining-sim-fp}).

\paragraph{Ablations}
We conduct two sets of experiments on the Dining-Sim environment.
The first experiment studies appearance gap: source domain has different appearance but similar content from the target domain. 
The source domain is generated using the target generation scheme but using source dataset materials as shown in Figure~\ref{fig:diningsim}.
The appearance gap because of photo-realistic texture in target domain is from 0.846 Recall@50 to 0.625 Recall@50.
We observe that the appearance alignment $\sigma^a$ helps reduce the appearance gap, increasing relationship triplet recall from 0.625@50 to 0.821@50.
For reference, the oracle performance on the target domain is 0.846 Recall@50.
Similarly, the second experiment studies content gap where the source and target use the same materials but have different assets, object positions and number of objects. We accomplish this by modifying the source generation scheme to select materials from the target dataset. Samples of source and target are shown in the third and fourth rows of Figure~\ref{fig:diningsim}. We observe that the label alignment $\sigma^{c,label}$ term helps reduce the content gap and increase relationship triplet recall from 0.468@50 to 0.539@50.
The relatively modest improvement makes sense as the two domains still differ in content (source and target domain assets differ). Full results are in Table~\ref{exp:dining-sim-ablations}.




\begin{table*}

\centering
\begin{small}
\begin{tabular}{l c c c c c c}
\toprule
 Method & Chair AP & Table AP & Laptop AP & mAP @0.5 IoU & Recall@50   \\
 \midrule
 SDR~\cite{sdr18}  &\textbf{ 0.842 $\pm{0.038}$} & 0.519 $\pm{0.088}$ & 0.392 $\pm{0.051}$ & 0.584 $\pm{0.049}$ & 0.331 $\pm{0.064}$ \\
 Ours ($\sigma^{c,label}$)  & 0.737 $\pm{0.043}$ & 0.724 $\pm{0.030}$ & 0.608 $\pm{0.047}$ & 0.713 $\pm{0.038}$ & 0.501 $\pm{0.044}$ \\
Ours ($\sigma^{c,label}$, $\sigma^{a}$, $\sigma^{c,pred}$)  & 0.770 $\pm{0.022}$ & \textbf{0.757 $\pm{0.037}$} & \textbf{0.659 $\pm{0.005}$} & \textbf{0.729 $\pm{0.015}$} & \textbf{0.547 $\pm{0.015}$} \\
 \bottomrule
\end{tabular}
\end{small}
\caption{\label{exp:dining-sim-main} Quantitative results of Sim2SG on a target domain in Dining-Sim environment.}
\end{table*}

\begin{table*}[!t]

\centering
\begin{small}
\begin{tabular}{l c c }
\toprule
  Method &  mAP @0.5 IoU & Recall@50   \\
 \midrule
 SDR~\cite{sdr18} & 0.772 $\pm{0.043}$ & 0.625 $\pm{0.076}$ \\
 Ours ($\sigma^{a}$) & \textbf{0.878 $\pm{0.001}$} & \textbf{0.821 $\pm{0.006}$} \\
 \bottomrule
\end{tabular}
 \end{small}
\begin{small}
\begin{tabular}{l c c }

\toprule
 Method &  mAP @0.5 IoU & Recall@50   \\
 \midrule
 SDR~\cite{sdr18} &  0.676 $\pm{0.011}$ & 0.468 $\pm{0.006}$ \\
 Ours ($\sigma^{c,label}$) & \textbf{0.737 $\pm{0.024}$} & \textbf{0.539 $\pm{0.006}$} \\
 \bottomrule
\end{tabular}
 \end{small}
 \caption{ \label{exp:dining-sim-ablations} Dining-Sim ablations. Left (resp. right): Source and target domains have different (resp. similar) appearance but similar (resp. different) content distribution. All the evaluations are on the target domain.}
\end{table*}



\subsubsection{Drive-Sim}
\label{appnd:drivesim}

\paragraph{Setup}
As mentioned in Section~\ref{subsc:drivesim}, we use an Unreal Engine 4\footnote{ {\scriptsize \url{https://www.unrealengine.com/}}} based driving simulator akin to~\cite{sdr18} to generate synthetic data. We have cars (1-2 per lane), trees(1-3), houses/buildings(1-3), pedestrians(0-2), sidewalk(2), roads(2-6). We do not have poles, street signs or any other objects. We have straight roads.
We use realistic random placements, e.g. cars can only be  placed on a lane, pedestrians on sidewalk, houses on ground and trees on both sidewalk and ground. We randomize the time of the day, cloud density and use directional light. We assume real world scale.
We place our camera at a car height on a random right lane with fixed camera parameters (0 yaw, 0 pitch, 90 fov). 
We add realistic texture and color to each object similar to~\cite{sdr18}. We use 1242 x 375 image resolution for training and evaluation.

\paragraph{Details on Synthesis Step}
We describe how we generate synthetic data by inferring scene graphs from KITTI~\cite{kitti} in detail.
During synthesis stage, we infer the scene graphs from KITTI and further filter the objects and relationships among them using a confidence threshold of 0.2. 
We do not have access to KITTI camera parameters and we use the camera parameters described in the previous paragraph.
Using the assumed camera parameters (both intrinsic and extrinsic) we project a ray from the camera through the pixel corresponding to the bounding box bottom-centre. The 3D coordinate of the object is then the intersection of the ray and the ground plane, which we assume to be flat at elevation 0. We place each object in the 3D scene by picking a random 3D asset according its type (class) and assigning random pose in the range $0^{\circ}$--$360^{\circ}$ (except cars that are aligned to the lane). We assume contexts like road, ground, sky, sidewalk as described in the previous paragraph. We refine the 3D scene further according to the predicted relationships among objects.
We also assume a consistent lane width, and number of roads are determined by positions of the detected vehicles in the scene. We place multiple Trees (\ie \textit{Vegetation}) if the projected 3D volume permits.
We then render the 3D scene.

\begin{table*}

\centering
\scriptsize
\begin{tabular}{l c c c c c c c c c}
\toprule
   Method & Car  & Pedes.  & House  & Veg.  &  mAP & Recall@50  \\
 \midrule
    SDR~\cite{sdr18} & 0.488 $\pm{0.007}$ &	0.214 $\pm{0.025}$ &	0.223 $\pm{0.022}$ & 	0.177 $\pm{ 0.010}$	 & 0.276 $\pm{0.005}$ &	0.112 $\pm{0.009}$ \\
  Meta-Sim~\cite{meta-sim} &  0.575  $\pm{0.008}$ &	0.227 $\pm{0.024}$ &	0.252 $\pm{0.008}$ & 	0.174 $\pm{ 0.033}$	 & 0.307 $\pm{0.003}$ &	0.143 $\pm{0.007}$ \\
  Self-learning~\cite{pseudolabelyang} & 0.466  $\pm{0.009}$ &	0.215 $\pm{0.016}$ &	0.189 $\pm{0.025}$ & 	0.265 $\pm{ 0.008}$	 & 0.284 $\pm{0.006}$ &	0.129 $\pm{0.006}$ \\
  DA-FasterRCNN~\cite{Chen_2018} & 0.523 $\pm{0.036}$ &	0.209 $\pm{0.038}$ &	0.203 $\pm{0.037}$ & 	0.171 $\pm{ 0.012}$	 & 0.277 $\pm{0.017}$ &	0.119 $\pm{0.022}$ \\
  GPA~\cite{Xu_2020} & 0.248  $\pm{0.053}$ &	0.016 $\pm{0.026}$ &	0.097 $\pm{0.030}$ & 	0.063 $\pm{ 0.031}$	 & 0.109 $\pm{0.022}$ &	0.028 $\pm{0.009}$ \\
  SAPNet~\cite{li2020spatial} & 0.420  $\pm{0.052}$ &	0.124 $\pm{0.035}$ &	0.018 $\pm{0.004}$ & 	0.042 $\pm{ 0.010}$	 & 0.151 $\pm{0.010}$ &	-- \\
Ours ($\sigma^{c,label}$) &  0.566  $\pm{0.033}$ &	\textbf{0.310 $\pm{0.029}$} &	0.261 $\pm{0.009}$ & 	0.242 $\pm{ 0.040}$	 & 0.345 $\pm{0.002}$ &	0.193 $\pm{0.010}$ \\
 Ours ($\sigma^{c,label}$, $\sigma^a$) &  0.606  $\pm{0.021}$ &	0.309 $\pm{0.013}$ &	0.272 $\pm{0.008}$ & 	0.260 $\pm{ 0.021}$	 & 0.362 $\pm{0.007}$ &	0.220 $\pm{0.010}$ \\
   Ours ($\sigma^{c,label}$,  $\sigma^a$, $\sigma^{c,pred}$) &  \textbf{0.623  $\pm{0.033}$} &	0.301 $\pm{0.018}$ & \textbf{0.283 $\pm{0.007}$} & 	\textbf{0.274 $\pm{ 0.015}$} & \textbf{0.370 $\pm{0.005}$} & \textbf{0.240 $\pm{0.003}$} \\
\midrule
 SDR & 0.412  $\pm{0.006}$ &	0.174 $\pm{0.018}$ &	0.215 $\pm{0.022}$ & 	0.177 $\pm{ 0.011}$	 & 0.245 $\pm{0.002}$ &	0.085 $\pm{0.008}$ \\
  Meta-Sim &  0.455  $\pm{0.040}$ &	0.203 $\pm{0.026}$ &	0.242 $\pm{0.009}$ & 	0.176 $\pm{ 0.029}$	 & 0.269 $\pm{0.007}$ &	0.093 $\pm{0.005}$ \\
  Self-learning & 0.377  $\pm{0.007}$ &	0.174 $\pm{0.014}$ &	0.197 $\pm{0.002}$ & 	0.263 $\pm{ 0.006}$	 & 0.253 $\pm{0.004}$ &	0.077 $\pm{0.005}$ \\
   DA-FasterRCNN &  0.472  $\pm{0.028}$ &	0.181 $\pm{0.031}$ &	0.203 $\pm{0.041}$ & 	0.168 $\pm{ 0.013}$	 & 0.256 $\pm{0.012}$ &	0.091 $\pm{0.019}$ \\
  GPA & 0.201  $\pm{0.043}$ &	0.016 $\pm{0.026}$ &	0.106 $\pm{0.031}$ & 	0.065 $\pm{ 0.036}$	 & 0.096 $\pm{0.026}$ &	0.018 $\pm{0.007}$ \\
  SAPNet &  0.419  $\pm{0.068}$ &	0.098 $\pm{0.028}$ &	0.017 $\pm{0.005}$ & 	0.038 $\pm{ 0.008}$	 & 0.143 $\pm{0.017}$ &	-- \\
   Ours ($\sigma^{c,label}$) &  0.471  $\pm{0.033}$ &	\textbf{0.273 $\pm{0.027}$} &	0.244 $\pm{0.010}$ & 	0.233 $\pm{ 0.036}$	 & 0.305 $\pm{0.006}$ &	0.128 $\pm{0.008}$ \\
 Ours ($\sigma^{c,label}$, $\sigma^a$) & 0.511  $\pm{0.002}$ &	0.266 $\pm{0.014}$ &	0.251 $\pm{0.013}$ & 	0.256 $\pm{ 0.021}$	 & 0.321 $\pm{0.004}$ &	0.155 $\pm{0.005}$ \\
  Ours ($\sigma^{c,label}$, $\sigma^a$, $\sigma^{c,pred}$) & \textbf{0.529  $\pm{0.029}$} &	0.249 $\pm{0.017}$ & \textbf{0.262 $\pm{0.011}$} & 	\textbf{0.270 $\pm{ 0.015}$} & \textbf{0.328 $\pm{0.007}$} & \textbf{0.170 $\pm{0.004}$} \\
\midrule
 SDR  & 
  0.382  $\pm{0.029}$ &	0.168 $\pm{0.017}$ &	0.211 $\pm{0.023}$ & 	0.174 $\pm{ 0.010}$	 & 0.234 $\pm{0.006}$ &	0.070$\pm{0.007}$ \\
  Meta-Sim &  0.413  $\pm{0.009}$ &	0.197 $\pm{0.027}$ &	0.236 $\pm{0.009}$ & 	0.164 $\pm{ 0.023}$	 & 0.253 $\pm{0.003}$ &	0.075 $\pm{0.005}$ \\
  Self-learning & 0.312  $\pm{0.006}$ &	0.167 $\pm{0.015}$ &	0.191 $\pm{0.003}$ & 	0.263 $\pm{ 0.006}$	 & 0.233 $\pm{0.004}$ &	0.062 $\pm{0.003}$ \\
   DA-FasterRCNN &  0.424  $\pm{0.028}$ &	0.170 $\pm{0.029}$ &	0.200 $\pm{0.041}$ & 	0.169 $\pm{ 0.014}$	 & 0.241 $\pm{0.014}$ &	0.074 $\pm{0.015}$ \\
  GPA & 0.174  $\pm{0.040}$ &	0.011 $\pm{0.016}$ &	0.106 $\pm{0.031}$ & 	0.059 $\pm{ 0.027}$	 & 0.087 $\pm{0.020}$ &	0.015 $\pm{0.005}$ \\
  SAPNet & 0.362  $\pm{0.054}$ &	0.085 $\pm{0.051}$ &	0.116 $\pm{0.021}$ & 	0.067 $\pm{ 0.022}$	 & 0.157 $\pm{0.024}$ &	-- \\
 Ours ($\sigma^{c,label}$) &  0.410  $\pm{0.009}$ &	\textbf{0.262 $\pm{0.025}$} &	0.240 $\pm{0.010}$ & 	0.229 $\pm{ 0.036}$	 & 0.285 $\pm{0.003}$ &	0.104 $\pm{0.006}$ \\
  Ours ($\sigma^{c,label}$, $\sigma^a$) &  0.493  $\pm{0.004}$ &	0.252 $\pm{0.014}$ &	0.247 $\pm{0.012}$ & 	0.253 $\pm{ 0.020}$	 & 0.311 $\pm{0.311}$ &	0.127 $\pm{0.004}$ \\
  Ours ($\sigma^{c,label}$, $\sigma^a$, $\sigma^{c,pred}$) & \textbf{0.501  $\pm{0.006}$} &	0.241 $\pm{0.018}$ & \textbf{0.254 $\pm{0.010}$} & 	\textbf{0.269 $\pm{ 0.014}$} & \textbf{0.316 $\pm{0.004}$} & \textbf{0.139 $\pm{0.004}$} \\
 \bottomrule
\end{tabular}
\caption{\label{tab:kitti_supp} Evaluation on three modes of KITTI : easy (top), moderate (middle), hard (bottom) when training on the labeled synthetic data and unlabeled real data. The class specific AP and mAP are reported at 0.5 IoU.}
\end{table*}

\begin{table}

\centering
\scriptsize
\begin{tabular}{l c c c c c c c c c}
\toprule
   Method & Recall@20  & Recall@50  & Recall@100  \\
 \midrule
 SDR  & 
  0.098 &	0.131 &	0.146 \\
  Meta-Sim &  0.109  &	0.149 &	0.164 \\
  Ours ($\sigma^{c,label}$, $\sigma^a$, $\sigma^{c,pred}$) & \textbf{0.184 } & \textbf{0.235} & \textbf{0.252} \\
\midrule
 SDR  & 
  0.067 &	0.088 &	0.099 \\
  Meta-Sim &  0.071  &	0.094 &	0.104 \\
  Ours ($\sigma^{c,label}$, $\sigma^a$, $\sigma^{c,pred}$) & \textbf{0.132 } & \textbf{0.167} & \textbf{0.181} \\
\midrule
 SDR  & 
  0.053 &	0.071 &	0.079 \\
  Meta-Sim &  0.058  &	0.076 &	0.085 \\
  Ours ($\sigma^{c,label}$, $\sigma^a$, $\sigma^{c,pred}$) & \textbf{0.107 } & \textbf{0.137} & \textbf{0.150} \\
 \bottomrule
\end{tabular}
\caption{\label{tab:kitti_recall_supp} Recall on three modes of KITTI : easy (top), moderate (middle), hard (bottom) when training on the labeled synthetic data and unlabeled real data. The evaluation is performed once.}
\end{table}

\begin{figure*}
\vspace{-2mm}
    \centering
    \addtolength{\tabcolsep}{-4.6pt}
    \begin{tabular}{cc}
      \includegraphics[width=0.48\textwidth]{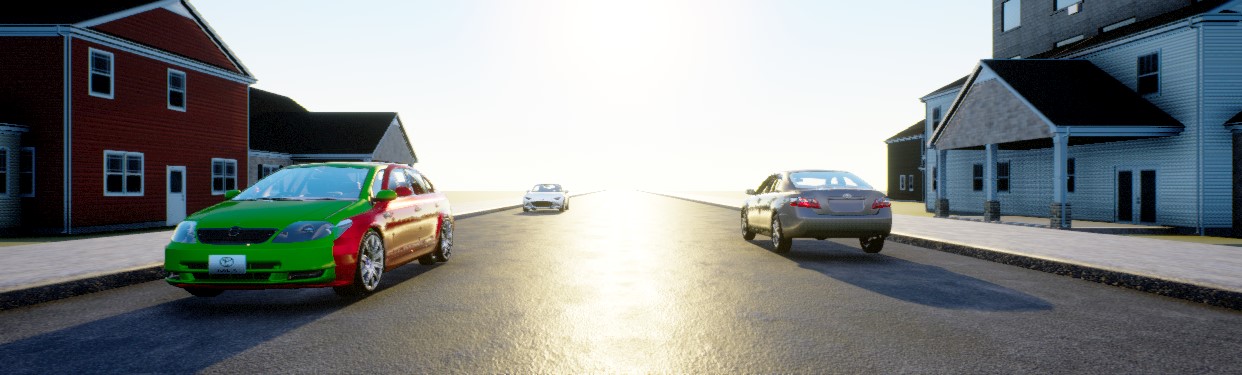} &
      \includegraphics[width=0.48\textwidth]{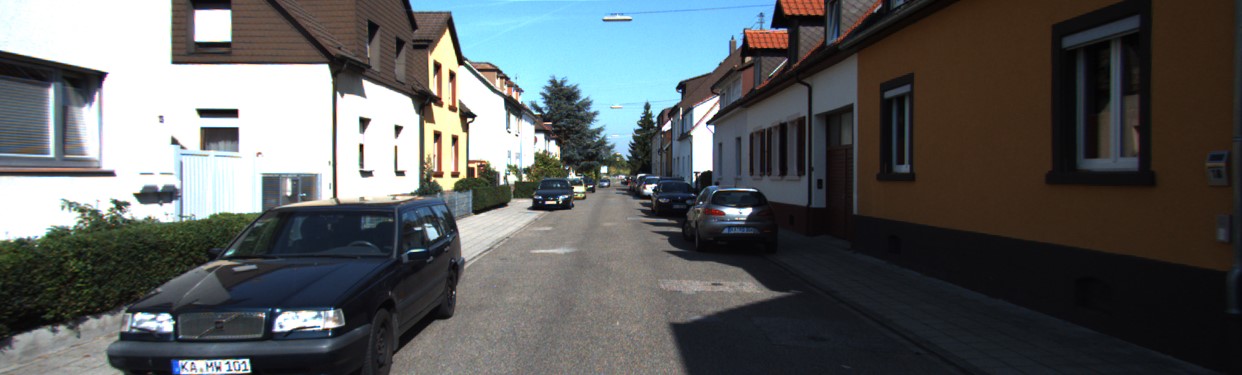} \\
			
      \includegraphics[width=0.48\textwidth]{figures/results/reconstruction/drive-3d/A/000159.jpg} &
      \includegraphics[width=0.48\textwidth]{figures/results/reconstruction/drive-3d/B/000159.jpg} \\
      
      \includegraphics[width=0.48\textwidth]{figures/results/reconstruction/drive-3d/A/000150.jpg} &
      \includegraphics[width=0.48\textwidth]{figures/results/reconstruction/drive-3d/B/000150.jpg} \\
      
    \includegraphics[width=0.48\textwidth]{figures/results/reconstruction/drive-3d/A/000059.jpg} &
      \includegraphics[width=0.48\textwidth]{figures/results/reconstruction/drive-3d/B/000059.jpg} \\
     
    \includegraphics[width=0.48\textwidth]{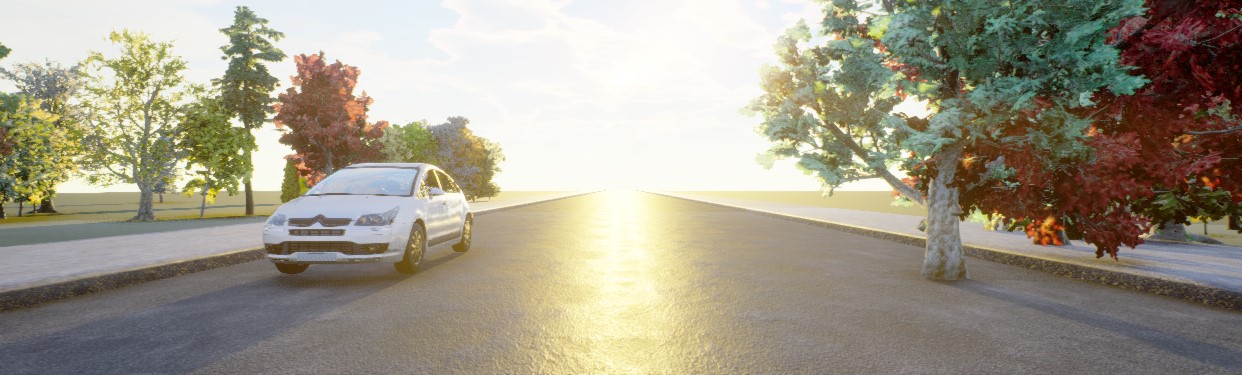} &
     \includegraphics[width=0.48\textwidth]{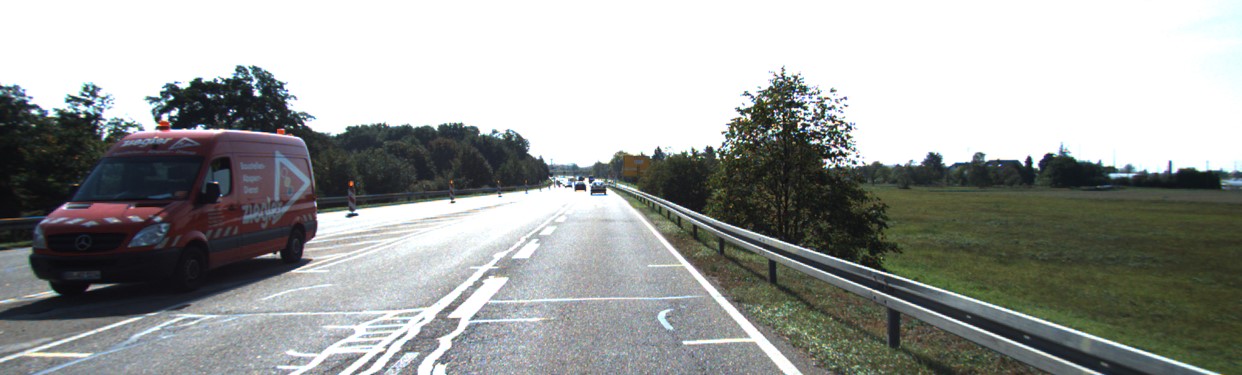} \\
     
    \end{tabular}
    \caption{Scenes generated by our method (left) for target KITTI samples (right).}
    \label{fig:drived3d:reconstruction_full}
\end{figure*}

\begin{figure*}
\addtolength{\tabcolsep}{-4.6pt}
    \centering
    \begin{tabular}{cc}
    
    \includegraphics[width=0.48\textwidth]{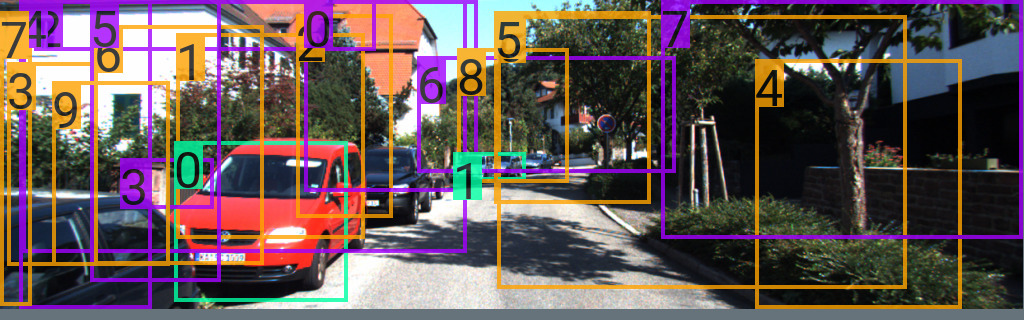} &
    \includegraphics[width=0.48\textwidth]{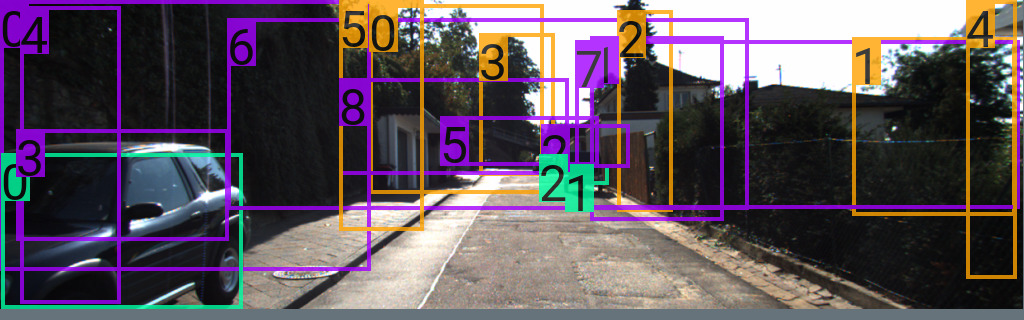} \\
      
        \includegraphics[width=0.48\textwidth]{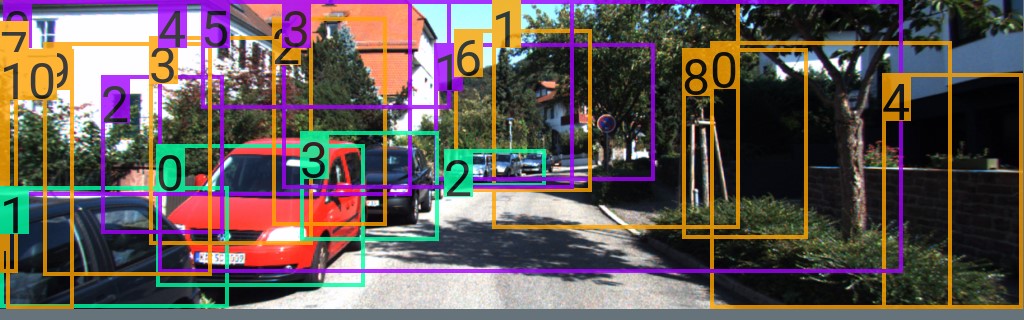} &
          \includegraphics[width=0.48\textwidth]{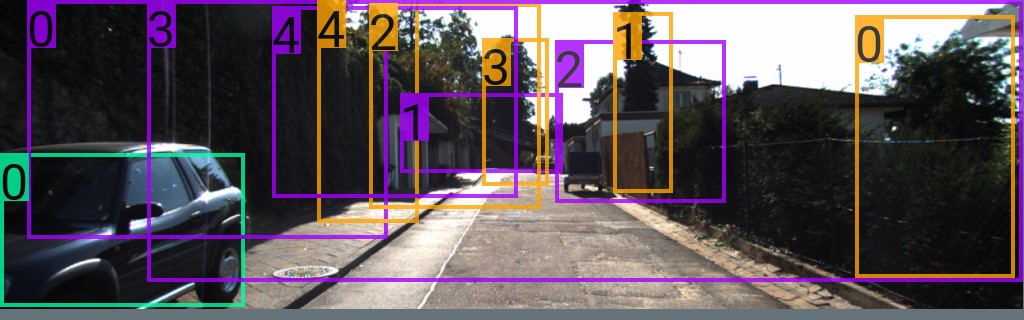} \\
      
         \includegraphics[width=0.48\textwidth]{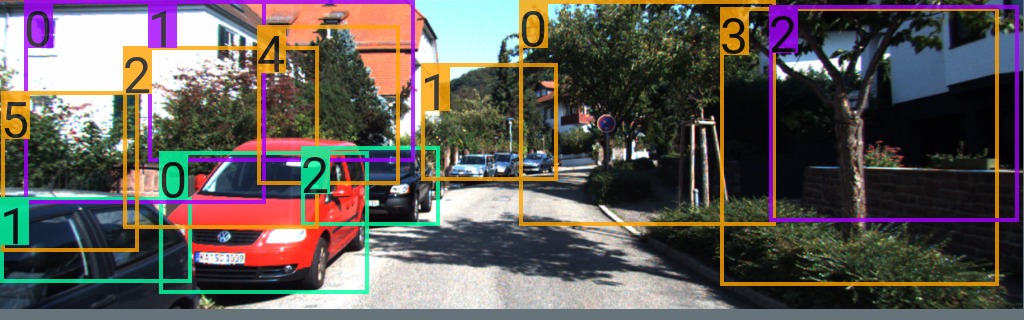} &
      \includegraphics[width=0.48\textwidth]{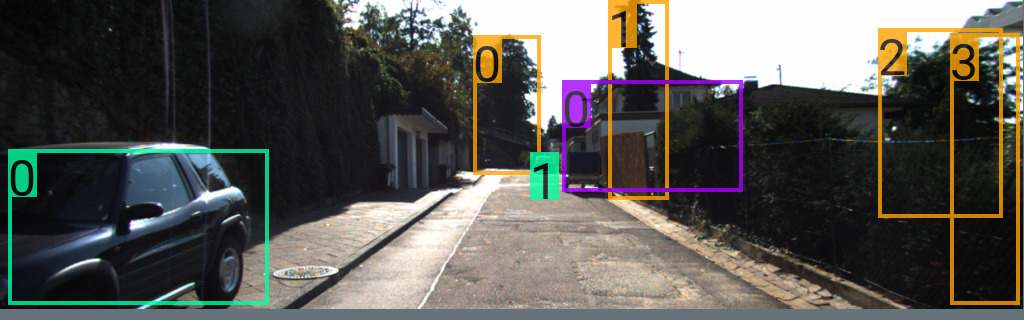} \\
      
    \end{tabular}
    \caption{Qualitative results of objects detected on three different KITTI images.
    Top: SDR fails to detect many objects and yields a large number of false positives (mislabels), leading to poor scene graphs (not shown). Middle:  Meta-Sim improves on false-positives, but still fails to detect some objects. 
    Bottom:  Our method detects objects correctly with fewer false positives, thus generating more accurate scene graphs. 
    (Cars in green, vegetation in yellow, buildings in purple.)}
    \label{fig:drivesim:qualitative_detection}
\end{figure*}

\paragraph{Training Details}
We optimize the model using a SGD optimizer with learning rate of $10^{-4}$ and momentum of 0.9. We train our model using a batch size 2 on NVIDIA DGX workstations. We report saturation peak performance in all our tables.
We give equal regularization weights to source task loss $\sigma_s$, appearance alignment $\sigma^a$, prediction alignment $\sigma^{c,pred}$ and label alignment $\sigma^{c,label}$. 

We first train our model using label alignment $\sigma^{c,label}$) for 3 epochs each with $10^{4}$ iterations.
We then add appearance alignment $\sigma^a$ and prediction alignment $\sigma^{c,pred}$ and train for an additional 60k iterations.
This makes sense as  $\sigma^a$ works better when content/labels are aligned between the two domains.
The total training takes 12 hours including the rendering time.

\textbf{Baselines}:
We adapt domain adaptation baselines~\cite{Chen_2018, Xu_2020} to our framework by using the same backbone (Resnet 101) and SG Predictor (GraphRCNN~\cite{Yang_2018}) network as Sim2SG, but their loss function.
We do not adapt SAPNet~\cite{li2020spatial}.
We train these baselines on 6000 images from the source domain~\cite{sdr18} using the same optimizer and learning rate as Sim2SG for $6 \times 10^{4}$ iterations.
We found GPA ~\cite{Xu_2020} and SAPNet ~\cite{li2020spatial} detection performance to be lower than that reported in their work especially for pedestrian, vegetation and house classes. It is worth noting that their reported class-wise performance numbers only overlap with some of the classes in our work.

We train ~\cite{meta-sim} for 40 epochs with a batch size of 16 and learning rate 0.001 as per the authors. We then obtain 6000 images and train it on Sim2SG framework (Resnet 101 backbone and GraphRCNN SG predictor) for $6 \times 10^{4}$ iterations using the same optimizer and learning rate as Sim2SG.
For self-learning based on pseudo labels~\cite{pseudolabelyang}, we obtain the pseudo labels on KITTI images using the most confident predictions by synthetic pretrained GraphRCNN network (as per the authors). We then train these labeled KITTI images on Sim2SG framework for 60k iterations using the same optimizer and learning rate as Sim2SG.

\paragraph{KITTI Annotation}
We use the existing bounding box annotations of Vehicle and Pedestrians. We annotate Trees and Houses/Buildings of all sizes, occlusion and truncation in KITTI. We use the available camera parameters to project the 2D bounding box into 3D space to help us annotate spatial relationships--front, behind, left and right.

\begin{figure*}
\addtolength{\tabcolsep}{-4.6pt}
    \centering
    \begin{tabular}{c}
     \includegraphics[width=0.8\textwidth]{supplemental/figures/drivesim/sim2sg/000185.jpg} \\
    \includegraphics[height=0.4\textwidth]{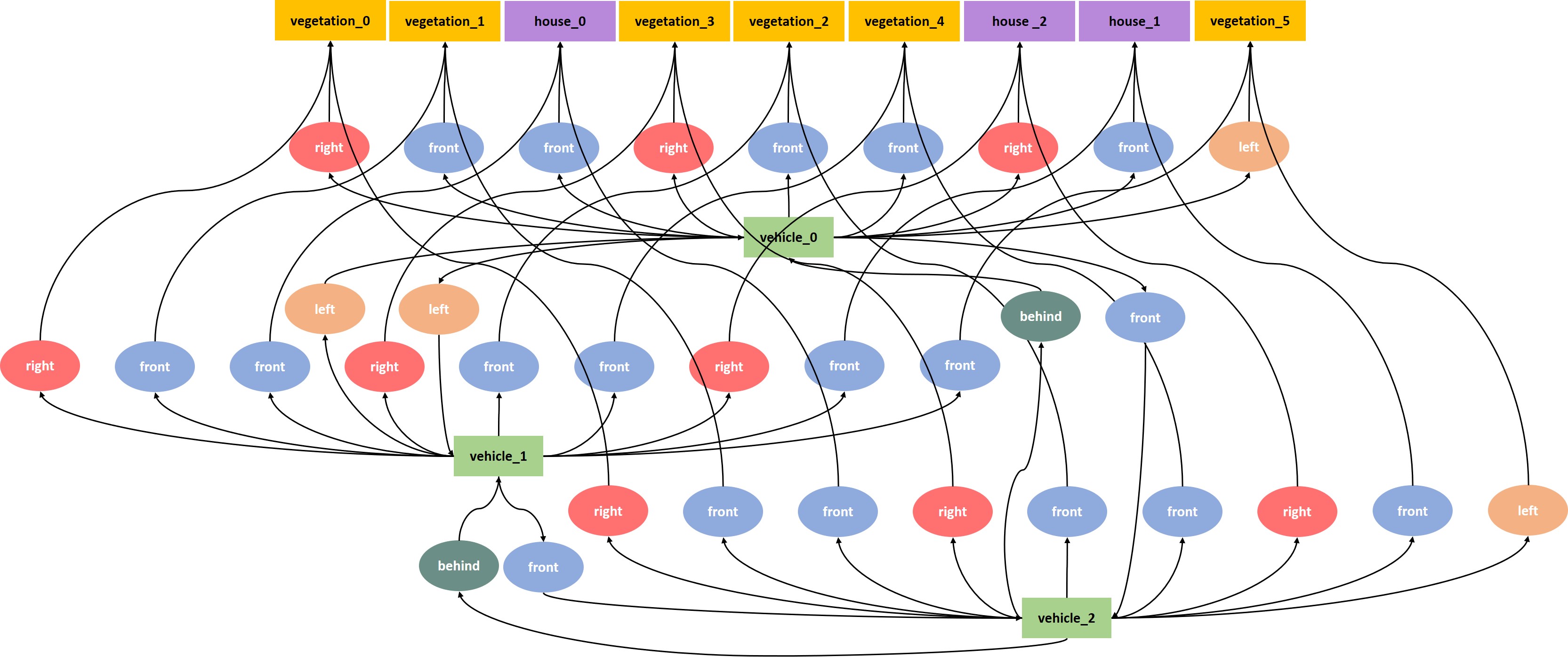} \\
     \includegraphics[width=0.8\textwidth]{supplemental/figures/drivesim/sim2sg/000682.jpg} \\
    \includegraphics[height=0.25\textwidth]{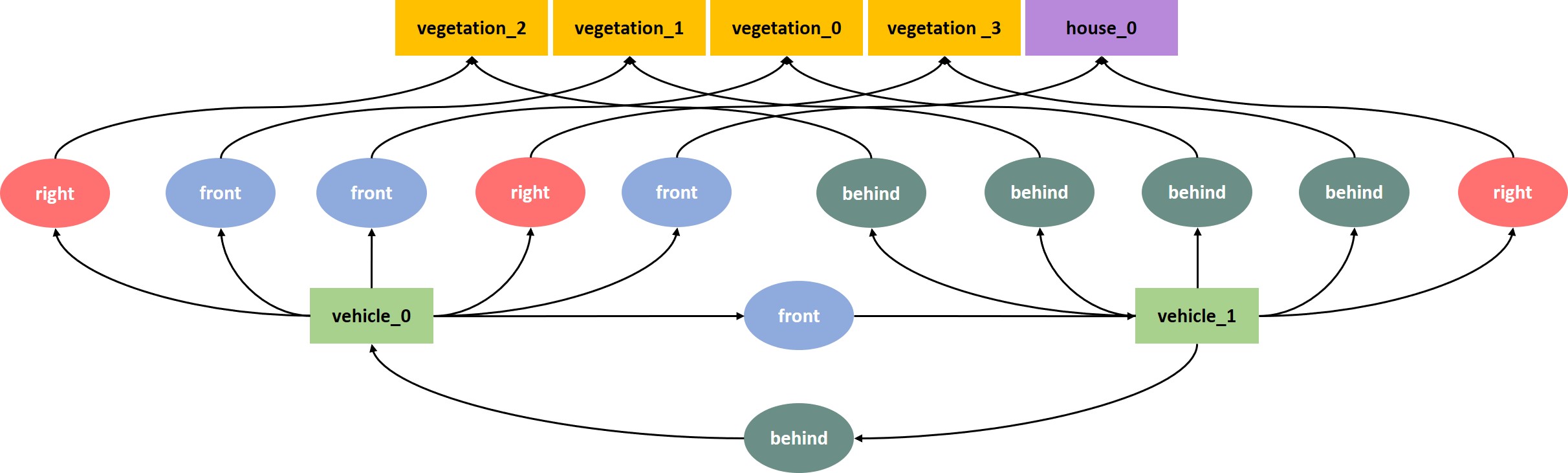} \\
      
    \end{tabular}
    \caption{Qualitative results of Sim2SG on KITTI. Sim2SG generates accurate scene graphs.}
    \label{fig:drivesim_qualitative_sg}
\end{figure*}

\paragraph{Results}
Full quantitative evaluations results are in Table~\ref{tab:kitti_supp} on all KITTI~\cite{kitti} evaluation criteria-- easy, moderate and hard. In all three criteria, Sim2SG is able to achieve significantly better results (higher detection mAP @0.5 IoU and relationship triplet recall @ 50) than all the baselines. We also report recall @20, 100 for few baselines and our approach in Table~\ref{tab:kitti_recall_supp}. More qualitative results of label alignment $\sigma^{c,label}$ is in Figure~\ref{fig:drived3d:reconstruction_full}. We show qualitative improvements (better object recall and fewer false positive detections) over SDR~\cite{sdr18} and Meta-Sim~\cite{meta-sim} in Figure~\ref{fig:drivesim:qualitative_detection} and the corresponding accurate and full scene graphs in Figure~\ref{fig:drivesim_qualitative_sg}.

\end{document}